\documentclass[12pt]{article}
\usepackage{amsmath}
\usepackage{times}
\usepackage{graphicx}
\usepackage{color}
\usepackage{multirow}
\usepackage{ccaption}
\usepackage{xspace}
\usepackage{amsfonts}
\usepackage{apacite}

\usepackage[round]{natbib}
\usepackage{rotating}
\usepackage{bbm}
\usepackage{latexsym}
\usepackage{lineno}

\renewcommand{\cite}{\citep}
\def\ctsc{SSC\xspace}

\newcommand{\captionfonts}{\normalsize}

\makeatletter  
\long\def\@makecaption#1#2{%
  \vskip\abovecaptionskip
  \sbox\@tempboxa{{\captionfonts #1: #2}}%
  \ifdim \wd\@tempboxa >\hsize
    {\captionfonts #1: #2\par}
  \else
    \hbox to\hsize{\hfil\box\@tempboxa\hfil}%
  \fi
  \vskip\belowcaptionskip}
\makeatother   

\usepackage{microtype}
\usepackage{graphicx}
\usepackage{appendix}
\usepackage[margin=1in]{geometry}
\usepackage{booktabs} 
\usepackage{circuitikz}
\usepackage{caption}
\usepackage{amsmath}
\usepackage{algpseudocode}
\usepackage[subrefformat=parens,labelformat=parens]{subfig}
\usepackage{algorithm}

\newcommand{\LSC}{LSC}
\newcommand{\mb}{\mathbf}
\newcommand{\sign}{\text{sign}}
\ctikzset{bipoles/resistor/height=0.10}
\ctikzset{bipoles/resistor/width=0.3}

\begin{document}


\noindent
{\Large 
Learning and Inference in Sparse Coding Models with Langevin Dynamics}

\ \\
{\bf Michael Y.-S. Fang$^{\displaystyle 1, \displaystyle 2}$, Mayur Mudigonda$^{\displaystyle 2,3}$, Ryan Zarcone$^{\displaystyle 2,4}$, Amir Khosrowshahi$^{\displaystyle 2,6}$, Bruno A. Olshausen$^{\displaystyle 2,3,5}$,  }\\
{$^{\displaystyle 1}$Department of Physics, University of California Berkeley, Berkeley, CA, USA.}\\
{$^{\displaystyle 2}$Redwood Center for Theoretical Neuroscience, University of California Berkeley, Berkeley, CA, USA.}\\
{$^{\displaystyle 3}$
Vision Science Graduate Group, School of Optometry, University of California, Berkeley, CA, USA.}\\
{$^{\displaystyle 4}$Biophysics Graduate Group, University of California Berkeley, Berkeley, CA, USA.}\\
{$^{\displaystyle 5}$Helen Wills Neuroscience Institute and School of Optometry, University of California Berkeley, Berkeley, CA, USA.}\\
{$^{\displaystyle 6}$
Intel Corporation, Santa Clara, CA, USA}\\

%


\thispagestyle{empty}
\markboth{}{NC instructions}
%
%
\begin{center} {\bf Abstract} \end{center} 

We describe a stochastic, 
dynamical system capable of inference and learning in a probabilistic latent variable model.
The most challenging problem in such models -- sampling the posterior distribution over latent variables -- is proposed to be solved by harnessing natural sources of stochasticity inherent in electronic and neural systems.  
We demonstrate this idea for a sparse coding model by deriving a continuous-time equation for inferring its latent variables via Langevin dynamics.  The model parameters are learned by simultaneously evolving according to another continuous-time equation, 
thus bypassing the need for digital accumulators or a global clock.  
Moreover we show that Langevin dynamics lead to an efficient procedure for sampling from the posterior distribution in the `$L_0$ sparse' regime, where latent variables are encouraged to be set to zero as opposed to having a small $L_1$ norm.  This allows the model to properly incorporate the notion of sparsity rather than having to resort to a relaxed version of sparsity to make optimization tractable.
Simulations of the proposed dynamical system on both synthetic and natural image datasets demonstrate that the model is capable of probabilistically correct inference, enabling learning of the dictionary as well as parameters of the prior.  


\section{Introduction}


Latent variable models such as sparse coding \cite{olshausen1997sparse} and 
Boltzmann machines \cite{hinton1983optimal,ackley1985learning} have been shown to be powerful and flexible tools in machine learning. However, training such models properly requires sampling from probability distributions over the latent variables. 
Typically, instead of sampling, a MAP (maximum {\em a-posteriori}) estimate or other heuristics are used since most sampling algorithms are laboriously slow and have convergence guarantees only under limited conditions. The time cost in large part comes from simulating stochastic dynamics of state transitions on deterministic, discrete-logic based hardware, requiring random number generation and fine sampling intervals to avoid discretization errors. These limitations 
have hindered the ability of latent variables models to learn complex structure in data, since adapting the parameters in a more complex, structured model, such as a hierarchical probabilistic model~\cite{lee2003hierarchical}, necessitates sampling under the posterior distribution.

This paper proposes a solution to this problem based on utilizing the intrinsic sources of stochasticity that exist in any physical system.  
Our central thesis is that rather than forcing a deterministic, discrete-logic based system to simulate stochastic dynamics on continuous variables, a more sensible and efficient solution is to exploit physics to directly implement stochastic, analog computation.  In the same way that the analog VLSI retina implements filtering via lateral inhibition in a resistive grid~\cite{mead1988silicon} -- resulting in orders of magnitude greater computational efficiency than digital simulation -- we envision the development of analog circuits that perform the necessary computations and stochastic dynamics for probabilistic inference and learning in complex latent variable models.  A recent successful example of this approach is the work of \cite{borders2019integer}, who used the intrinsic probabilistic behavior of nanoscale magnetic tunneling junctions to sample from the binary state variables of a Boltzmann machine.  Another example is the use of stochastic logic circuits to perform fast Bayesian inference for perception and reasoning tasks \cite{mansinghka2014building,mansinghka2008stochastic}.
Additionally, in neuroscience it has been hypothesized that seemingly random fluctuations in neural activity can be interpreted as a process for sampling from posterior distributions \cite{hoyer2003interpreting,berkes2011spontaneous,orban2016neural,echeveste2020cortical}. 
Our goal here is to demonstrate, through derivation and simulation of a dynamical system of equations, the viability of such an approach for probabilistic inference and learning in a latent variable model.  In an appendix, we point the way to a potential circuit implementation.

Beyond the difficulties associated with sampling, learning the parameters of a probabilistic model requires averaging the samples or other quantities computed from them.  One direct way of doing this is to accumulate these quantities followed by a parameter update 
(Fig. \ref{fig:evo_slsc}). However, this requires a digital accumulator, and the interfacing between analog and digital hardware is often a bottleneck for sampling. For example, in recent work by \cite{roques2019photonic}, the limiting component for a photonic sampler was identified as the photodetector.
Here we propose a novel, fully analog framework in which the update of parameters occurs {\em simultaneously} alongside the sampling of latent variables through continuous time dynamics (Fig. \ref{fig:evo_lsc}). Rather than waiting for the collection of samples for each discrete parameter update, the effective accumulation of samples is achieved by simply having a longer time constant.

\begin{figure}
    \centering
    \subfloat[Discrete Dictionary Update with MAP]{
    \label{fig:evo_dsc}
    \includegraphics[width=0.67\columnwidth]{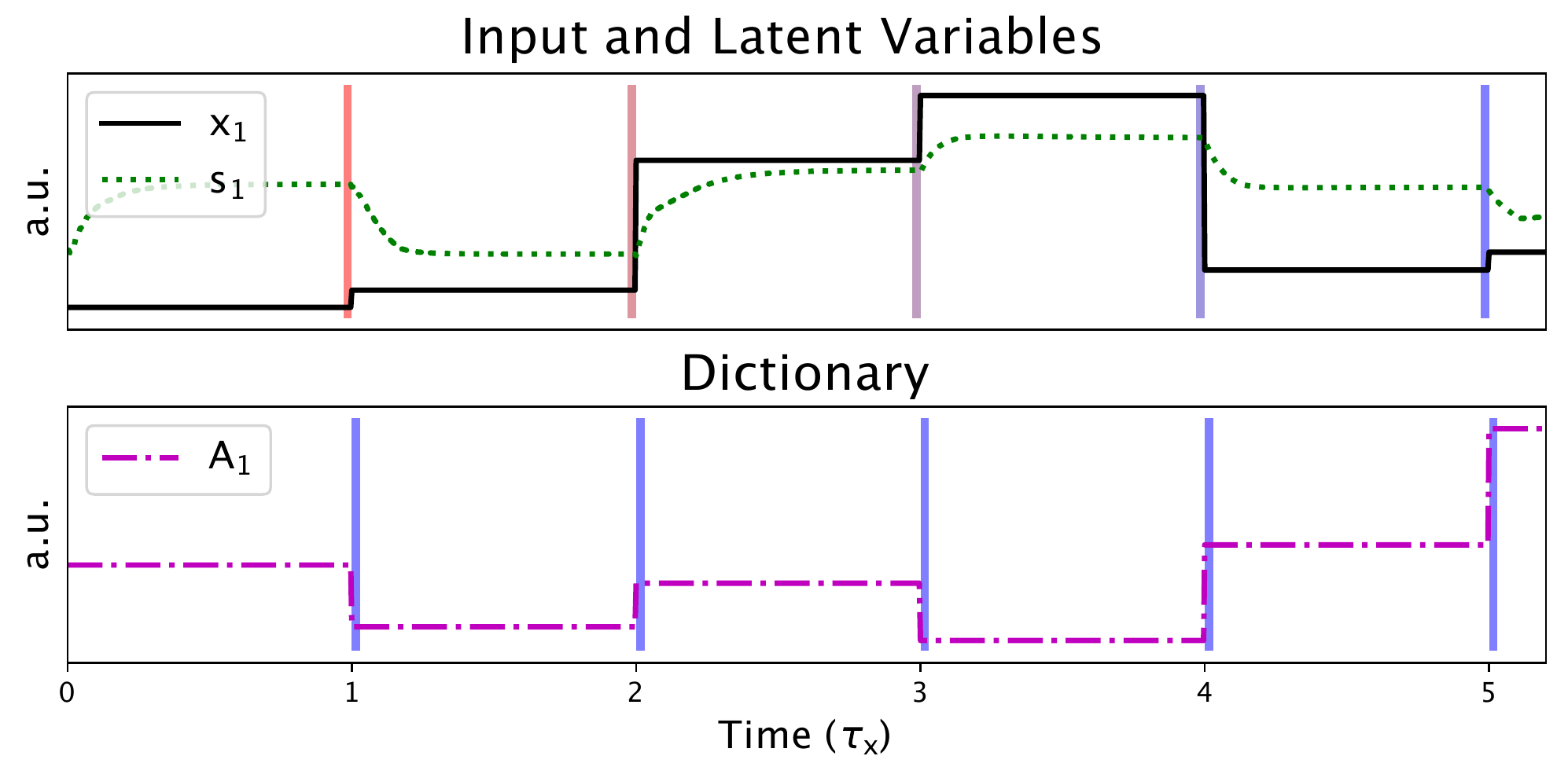}}\\
    \subfloat[Discrete Dictionary Update with Sampling]{
    \label{fig:evo_slsc}
    \includegraphics[width=0.67\columnwidth]{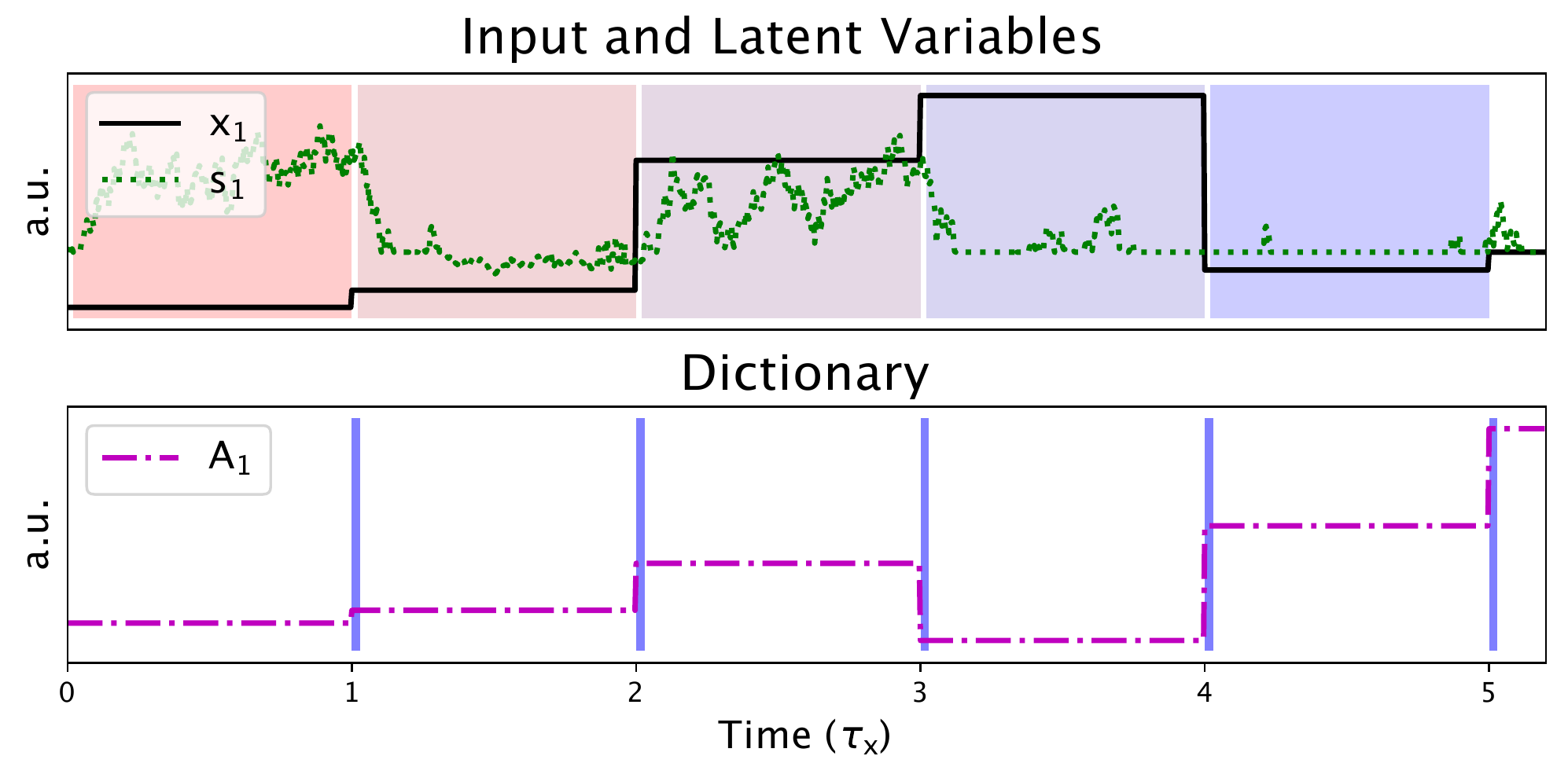}}\\
    \subfloat[Continuous, Simultaneous Dictionary Update with Sampling]{
    \label{fig:evo_lsc}
    \includegraphics[width=0.67\columnwidth]{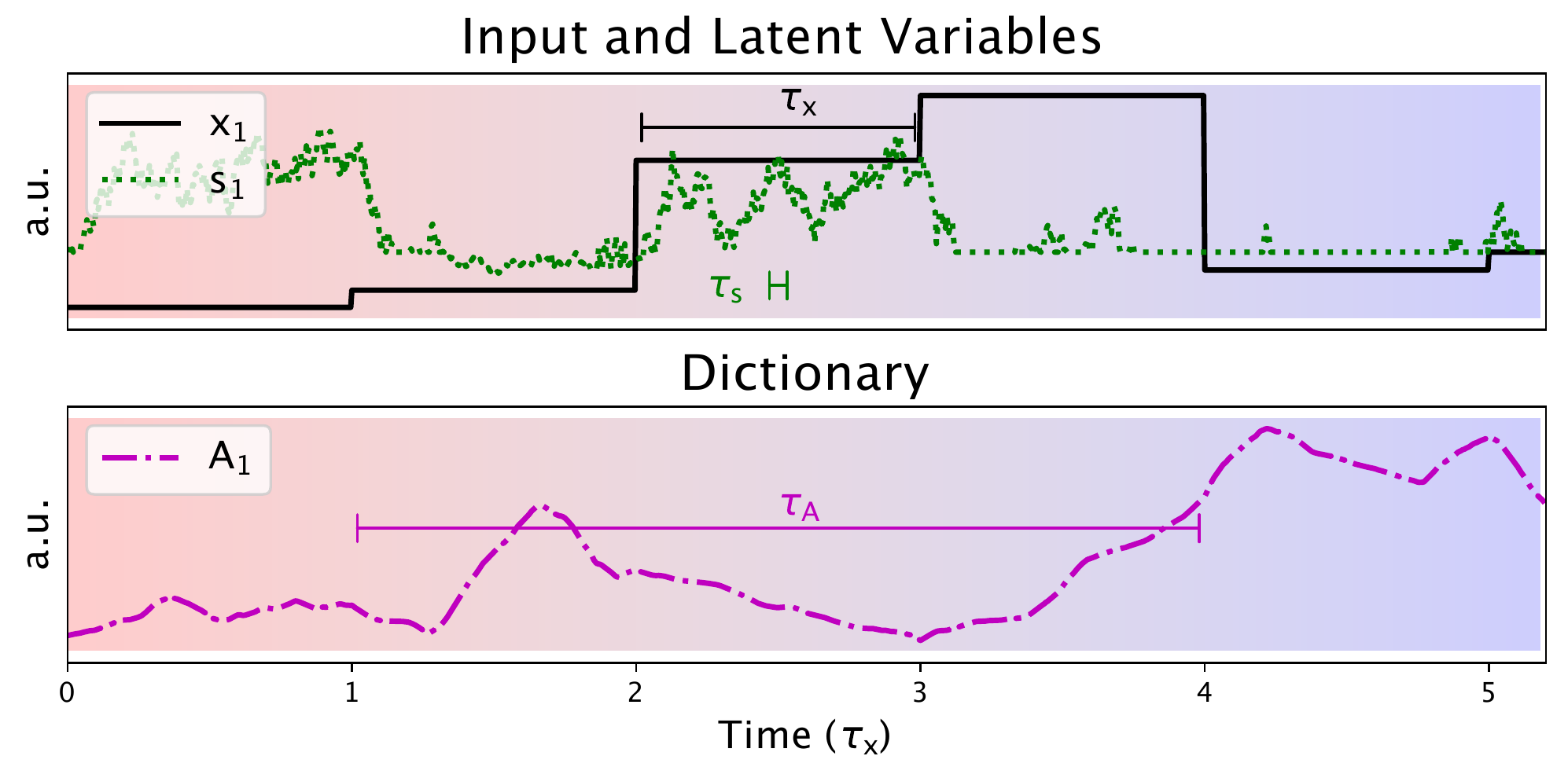}}
    \caption{(caption on following page)}
    \label{fig:evo_illustration}
\end{figure}
\begin{figure}[t] 
  \contcaption{Illustration of three approaches to learning latent variable models. a) In the standard approach, data $\mb x$ is presented at regular intervals (upper plot, black trace). A MAP estimate of latent variables $\mb s$ is calculated via gradient descent or other iterative algorithm (green trace). The resulting estimate is used for a discrete update to the dictionary $A$ (lower plot). The blue vertical bars illustrate the computational inefficiency where only a single point estimate of the coefficients is used to make a dictionary update. b) In a sampling-based approach, for each data interval multiple samples from the posterior are averaged for a dictionary update. The colored regions in the top panel show that many samples are collected to approximate the posterior distribution. However, the discrete dictionary updates (at corresponding vertical bands) make a fully analog implementation difficult.  c) Rather than waiting for the accumulation of samples, the dictionary $A$ is updated simultaneously alongside the latent variables $\mb s$. The slow timescale of the dictionary compared to the latent variables $\tau_A \gg \tau_s$ allows for effective averaging. (Learning rates shown are purely for illustrative purposes.)
  }
\end{figure}




To study this analog learning and inference framework we apply it to the sparse coding model, a simple yet expressive probabilistic model with an explicit prior over the latent variables \cite{tibshirani1996regression,hastie2009elements}.  The sparse coding model is of interest in both neuroscience and engineering as it provides an account for the neural representation of natural images in visual cortex \cite{olshausen1997sparse} and it has proven useful in computer vision~\cite{wright2010sparse,wang2015sparse} and signal compression~\cite{donoho2006compressed}.  However current implementations of sparse coding are slow due to the optimization required to infer the latent variables for each data sample, and learning is inefficient since only a single such point estimate of the latent variables is used to make a dictionary update (Fig.~\ref{fig:evo_dsc}).  In Section~\ref{sec:lsc} we derive a a fully continuous-time sparse coding model by making use of fast Langevin dynamics to sample latent variables and slower dynamics to co-evolve the dictionary based on these samples, as in Figure~\ref{fig:evo_lsc}.

Sampling with Langevin dynamics is well studied both in theory \cite{bussi2007accurate} and in application to Bayesian learning \cite{welling2011bayesian}. However, to our knowledge this is the first fully analog approach to simultaneous inference and learning for sparse coding. Prior sampling-based approaches utilized a mixture-of-Gaussians model and employed discrete Gibbs sampling over the mixture variables \cite{olshausen2000learning} or a method for preselecting parts of the space to sample via MCMC \cite{shelton2011select}. 

An additional advantage of Langevin dynamics is that it leads us to a simple procedure for sampling from the posterior when using an `$L_0$ sparse' prior that explicitly encourages latent variables to be set to zero rather than simply taking on small values (also known as a `spike and slab' prior).  Normally such priors are avoided as finding the optimal sparse representation of a signal requires solving a combinatorial search problem.  Instead, sparsity is enforced by imposing an $L_1$ cost function on the latent variables, which is used as a proxy for $L_0$ since it allows for convex optimization.  However, in probabilistic terms, the $L_1$ cost corresponds to a Laplacian prior which only weakly captures the notion of sparsity.  We show in Section \ref{sec:l0_sparse} how Langevin sparse coding releases us from this restriction.  By simple thresholding of a continuous variable undergoing Langevin dynamics, we obtain samples from the posterior using an `$L_0$ sparse' prior.

In Section \ref{sec:bars}, we demonstrate the efficacy of this model for correct inference and learning using a synthetic dataset.  Furthermore in Section \ref{sec:dict_norm} we demonstrate that this approach allows for learning the size of the dictionary, which was attempted in previous work using variational approximation of the posterior \cite{berkes2008sparsity}.  
Then in Section \ref{sec:vh_results}, we fit our $L_0$-sparse coding model to the Van Hateren dataset of natural images. In addition to learning the dictionary elements, we provide an estimate for the sparsity of natural images. 


To summarize, the main contributions presented are:
\begin{enumerate}
    \item A theoretical formulation of simultaneous dynamics for sampling from latent variables and learning model parameters.
    \item Langevin Sparse Coding (LSC), a continuous-time, probabilistic model for simultaneous inference and learning in a sparse coding model.
    \item An efficient procedure for sampling from the posterior with an `$L_0$ sparse' prior.
    \item Learning not only the dictionary for representing natural images but also other parameters of the model such as the sparsity level and size of the dictionary.
\end{enumerate}

\section{Langevin Sparse Coding}
\label{sec:lsc}
Sparse coding is a simple yet efficient algorithm for learning structure in data by finding a `dictionary' to describe patterns contained in the data. 
While it is formulated as a probabilistic latent-variable model, it is often approximated in practice by finding point estimates for the latent variables rather than sampling from their posterior distribution.  As a result, it is difficult to make rigorous claims about the relation between the learned dictionary and the statistics of the data, and it is problematic to adapt other parameters of the model such as the degree of sparsity or overcompleteness of the dictionary.  More broadly, it has hindered the advancement of sparse coding into a more powerful generative modeling framework -- for example, by incorporating hierarchical structure -- since there is no principled way to learn the parameters of such models without sampling from the posterior.

In this section, we introduce Langevin Sparse Coding (LSC), which efficiently samples the latent-variables of a sparse coding model and allows simultaneous, continuous updates of dictionary elements along with the latent variables. This last property is important in making the LSC framework amenable for fully analog implementation. We begin with a review of the canonical approach of Discrete Sparse Coding (DSC). Next, we introduce simultaneous-update sparse coding (\ctsc) in which dictionary updates are made continuously and concurrent with the dynamics of the coefficients. Finally, we present LSC where we demonstrate that the inherent noise to analog systems can be used to perform sampling.

\subsection{Probabilistic Model}
\label{sec:prob-model}
Sparse coding assumes that the data, $\mb x \in \mathbb R^D$, are described as a linear combination of elements from a dictionary $A \in \mathbb R^{D \times K}$ with additive Gaussian noise $\mb n \in \mathbb R^D$:
\begin{align}
    \mathbf x = A\, \mathbf s + \mathbf n
    \label{eq:x_As_n}
\end{align}
where $n_i \overset{iid}{\sim} N(0, \sigma^2)$. The coefficients $\mb s \in \mathbb R^K$ are latent variables that are assumed to be sparsely distributed, so that any given datapoint should be well approximated using a small number of columns of the dictionary.  Sparsity is enforced by the choice of prior, typically chosen to be factorial:
\begin{eqnarray}
    p_s(\mb{s}) & = & \Pi_{i=1}^K p_s(s_i)\\
    p_s(s_i) & \propto & \exp(-\lambda\, C(s_i))
    \label{eq:sparse_prior}
\end{eqnarray}
where the form of $C$ is chosen so that $p_s(s_i)$ is peaked at $s_i=0$ and with heavy tails away from zero.  (Note that non-factorial priors are also possible, see e.g.,~\cite{garrigues2010group}.)

The posterior over the latent variables in this model may be written in exponential form, 
\begin{equation}
    p(\mb{s\, |\, x}, A) \propto \exp(-E(A, \mathbf s, \mathbf x)),
    \label{eq:posterior}
\end{equation}
with the energy function $E(A, \mathbf s, \mathbf x)$ given by
\begin{align}
    E(A, \mathbf s, \mathbf x) = \frac{||\mathbf x - A\, \mathbf s||_2^2}{2\sigma^2} + \lambda \sum_i C(s_i) \,.
    \label{eq:energy}
\end{align}
Thus inferring a good (highly probable) interpretation of a given data sample, $\mb{x}$, corresponds to finding a set of latent variables, $\mb{s}$, with low energy, $E$.


The goal of learning in this model is to find a dictionary, $A$, that provides the best fit to the data.  This is accomplished by solving for the maximum likelihood estimator (MLE) of the dictionary
\begin{align}
    A^* = \arg\max_A \left \langle \log p(\mathbf x | A) \right \rangle_{\mb x \sim \mathcal D} 
\end{align}
where $\langle \cdot \rangle_{\mb x \sim \mathcal D}$ denotes expectation over the dataset $\mathcal D$ (e.g. natural images). The MLE can be found through gradient ascent, where the gradient is given by
\begin{align}
    \label{eq:dEdA}
    \nabla_A \langle \log p(\mb x | A) \rangle_{\mb x \sim \mathcal D} &= \left\langle \left\langle  -\nabla_A E(A, \mb s, \mb x) \right\rangle_{\mb s | \mb x}\right\rangle_{\mb x \sim \mathcal D}\\
    &= \left\langle \left\langle  (\mb x - A\, \mb s)\, \mb s^T \right\rangle_{\mb s | \mb x}\right\rangle_{\mb x \sim \mathcal D}
    \label{eq:learning-rule}
\end{align}
where $\langle \cdot \rangle_{\mb s | \mb x}$ denotes expectation with respect to the posterior distribution $p(\mb{s\, |\, x}, A)$
(see~\cite{lewicki1999probabilistic} for a derivation).  Thus, adapting the dictionary to the data requires, for each data sample $\mb x$, sampling from the posterior over $\mb s$ and computing the correlation between the residual, $\mb{x}-A\,\mb{s}$, and $\mb s$.  The dictionary $A$ would then be incrementally updated according to this correlation (eq.~\ref{eq:learning-rule}).  Equilibrium is reached when $\left\langle\langle \mb{\hat{x}}(\mb s)\mb{s}^T \rangle_{\mb s | \mb x}\right\rangle_{\mb x \sim \mathcal D} = \left\langle \mb x \langle \mb s^T \rangle_{\mb s | \mb x}\right\rangle_{\mb x \sim \mathcal D}$, with $\mb{\hat{x}}(\mb s)=A\,\mb s$.

Beyond learning the dictionary, one can adapt other parameters of the model such as $\sigma$ and $\lambda$ also via gradient descent.  The gradients for these parameters are as follows:
\begin{align}
    \label{eq:dEdsigma}
    \nabla_{\sigma} \langle \log p(\mb x | A) \rangle_{\mb x \sim \mathcal D} &\propto \frac{1}{D}\left\langle \left\langle  |\mb x - A\, \mb s|^2 \right\rangle_{\mb s | \mb x}\right\rangle_{\mb x \sim \mathcal D} - \sigma^2\\
    \nabla_{\lambda} \langle \log p(\mb x | A) \rangle_{\mb x \sim \mathcal D} &\propto \frac{1}{K}\left\langle \left\langle  \sum_i^K C(s_i) \right\rangle_{\mb s | \mb x}\right\rangle_{\mb x \sim \mathcal D} - \langle C(s) \rangle_{p_s(s)}
    \label{eq:dEdlambda}
\end{align}
Adapting these parameters similarly requires computing averages under the posterior distribution for each data sample. Note that when the sparse coding model objective is formulated purely in terms of its energy function (eq.~\ref{eq:energy}) -- which is typically the case -- then there is no principled away to adapt these parameters to the data.  The probabilistic framework makes it possible, so long as it is tractable to sample from the posterior distribution.

\subsection{Discrete Sparse Coding}
\label{sec:dsc}

In practice, the expectation over the data in (\ref{eq:learning-rule}) is approximated via stochastic gradient descent (SGD). For a batch of data of size $N$, $\{\mb x_n\}_{n=1\dots N}$, the update rule is
\begin{align}
    \Delta A = \eta\, \frac1N \sum_{n=1}^N \left\langle (\mb x_n - A\, \mb s_n)\, \mb s_n^T\right\rangle_{\mb s_n | \mb x_n}
    \label{eq:step_A}
\end{align}
where $\eta$ specifies the learning rate.  However, the expectation over $\mb s_n$ is usually considered intractable and so in practice it is approximated by the maximum \emph{a posteriori} (MAP) estimator of $\mb s_n$
\begin{align}
    \mb s_n^* = \arg \min_{\mb s_n} E(A, \mb s_n, \mb x_n).
\end{align}
Solving via gradient descent yields the iterative update equation
\begin{align}
\Delta \mb s_n & \propto -\nabla_{\mb s} E(A, \mb s_n, \mb x_n)\\
&= -\frac1{\sigma^2}A^T(\mb x_n - A \mb s_n) - \lambda\,C^\prime(\mb s_n)
\label{eq:step_s}
\end{align}
where $C^\prime$ is the derivative of cost function $C$ above (\ref{eq:energy}) and operates elementwise on $\mb s_n$.  For each $\mb x_n$, equation (\ref{eq:step_s}) is iteratively evaluated until it converges to a solution.  
In order to make this a convex optimization, the cost function $C$ is typically taken to be the $L_1$ norm, corresponding to a Laplacian prior $p_s(\mb s)$.
Gradient descent does not generally constitute the most efficient method for finding the MAP estimate, but we use it here as a step towards the development of LSC below.

The price we pay for approximating the expectation $\langle\,\cdot\,\rangle_{\mb s_n | \mb x_n}$ in equation~\ref{eq:step_A} with a single MAP estimate is that it now becomes necessary to normalize the dictionary elements $A = (\mb A_1, \dots, \mb A_K)$ after each update via
\begin{align}
    \mb A_i \leftarrow \frac{\mb A_i}{||\mb A_i||_2} \equiv {\mb {\hat A}_i}.
    \label{eq:normalization}
\end{align}
This is necessary because the MAP estimator $\mb s^*$ will consistently underestimate $\mb s$ such that it is biased toward zero (due to the sparse prior).  As a result, each $\mb A_i$ will grow without bound unless normalized. 
(As we shall see below, this no longer becomes necessary when we sample from the posterior.)

Both updates $\Delta A$ and $\Delta \mb s_n$ can be expressed more efficiently through gradient descent on a batch energy function:
\begin{align}
    E (A, S, X) &\equiv \sum_{n=1}^N E(A, \mb s_n, \mb x_n)\\
    &= \frac{||AS - X||_{2,2}^2}{2\sigma^2} + \lambda ||S||_{1, 1}.
\end{align}
We have defined batch matrices $S \in \mathbb R^{K \times N}$ and $X \in \mathbb R^{D \times N}$. 
Above, $|| \cdot ||_{p, q}$ refer to the $L_{(p, q)}$ matrix norm, defined by
\begin{align}
    ||A||_{p, q} = \left(\sum_j \left(\sum_i |a_{ij}|^p \right)^{\frac q p} \right)^{\frac 1q}
\end{align}
With the batch energy defined, the update rules are
\begin{align}
    S &\leftarrow S -\eta_S \nabla_S E(A, S, X)
    \label{eq:dsc_update}\\
    A &\leftarrow A -\eta_A \nabla_A E(A, S, X)
    \label{eq:dsc_update_2}\\
    A &\leftarrow \mbox{\rm Norm}(A)
\end{align}
where the Norm() operation corresponds to the normalization of equation~\ref{eq:normalization}.

To coordinate the updates of $S$ and $A$, a nested loop must be used (Alg. \ref{alg:SC_MAP}). The inner loop approximates the MAP estimator $S^*$ while the outer loop finds the MLE of $A$.

\begin{algorithm}
\begin{algorithmic}[1]
\For{$k \leftarrow 1$ to $N_A$}
    \State $X \leftarrow \Call{SampleBatch}$
    \For{$n \leftarrow 1$ to $N_s$}
        \State $S \leftarrow S - \eta_S \cdot \nabla_S E (A, S, X)$
        \label{alg:SC_MAP_innerloop}
    \EndFor
    \State $S^* \leftarrow S$ 
    \State $A \leftarrow A - \eta_A \cdot \nabla_A E (A, S^*, X)$ 
    \State $A \leftarrow \mbox{\rm Norm}(A)$
\EndFor
\end{algorithmic}
\caption{Algorithm for discrete sparse coding (DSC). Note line 6 was included purely to emphasize $S^*$ as a MAP estimate.}
\label{alg:SC_MAP}
\end{algorithm}

A closely related cousin of DSC, the Locally Competitive Algorithm (LCA)~\cite{rozell2008sparse}, computes the MAP estimate by following dynamics that descend the energy $E$ in a more efficient manner.  Instead of doing direct gradient-descent (eq.~\ref{eq:step_s}), $\mb s$ is taken to be a monotonically increasing, nonlinear function of another variable $\mb u$ that follows the gradient with respect to $\mb s$:
\begin{align}
\Delta \mb u_n & \propto -\nabla_{\mb s} E(A, \mb s_n, \mb x_n)\\
\mb s_n & = g(\mb u_n)
\end{align}
where $g$ operates elementwise on $\mb u$ and is determined by the choice of cost function $C$.  For an L1 cost, $g$ is a signed Relu function with threshold $\lambda$:
\begin{align}
    g(u_i) = 
    \begin{cases}
    0 & |u_i| < \lambda\\
    \sign(u_i)(|u_i| - \lambda) & |u_i| \ge u_0
    \end{cases}
    \label{eq:threshold_fn_lca}
\end{align}
Other than this difference in the dynamics for MAP inference, which falls purely within the inner loop (line~\ref{alg:SC_MAP_innerloop}) of Algorithm~\ref{alg:SC_MAP}, both DSC and LCA update the dictionary based on a single MAP estimate and thus suffer the same inefficiency as depicted in Figure~\ref{fig:evo_dsc}.

\subsection{Simultaneous (Update) Sparse Coding - SSC}

We note that the DSC algorithm above requires the alternating update of the dictionary elements and coefficients. Typically, this necessitates a digital clock for synchronization and is a major challenge towards fully analog implementation. In this subsection, we present an asychronous framework -- Simultaneous-Update Sparse Coding (\ctsc{}) -- where both the dictionary and coefficients are updated simultaneously. 

Rather than updating the dictionary $A$ at the end of the loop when $S$ has converged to the MAP estimator $S^*$, \ctsc{} updates $A$ continuously and concurrent with $S$. In search of dynamics amenable to analog computation, we take the step sizes to be infinitesimally small, and arrive at the following set of differential equations.
\begin{align}
    \label{eq:ctsc_diffeq}
    \tau_S \dot S &= -\nabla_S E(A, S, X(t))\\
    \label{eq:ctsc_diffeq_2}
    \tau_A \dot A &= -\nabla_A E(A, S, X(t))
\end{align}
while still enforcing the normalization constraint on $A$ (eq.~\ref{eq:normalization}).
Here, we take $X(t)$ to be updated synchronously at regular intervals of $\tau_X$. At each update, a new batch of samples is drawn.

To compare \ctsc and DSC, consider the following simulation for \ctsc using the Euler Method.
\\
\begin{algorithm}[ht!]
\begin{algorithmic}[1]
\For{$t \leftarrow 1$ to $t_\text{max}/\Delta t$}
    \State $dS \leftarrow \frac{\partial E}{\partial S} (A, S, X(t))$
    \State $dA \leftarrow \frac{\partial E}{\partial A} (A, S, X(t))$
    \State $S \leftarrow S - \frac{\Delta t}{\tau_S}\cdot dS$
    \State $A \leftarrow A - \frac{\Delta t}{\tau_A} \cdot dA$
    \State $A \leftarrow \mbox{\rm Norm}(A)$
\EndFor
\end{algorithmic}
\caption{Euler Method simulation of \ctsc with stepsize of $\Delta t$ and regular interval input of $X$}
\label{alg:CTSC}
\end{algorithm}

Comparing Algorithm \ref{alg:SC_MAP} and Algorithm \ref{alg:CTSC}, the timescales $\tau$ can be related to the learning rates, $\eta$, and the number of iterations $N_S$. We stress an important difference between the two is that \ctsc{} is fully described through a set of coupled differential equations and requires no control structure (i.e. a nested for loop). This is especially desirable for analog implementation as a global clock is no longer necessary. Furthermore, there is no longer need for synchronous, regular input of the data $X$. While not explored here, dynamic input such as videos can be naturally processed without any frame-by-frame synchronization.

\subsection{Sampling via Langevin Dynamics}

Consider a time-varying system described by coordinates $\mathbf u (t)$ with energy $E(\mathbf u)$ . It can be modeled by Langevin dynamics according to the following stochastic differential equation:
\begin{align}
    \dot {\mb u} = - \nabla E(\mb u) + \sqrt{2 T} \xi(t),
    \label{eq:langevin}
\end{align}
where $\xi(t)$ is independent Gaussian white noise with $\langle \xi(t) \xi(t')^T \rangle = \mb I \delta(t - t')$.  Under these dynamics the distribution of $p(\mb u(t))$, over time, will asymptotically converge to
\begin{align}
    p^{(\infty)}(\mb u) \propto e^{-E(\mb u) / T}
\end{align}

This relation suggests that we change the dynamics of \ctsc (\ref{eq:ctsc_diffeq}) by injecting noise to $\dot S$:
\begin{align}
    \tau_S \dot S &= -\nabla_S E(A, S, X) + \sqrt{2 T \tau_S}\xi(t)
    \label{eq:lsc_update_s}
\end{align}
Note that under the scaling of $t \rightarrow t / \tau_S$, we have $\langle \xi(t/\tau_S) \xi(t'/\tau_S)^T \rangle = \mb I \delta(\tau_S^{-1}(t - t')) = \tau_S \mb I \delta(t - t') = \langle \sqrt{\tau_S} \xi(t) \sqrt{\tau_S} \xi(t')^T \rangle$. This necessitates the somewhat unexpected scaling factor of $\tau_S$.

Following the above dynamics, for fixed $A$ and input $X$, $S$ will sample from the posterior distribution,
\begin{align}
    p_{S|X}(S(t)|X,A) \propto e^{-E(A, S, X)/T}.
    \label{eq:eq_distr}
\end{align}
This is a remarkable result:  {\em By simply injecting noise into the continuous-time dynamics normally used for MAP inference in sparse coding, we obtain a dynamical system that naturally samples from the desired posterior distribution (eq.~\ref{eq:posterior})}.  With $T=0$, we recover the \ctsc dynamics above (eqs.~\ref{eq:ctsc_diffeq}-\ref{eq:ctsc_diffeq_2}) where  $S$ converges to the MAP estimate. 

A useful property of (\ref{eq:lsc_update_s}) is that the equilibrium distribution is independent of the time constant $\tau_S$. By taking $\tau_A \gg \tau_S$, the assumption that $A$ is fixed with respect to the dynamics of $S$ can be upheld. Conversely, because $S$ evolves much faster than $A$, the dynamics of $A$ are well approximated by
\begin{align}
    \tau_A \dot A = - \langle \nabla_A E(A, S, X) \rangle_{S|A, X}.
    \label{eq:lsc_update_A}
\end{align}
This is the exact mean gradient that we originally sought to calculate (eq.~\ref{eq:dEdA}).

In summary, we have derived a new method for inference and learning in a sparse coding model, Langevin Sparse Coding (LSC), as specified by the continuous, coupled dynamics of equations~\ref{eq:lsc_update_s} and \ref{eq:lsc_update_A}, that achieves the desired property illustrated in Figure~\ref{fig:evo_lsc}.  Importantly, our aim doing this is not simply to produce another MCMC algorithm, but rather to move toward a physical realization that naturally implements these dynamics (an example of which is described in Appendix~\ref{sec:hardware}).  

\section{`$L_0$ Sparse' Prior}
\label{sec:l0_sparse}
Since the goal of sparse coding is to represent each data item using a small number of non-zero latent variables, the prior should ideally have a sharp peak at zero in order to encourage many latent variables to be set to zero.  In this case, the cost term $C$ within the energy function (\ref{eq:energy}) would resemble an $L_0$ cost that rewards coefficients for being strictly zero (as opposed to being non-zero and merely small in amplitude).  However such cost functions are not used in practice because they are not amenable to gradient-based or convex optimization methods for computing the MAP estimate.  Instead, the $L_1$ cost is usually adopted as a proxy for $L_0$ as it has been shown to yield equivalent solutions under certain conditions~\cite{tropp2006algorithms}.  However from the perspective of a probabilistic model, the $L_1$ cost corresponds to a Laplacian prior that only weakly expresses the notion of sparsity.  In fact, the Laplacian is the {\em maximum entropy} distribution for a real-valued variable of a given mean absolute value.  Here we show that the use of `$L_0$ Sparse' priors becomes tractable in our sampling-based setting, and we develop a modified LSC formulation that enables efficient sampling from the posterior.

\begin{figure}
    \centering
    \includegraphics[width=.9\columnwidth]{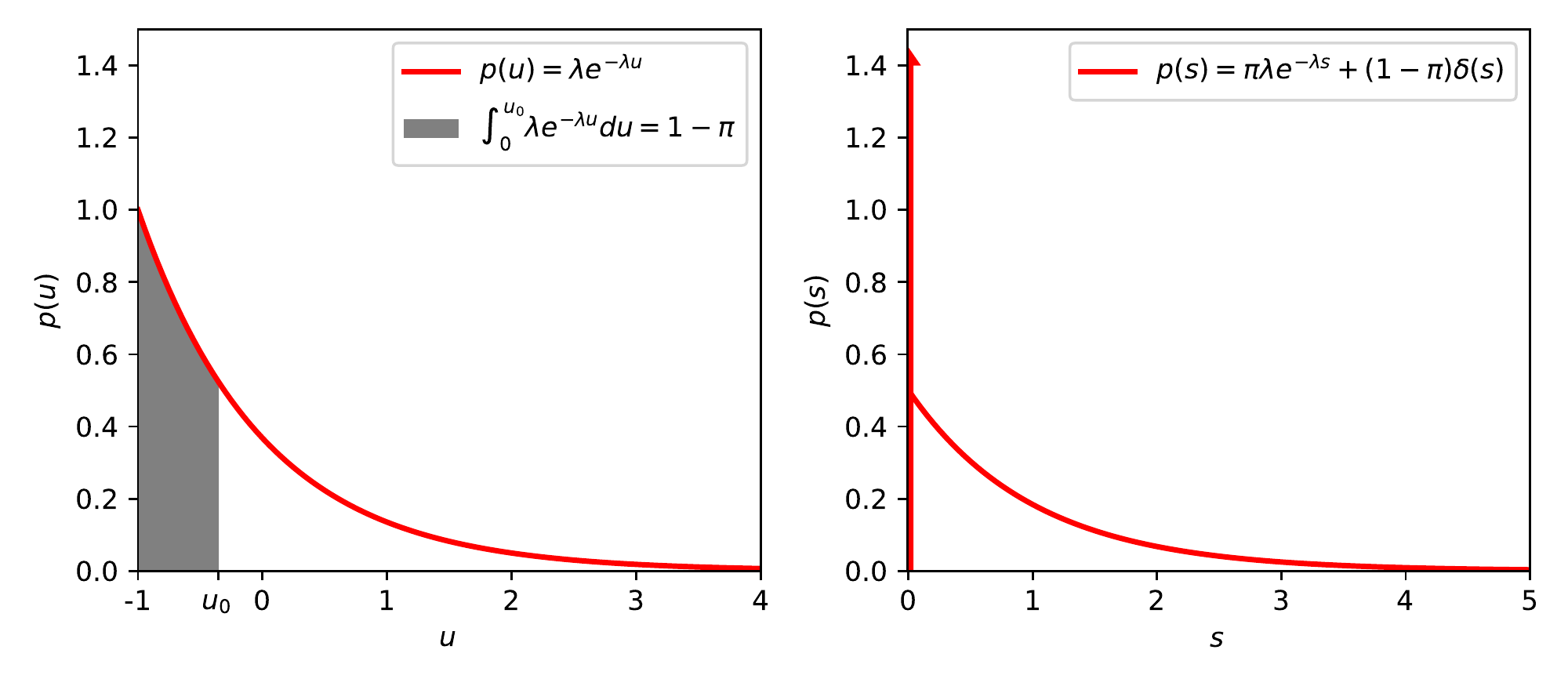}
    \caption{`$L_0$ Sparse' prior.  Left panel shows the exponential distribution $p(u)$. With the change of variable $s=f(u)$ via the application of a soft-thresholding function, we obtain the desired L0-like distribution $p_0(s)$ shown in the right panel (shown for the region $s\ge0$). The threshold parameter $u_0$ is chosen so that the probability weight of the delta function, $1-\pi$, is equal to the shaded region in the left panel.  These plots show the resulting distributions for $\lambda = 1, \pi = 0.5$.}
    \label{fig:L0_prior}
\end{figure}

Consider the following prior consisting of a mixture of a delta-function and Laplacian distribution (also known as a `spike and slab' prior \cite{mitchell1988bayesian}):
\begin{align}
    p_0(s) = \pi\,\lambda e^{-\lambda\, s} + (1- \pi)\delta(s).
    \label{eq:l0_prior}
\end{align}
With $\pi$ as the probability of being `active', $1-\pi$ quantifies the $L_0$ sparsity, or how likely $s$ is to be zero. When $s$ is in the active state it is exponentially distributed with mean $1/\lambda$ (see right panel of figure~\ref{fig:L0_prior}).  
Note that here and in what follows we will assume the latent variables to be non-negative as opposed to allowing them to go positive or negative as is typically the case in sparse coding models.

To develop an efficient sampling strategy, we first define auxiliary variables $\mb u$ such that each $u_i$ independently follows an exponential distribution:
\begin{align}
    p_U(u_i) = \lambda\, e^{-\lambda\,u_i}.
\end{align}
We then take the latent variables $\mb s$ to be given by $s_i = f(u_i)$ where $f$ is a biased ReLU function:
\begin{align}
    s_i = f(u_i) = 
    \begin{cases}
    0 & u_i < u_0\\
    u_i - u_0 & u_i \ge u_0
    \end{cases}
    \label{eq:threshold_fn}
\end{align}
for some positive $u_0$. We can show that $s_i$ is then distributed according to the prior $p_0(s)$ by marginalizing the joint distribution $p(s,u)$ over $u$ as follows:
\begin{equation}
\begin{aligned}
p_S(s) &= \int_{-\infty}^\infty p(s|u)\,p_U(u) du\\
 &= 
    \int_0^{u_0} \delta(s)\,p_U(u) du + \int_{u_0}^\infty \delta(s-(u-u_0))\,p_U(u) du \\
  &=\delta(s)\, \int_0^{u_0} p_U(u) du+ p_U(s+u_0) \\
  &=\delta(s)\, [1-e^{-\lambda u_0}] + \lambda e^{-\lambda s}\,
  e^{-\lambda u_0} \\
  &=[1-\pi]\, \delta(s) + \pi\, \lambda e^{-\lambda s}
  \; \equiv \; p_0(s)
\end{aligned}
\end{equation}
with $\pi=e^{-\lambda u_0}$.  The relation between $p(u)$, $u_0$ and $p(s)$ is illustrated in Figure~\ref{fig:L0_prior}.


To derive the Langevin dynamics for sampling from the posterior using the $L_0$-sparse prior above, we first re-write the energy function in terms of $\mb u$:
\begin{align}
    E(A, \mb{u}, \mb{x}) = \frac12 \frac{||\mb x - A f(|\mb u|)||_2^2}{\sigma^2} + \lambda ||\mb u||_1.
    \label{eq:energy_l0}
\end{align}
We then let $\mb u$ follow Langevin dynamics governed by this energy function.  Note that we can allow the $u_i$ to move freely between positive and negative values and then use only their absolute value in evaluating the energy.  This essentially reflects the dynamics about the origin which avoids the problems associated with having an infinite energy barrier at $u_i=0$.
Letting $|\mb u|$ denote the elementwise absolute value of $\mb u$, the distribution of $|\mb u|$ will converge to
\begin{align}
    p(|\mb{u}|\, |\, \mb{x}) &\propto \exp\left(-||A\, f(|\mb u|) - \mb x||_2^2/\sigma^2-\lambda ||\mb u||_1\right) \\
    & \propto p(\mb x | f(|\mb u|))\, p_U(|\mb u|)\\
    &=p(\mb x | \mb s)\, p_0(\mb s) 
\end{align}
Thus we obtain a second remarkable result:  {\em By following Langevin dynamics on the energy in (\ref{eq:energy_l0}) with $\mb s=f(|\mb u|)$, we obtain samples from the posterior $p(\mb{s}|\mb{x})$ given by combining the likelihood with the $L_0$-sparse prior $p_0(\mb s)$}.  This is significant, because a MAP-estimate based approach would be impossible with such a prior since the posterior will always have its maximum at $\mb s = 0$ regardless of the likelihood. 

Applying the LSC equations (\ref{eq:lsc_update_s}, \ref{eq:lsc_update_A}) using the energy in equation (\ref{eq:energy_l0}), we obtain the following coupled stochastic differential equations for inference and learning in $L_0$-LSC:
\begin{align}
    \label{eq:l0lsc_1}
    \tau_u \dot{\mb u} &= - A^T(A\, \mb s - \mb x)\Theta(|\mb u| - \mb{u_0}) - \lambda\,\sign(\mb u) + \sqrt 2 \xi(t)\\
    \mb s & = f(|\mb u|)\\
    \label{eq:l0lsc_2}
    \tau_A \dot A &= -(A\, \mb s - \mb x) \mb s^T.
\end{align}
where $\Theta(u)$ is the Heaviside function and $\xi(t)$ is independent Gaussian white noise.
Importantly, we can also learn $u_0$, and therefore the activation probability, $\pi$, via the dynamics
\begin{align}
    \label{eq:u0_update}
    \dot u_0 &\propto \left\langle \left\langle - \frac{\partial E}{\partial u_0} \right\rangle_{\mb s| \mb x} \right \rangle_{X\sim \mathcal{D}}\\
    &= \left\langle \left\langle A^T(A \mb s - \mb x) \cdot \mathbf 1(\mb s > 0) \right\rangle_{\mb s | \mb x} \right \rangle_{\mb x \sim \mathcal D}
\end{align}

\section{Results}
To study the efficacy of $L_0$-LSC, we first apply it to an artificial dataset consisting of images of bars in different orientations.  This provides a useful test case for evaluation since the causes that generate the data are known.  We then turn to a dataset of natural scenes where the ground truth is unknown.

\subsection{Inference on Bars Dataset}
\label{sec:bars}
                
\begin{figure}
    \centering
    \subfloat[Bars Dictionary]{
    \label{fig:bars_dict}
    \includegraphics[width=.6\columnwidth]{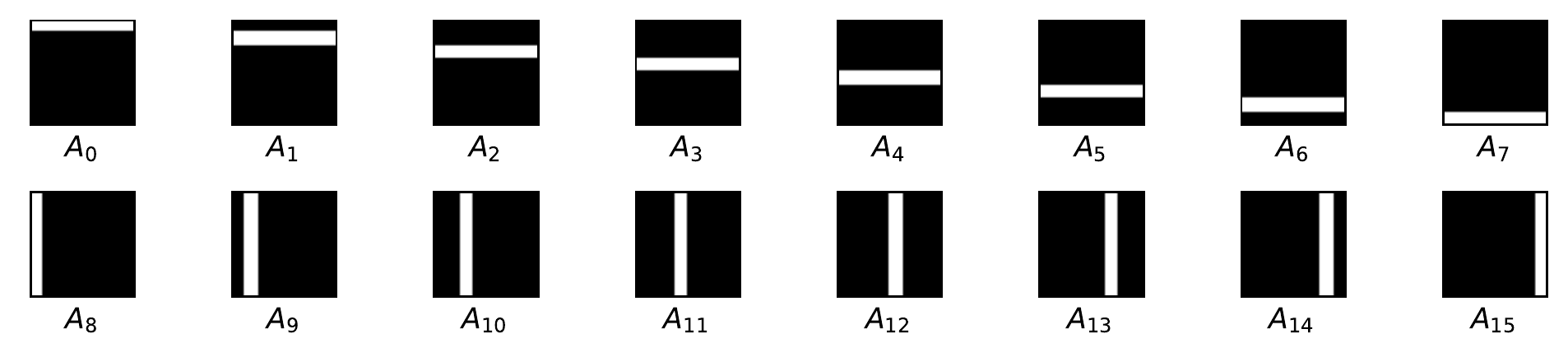}}\\
    \subfloat[Bars Sample: $\lambda = 1, \pi=0.3, \sigma=0$]{
    \label{fig:bars_samp}
    \includegraphics[width=.6\columnwidth]{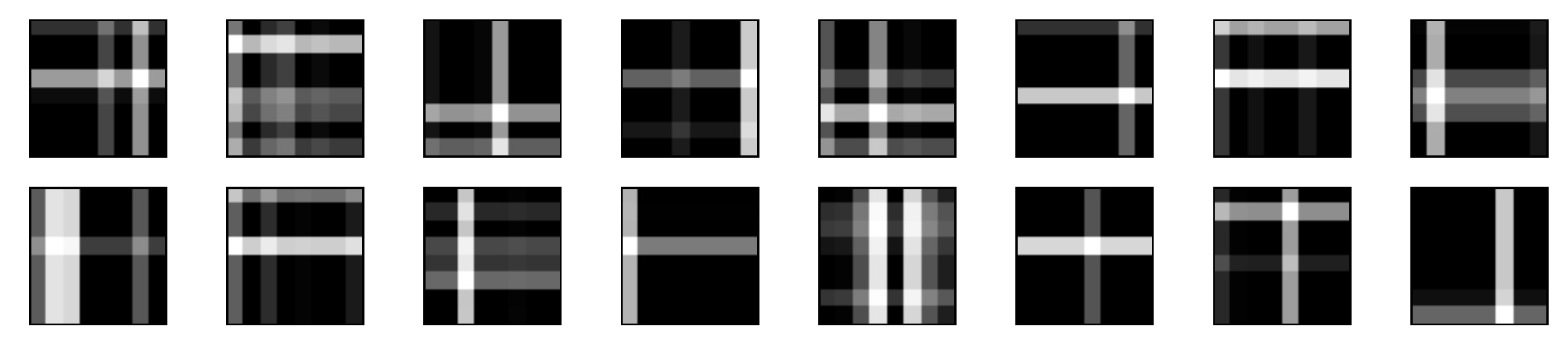}}\\
    \subfloat[Bars Sample: $\lambda = 1, \pi=0.3, \sigma=0.5$]{
    \label{fig:bars_samp_noise}
    \includegraphics[width=.6\columnwidth]{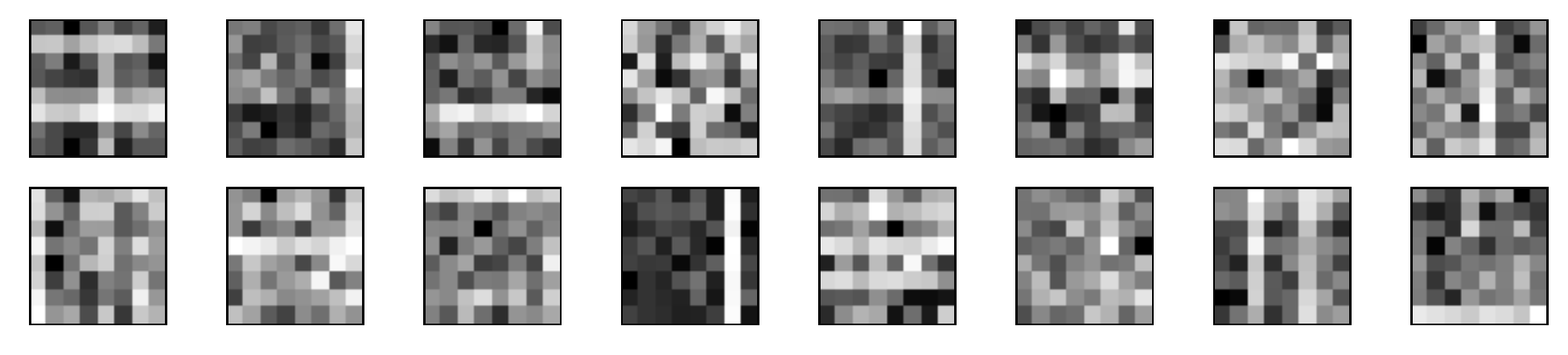}}
    \caption{The synthetic Bars dataset used as a toy problem. a) The dictionary is the collection of vertical and horizontal lines. b) An example of a sample drawn from the dataset. c) Another sample with noise introduced.}
    \label{fig:bars}
\end{figure}

For the bars dataset, samples are generated from a dictionary $A$ consisting of vertical and horizontal lines (Fig. \ref{fig:bars_dict}). We compare results obtained on this dataset against DSC as well as another method for training sparse coding, the locally competitive algorithm (LCA) \cite{rozell2008sparse}.

We synthetically generate data as a linear combination of the dictionary with additive Gaussian noise (Eq. \ref{eq:x_As_n})
where, $n_i \sim N(0, \sigma^2)$ and the coefficients are distributed according to $L_0$ zero-inflated exponential prior\cite{beckett2014zero} (Eq. \ref{eq:l0_prior}). A sample drawn from this model without noise and with noise is shown in Fig. \ref{fig:bars_samp} and \ref{fig:bars_samp_noise}.

\begin{figure}[h]
    \centering
    \subfloat{
    \label{fig:lambda_vs_pi}
    \includegraphics[width=.45\columnwidth]{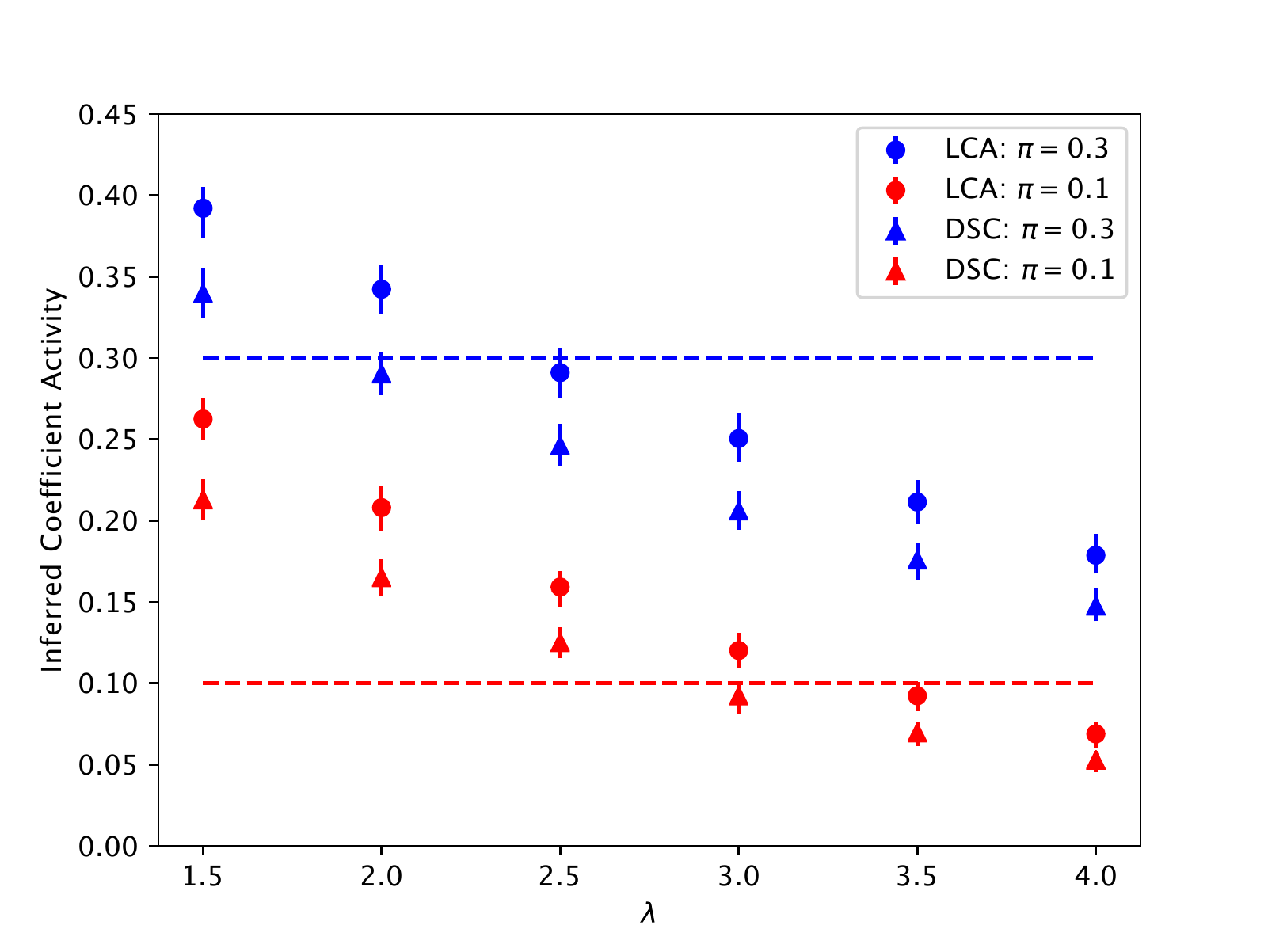}
    }
    \subfloat{
    \label{fig:learning_pi}
    \includegraphics[width=.45\columnwidth]{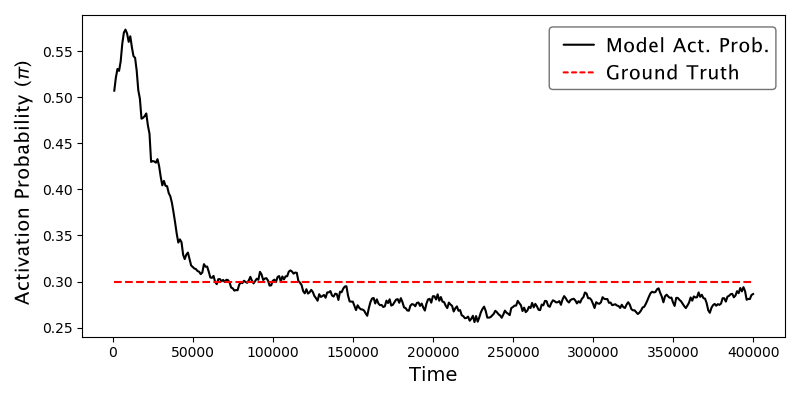}
    }
    \caption{a) LCA and DSC are trained on data generated with activation probability $\pi =0.3$ (blue) and $\pi=0.1$ (red). For both, a sweep in sparsity parameter $\lambda$ is made. While a correspondence between $\lambda$ and $\pi$ exists, there is no analytic expression to automatically adapt these parameters to the data. Even with data of known sparsity, it is impossible to select the correct parameter $\lambda$ to use. b) With L0-sparse LSC, the activation probability $\pi$ is directly related to the parameter $u_0 = -\lambda^{-1} \log \pi$ and can be learned directly without a parameter search.} 
\end{figure}

When trained on this dataset, all three algorithms were successful at learning the correct dictionary. However, $L_0$-LSC can better capture the posterior distribution than either DSC or LCA due to the fact that it directly enforces $L_0$ sparsity.  In both DSC and LCA, the sparsity is controlled by adjusting the parameter $\lambda$. However, the relationship between $\lambda$ and $L_0$ sparsity (Fig. \ref{fig:lambda_vs_pi}) is rather indirect and no analytic expression is known. On the other hand, in $L_0$-LSC a specific level of $L_0$-sparsity can be directly enforced by setting $u_0 = -\lambda^{-1} \log(\pi)$.

Moreover, the activation probability $\pi$ can be learned by LSC without any guesswork or parameter search (Eq. \ref{eq:u0_update}). Specifically, simultaneous to the evolution of $A, \mb u$, the threshold parameter $u_0$ is treated as a variable evolves through gradient descent, $\dot u_0 \propto \nabla_{u_0} E$.

\begin{figure}[ht!]
    \centering
    \subfloat[DSC]{
    \label{fig:distr_dsc}
    \includegraphics[width=.45\columnwidth]{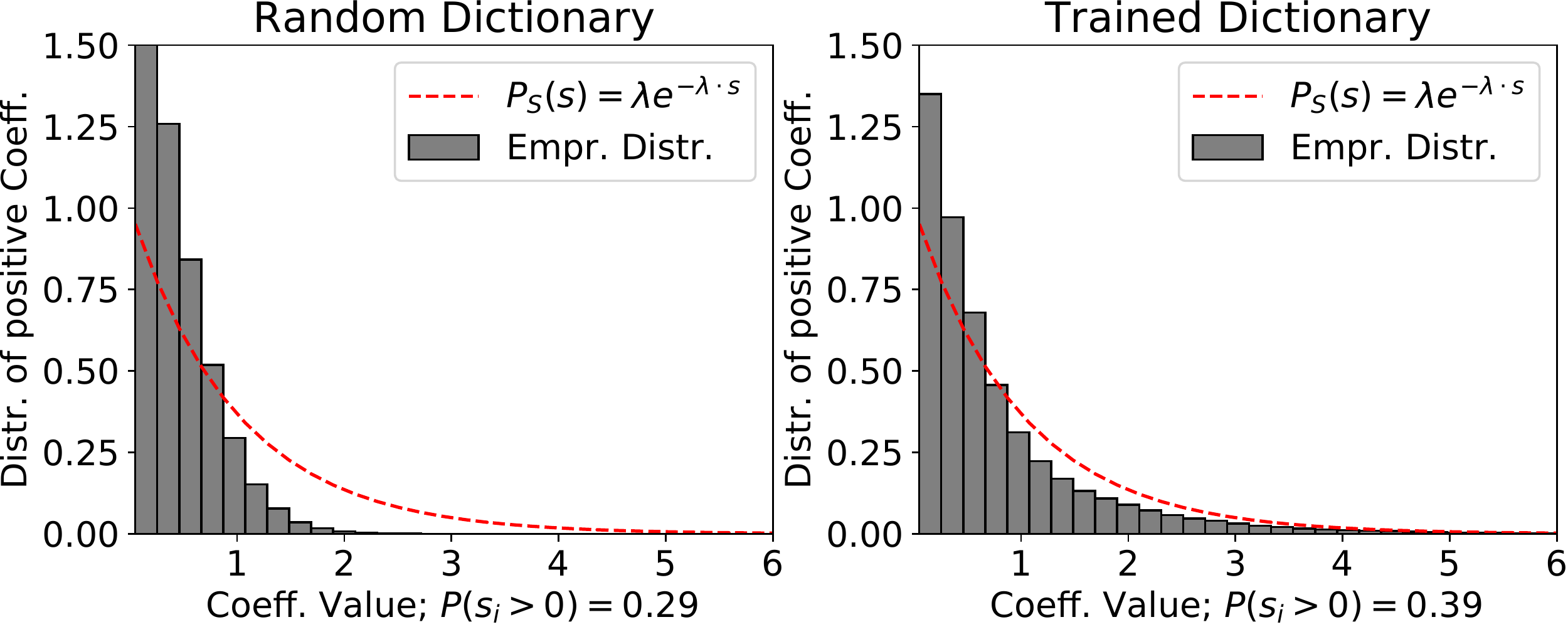}}
    \subfloat[LCA]{
    \label{fig:distr_lca}
    \includegraphics[width=.45\columnwidth]{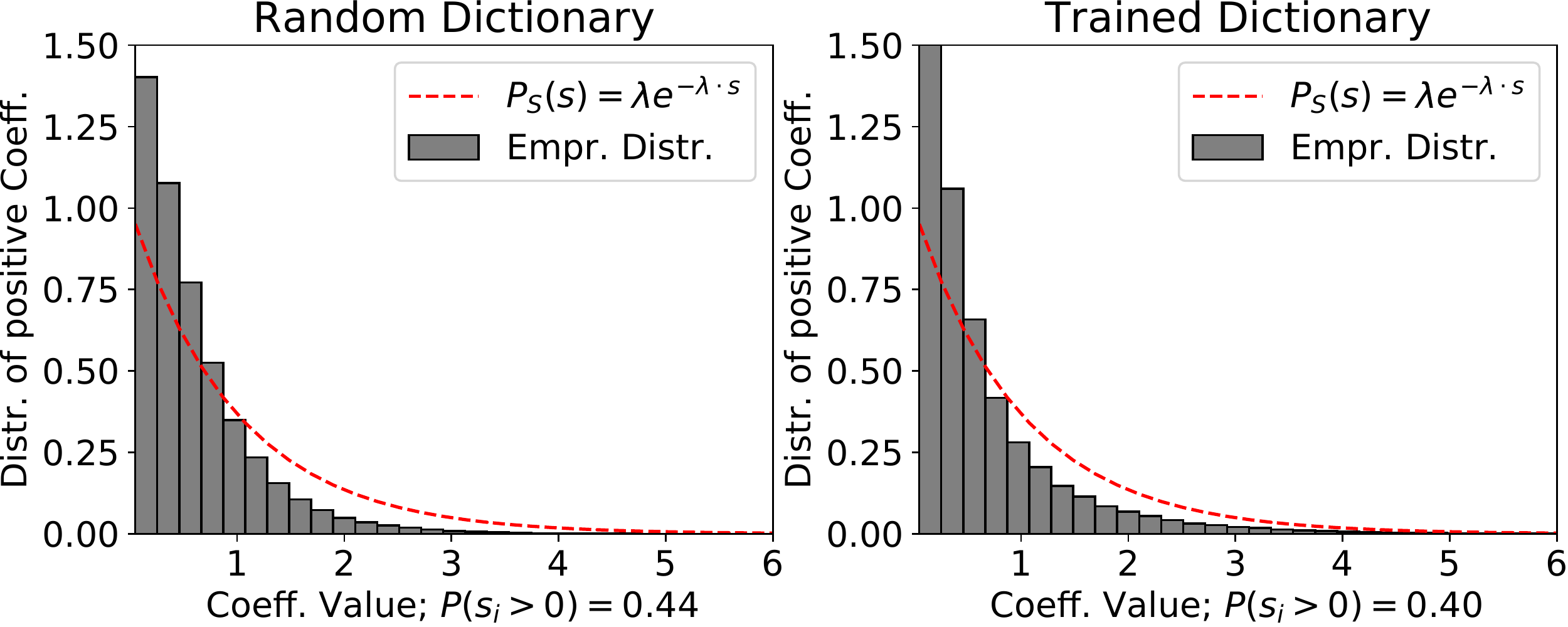}}\\
    \subfloat[$L_0$-LSC]{
    \label{fig:distr_lsc}
    \includegraphics[width=.45\columnwidth]{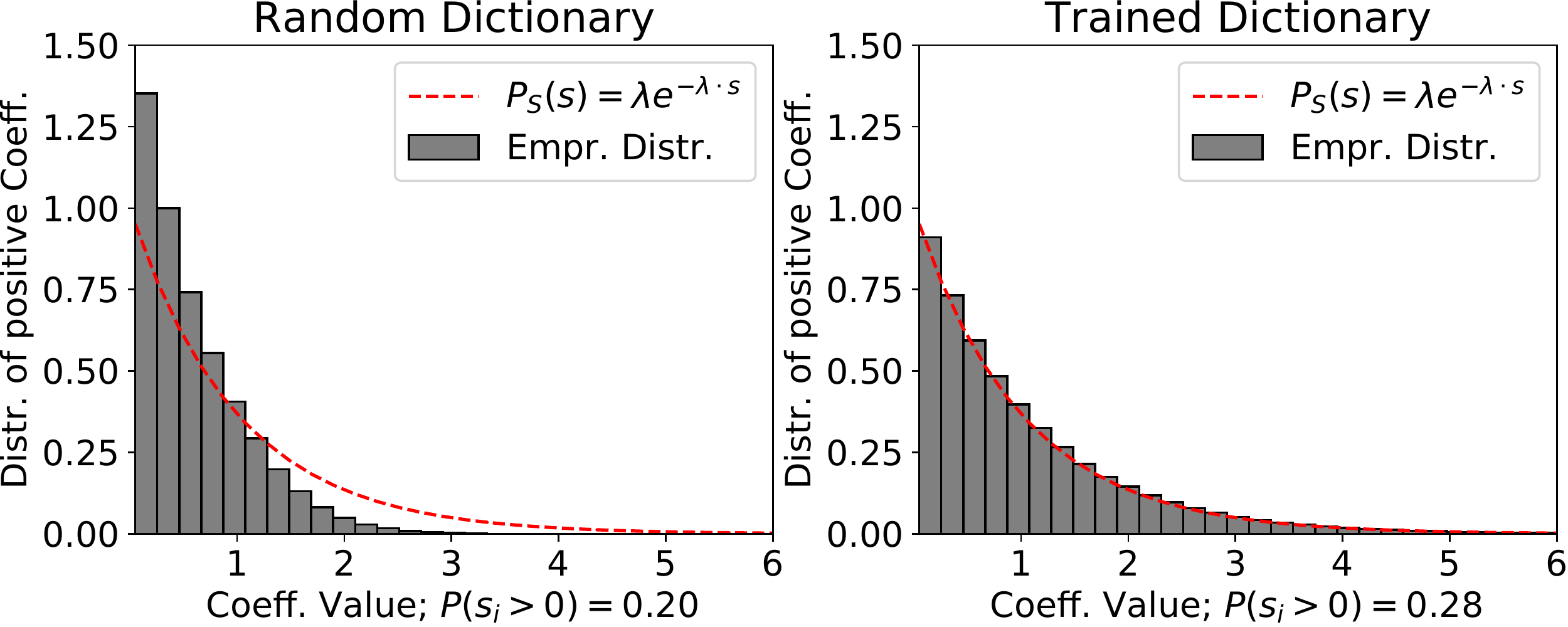}}
    \caption{The distribution of non-zero coefficients of each of the three algorithms. The dotted red line shows the prior of coefficients used in generating the dataset. The left panel of each subfigure shows the empirical distribution when each algorithm is run with random dictionaries. The right panel shows the the distribution with learned dictionaries. Only $L_0$-LSC, with the correctly trained dictionary achieves the distribution matching the prior.}
    \label{fig:bars_distr}
\end{figure}

Figure \ref{fig:learning_pi} shows the  convergence of model parameter $\pi$ to match (approximately) the actual level of sparsity in the data. To further characterize the coefficients, the distributions of the non-negative coefficients of the three algorithms were also plotted in Fig. \ref{fig:bars_distr}. Using a fixed dictionary, the algorithm was run to infer either the MAP estimate (DSC and LCA) or to sample from the posterior ($L_0$-LSC). This was done with a correctly learned dictionary (Fig. \ref{fig:bars_dict}) as well as a random dictionary (i.e. uncorrelated gaussian noise).  In addition to having the correct $L_0$-sparsity, $L_0$-LSC  correctly samples the posterior, which when averaged over the data matches the desired prior (Fig. \ref{fig:distr_lsc}), as expected from theory.  This is in contrast to non-stochastic algorithms where the inferred latent variable distribution often often exhibits a more pronounced peak at zero compared to the prior. 
A more quantitative analysis is provided in Appendix~\ref{app:dkl}.

\subsection{Learning the Dictionary Norm}
\label{sec:dict_norm}

For traditional sparse coding models such as DSC and LCA which update the dictionary based on a single MAP estimate for each data item, it is necessary to normalize the dictionary elements after each update.  However if the update is based on samples from the posterior, as specified in equation~(\ref{eq:learning-rule}), then this is no longer necessary.
As a result, when using LSC, there is no need for normalization. Instead, the dictionary element norms $||\mb A_i||$  will automatically grow or shrink as needed to optimize the model log-likelihood. 

\begin{figure}
    \centering
    \subfloat[Evolution of dictionary norms]{
    \label{fig:bars_nn_evo}
    \includegraphics[width=.45\columnwidth]{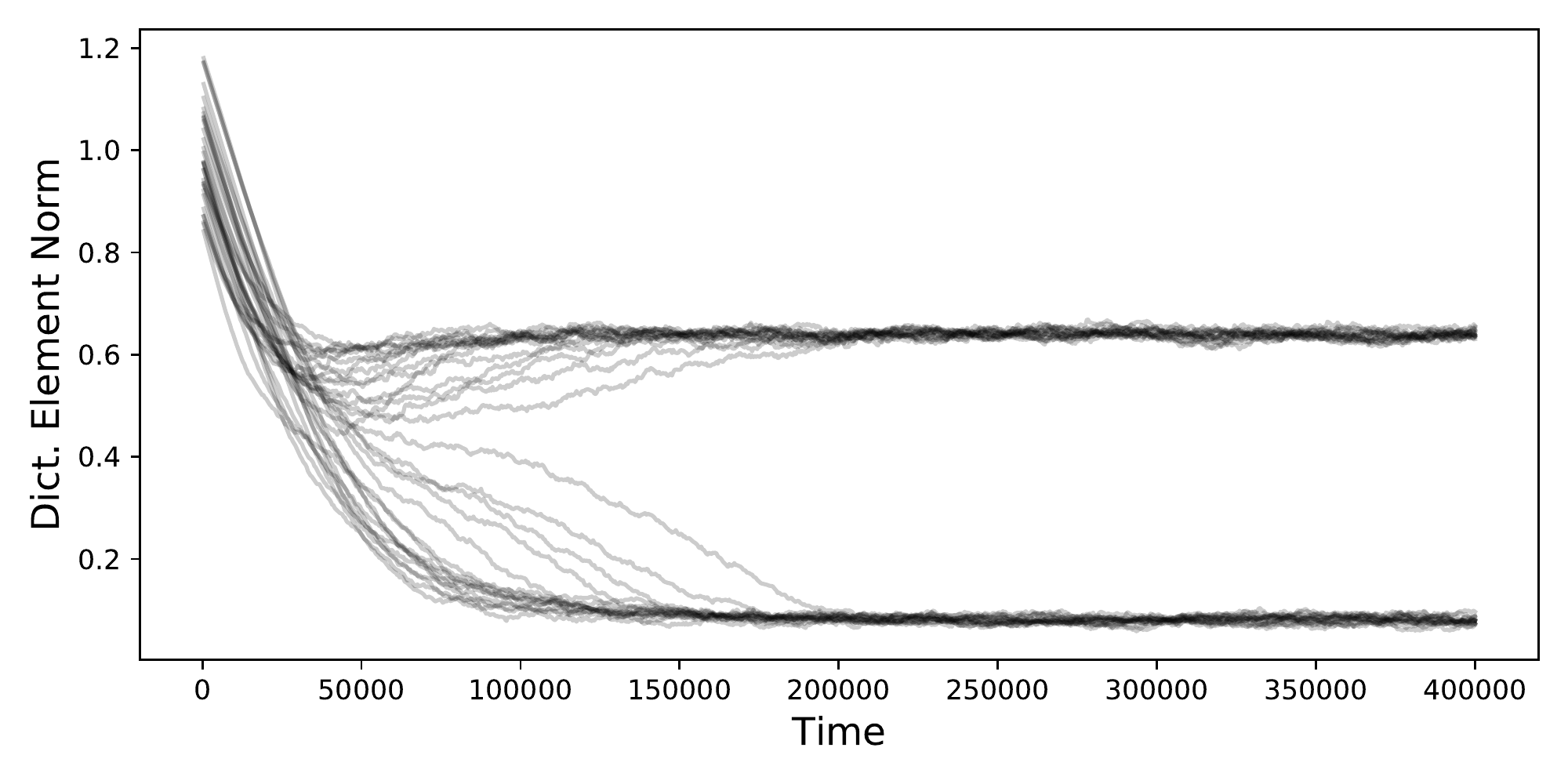}}
    \subfloat[Unnecessary dictionary elements vanish.]{
    \label{fig:bars_nn_dict}
    \includegraphics[width=.4\columnwidth]{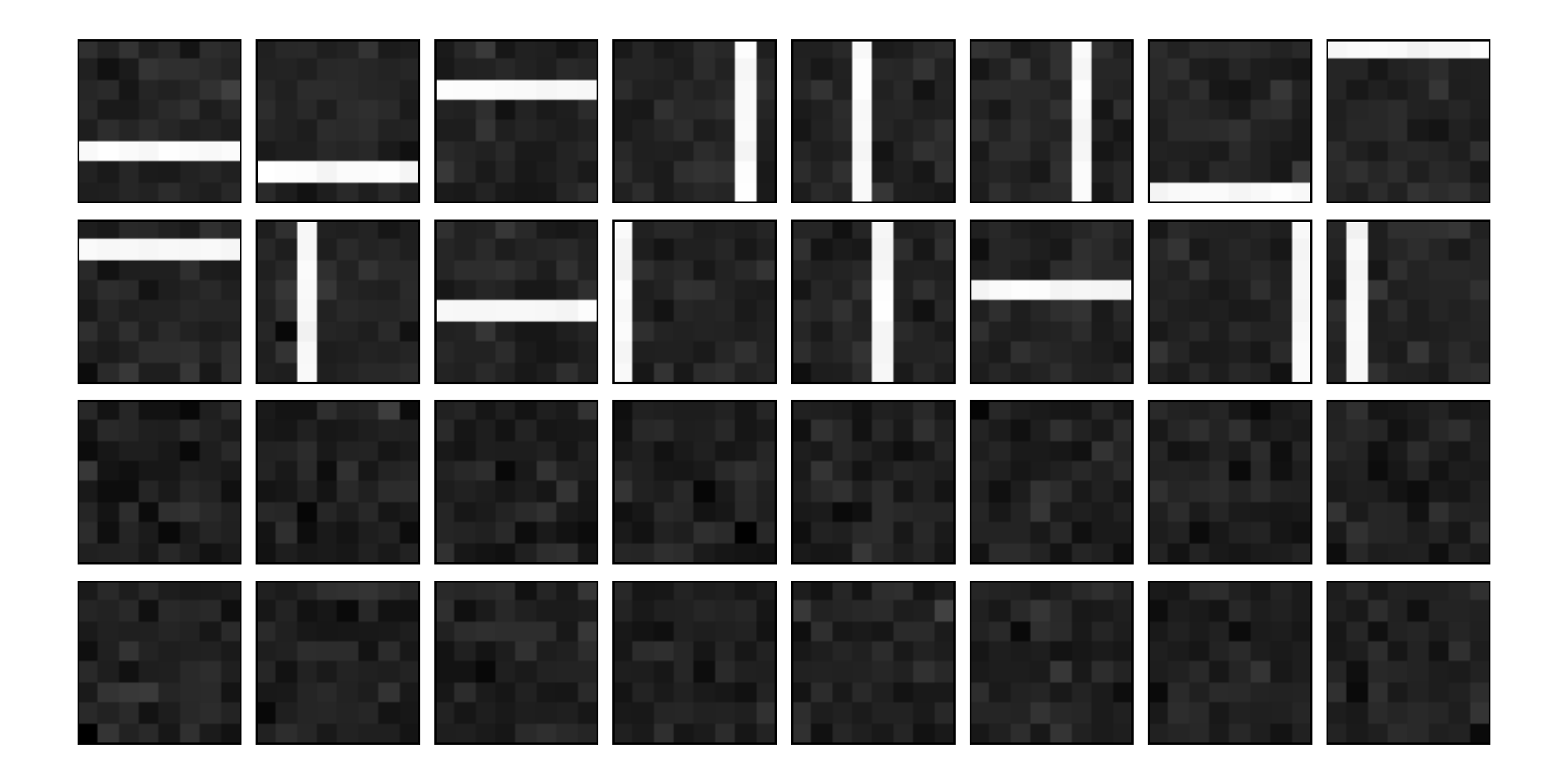}}\\
    \caption{Learning the dictionary size.  
    a) Dictionary norms bifurcate, with half decaying to nearly zero. b) The remaining elements contain exactly one copy of the dictionary elements used to generate the data.}
    \label{fig:oc_fixed_pi}
\end{figure}

The adaptive norm property can also be used to automatically select for the size of the dictionary. For data of dimension $D$, we consider a dictionary of size $K = \Omega \times D$, to have an (over)completeness of $\Omega$. A $2 \times$ overcomplete model was trained using the LSC algorithm using a fixed activation probability $\pi$, without normalizing the dictionary $A$.  The resulting learned dictionary is shown in Fig. \ref{fig:bars_nn_dict}. In previous work by \cite{berkes2008sparsity}, Annealed Importance Sampling (AIS) \cite{neal2001annealed} was used to approximate the marginal likelihood in order to find the optimal dictionary elements. However, $L_0$-LSC, without additional procedures, can be used to effectively do the same through attenuation of unnecessary dictionary elements. 
The learned dictionary contains exactly the bars dictionary and the extra elements decay to nearly zero, as shown in Figure~\ref{fig:bars_nn_evo}. 
 
\begin{figure}
    \centering
    \subfloat[Learned dictionary with duplicated elements]{
    \label{fig:full_dict}
    \includegraphics[width=.4\columnwidth]{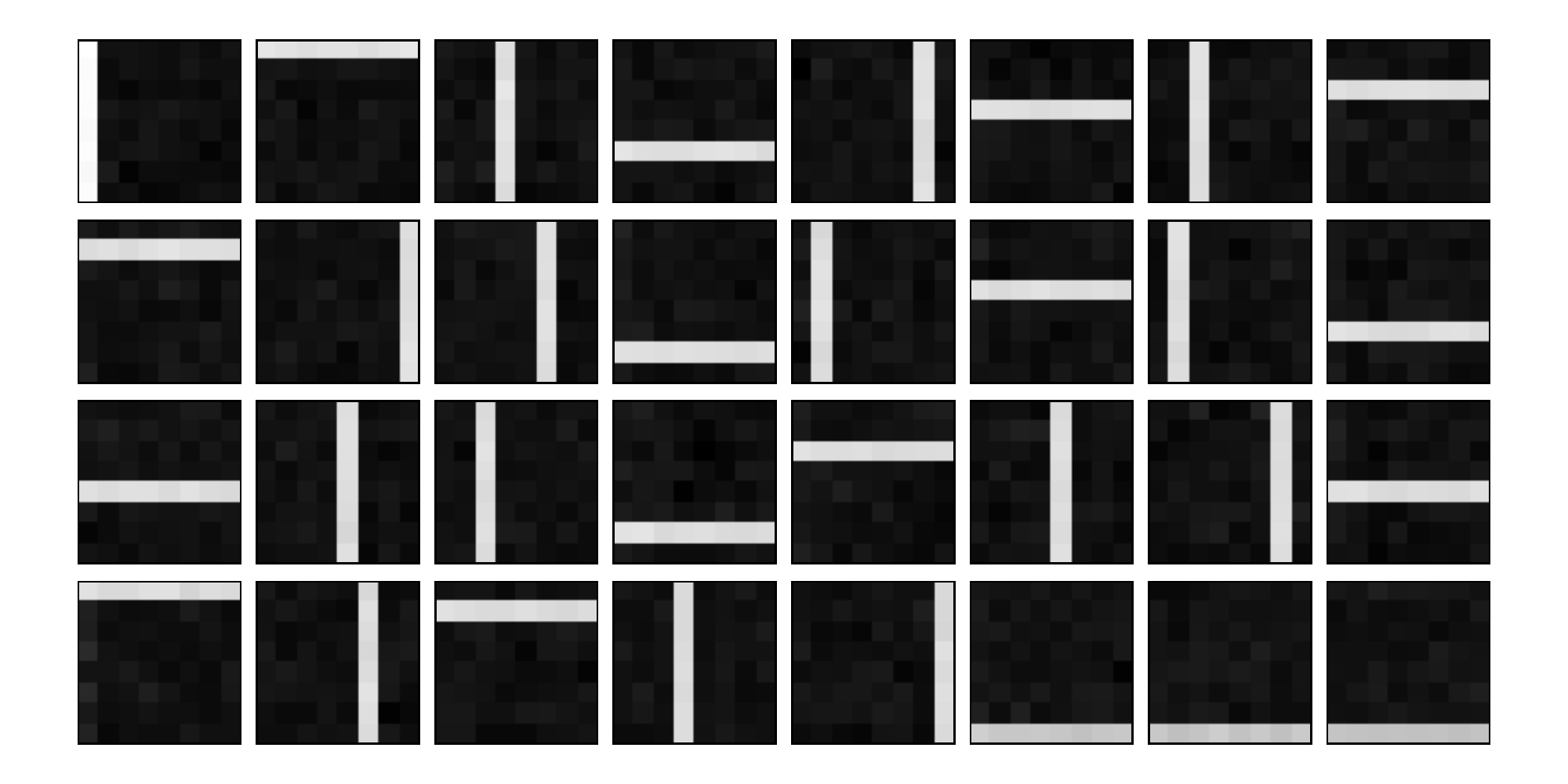}}\\
    \subfloat[Activation probability $\pi$ learned by twice overcomplete model]{
    \label{fig:full_pi}
    \includegraphics[width=.45\columnwidth]{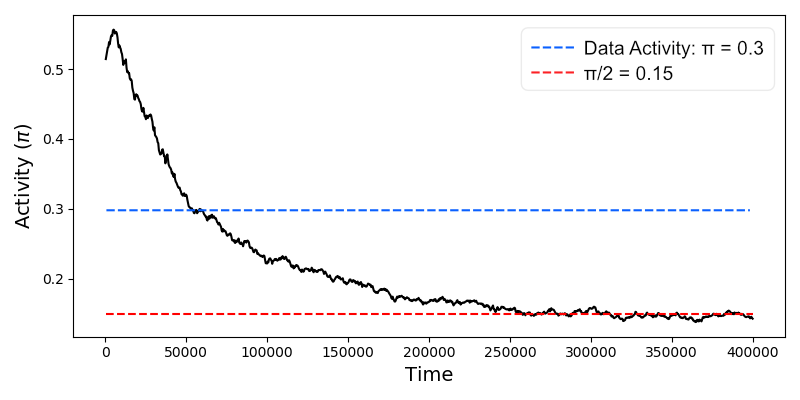}}
    \subfloat[Evolution of dictionary norms]{
    \label{fig:full_norm}
    \includegraphics[width=.45\columnwidth]{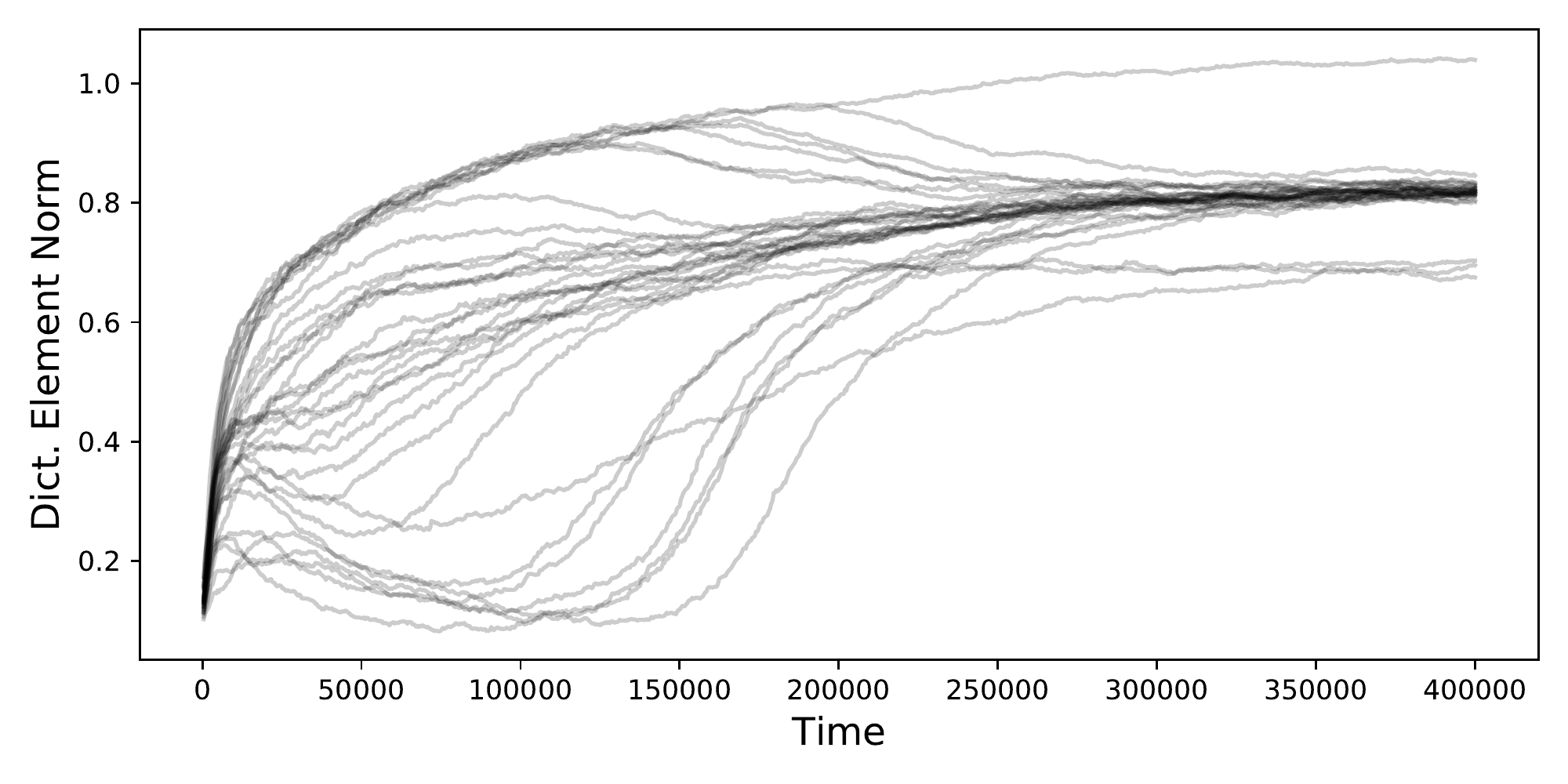}}
    \caption{LSC is used to learn both the dictionary size and activation probability of the same $2\times$ overcomplete model a) The learned dictionary now contains duplicated elements. b) But the activation probability $\pi$ is half of the actual value used in generating the data.}
\end{figure}

When both $||\mb A_i||$ and $\pi$ are being learned, a more stable solution is to have duplicated dictionary elements with a reduced activity.  This is shown in Figure \ref{fig:full_dict} with a duplicated dictionary but halved activity (Fig. \ref{fig:full_pi}).

\subsection{Natural Image Patches}
\label{sec:vh_results}
\begin{figure}
    \centering
    \includegraphics[width=1\columnwidth]{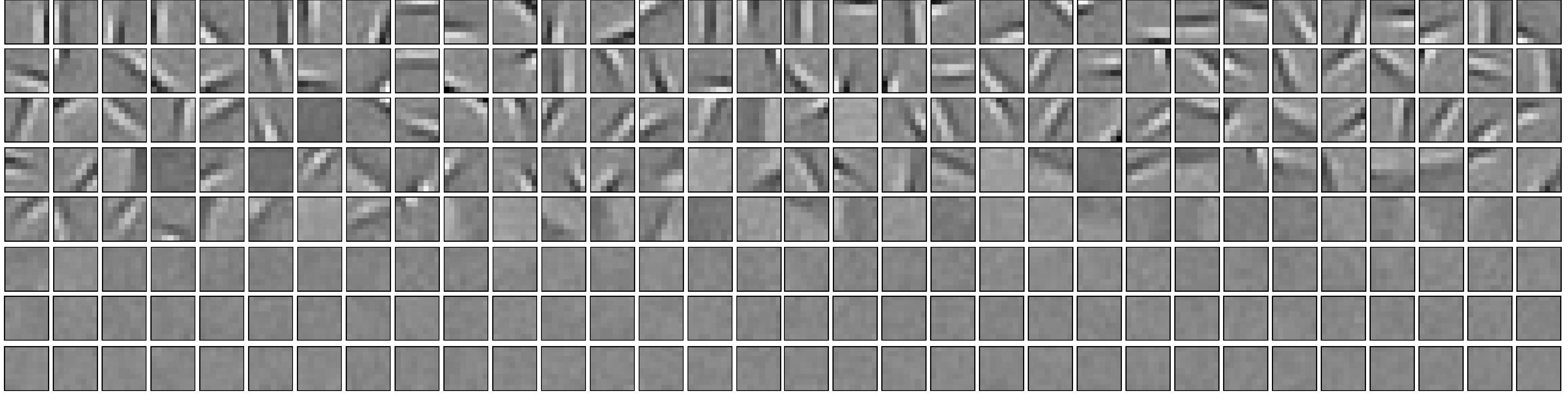}
    \caption{With activity fixed ($\pi = 0.5$), only a fraction of the total dictionary elements have significant norm, the rest vanish. The dictionary elements are sorted by their respective norms.}
    \label{fig:fixed_pi_dict}
\end{figure}

We ran the $L_0$-LSC algorithm on a dataset of $8\times8$ image patches of whitened natural scenes from the Van Hateren dataset\cite{hateren_schaaf_1998,olshausen2013highly}. First, the model activity was fixed at $\pi = 0.5$ and we used $L_0$-LSC to learn a $4 \times$ overcomplete dictionary ($K = 4 \times 64 = 256$). We can see in Figure \ref{fig:fixed_pi_dict} that a little more than half of the dictionary was utilized. The unused dictionary elements had a comparatively insignificant norm. In contrast to prior efforts to determine the optimal number of dictionary elements based on approximating the log-likelihood \cite{berkes2008sparsity}, this result emerges directly from dictionary learning in Langevin sparse coding.

\begin{figure}[ht!]
    \centering
    \subfloat[Mean activity at different levels of overcompleteness]{
    \label{fig:oc_vs_pi}
    \includegraphics[width=.45\columnwidth]{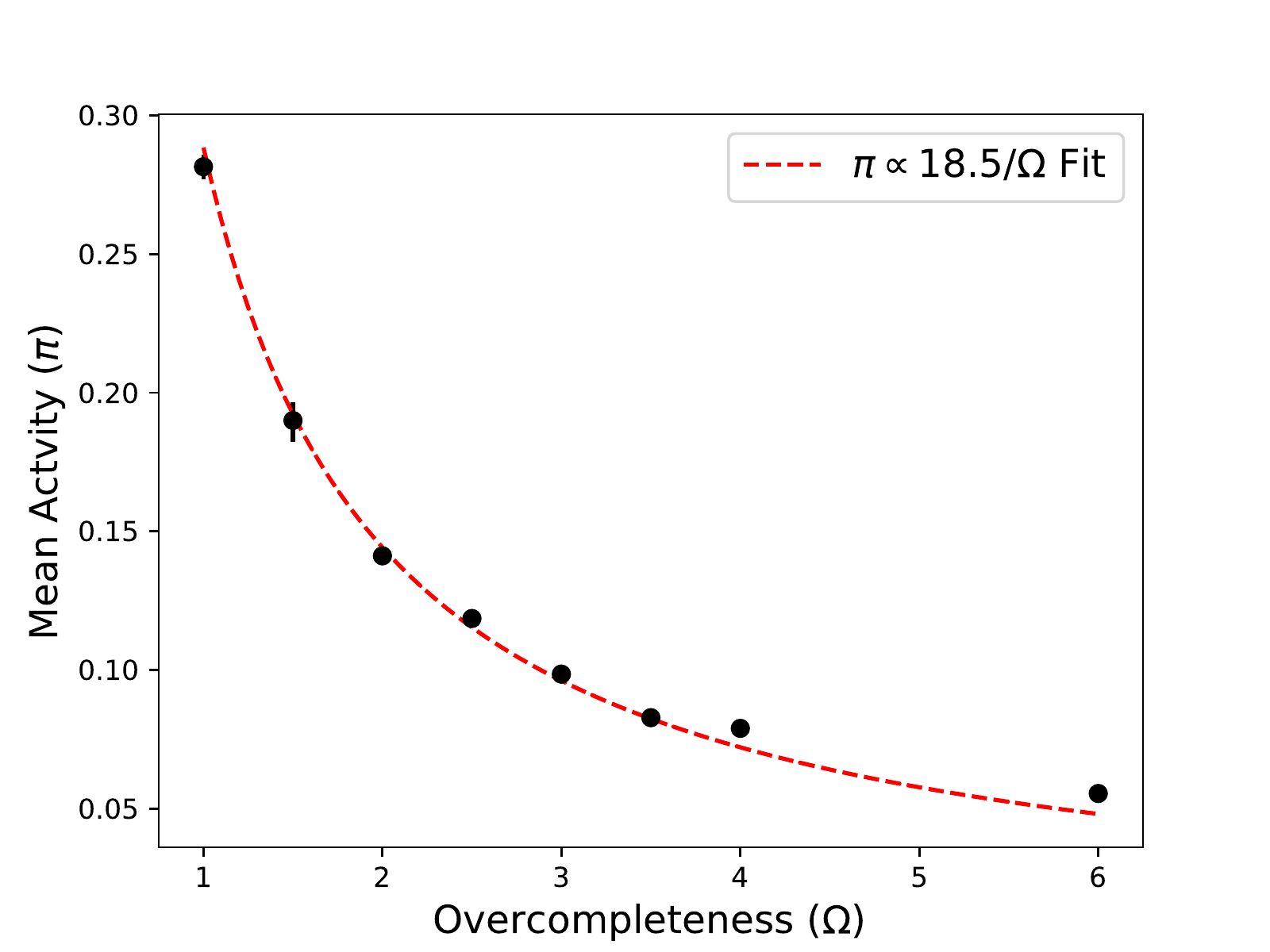}
    }
    \subfloat[Mean number of active coefficients]{
    \label{fig:mean_num_dict}
    \includegraphics[width=.45\columnwidth]{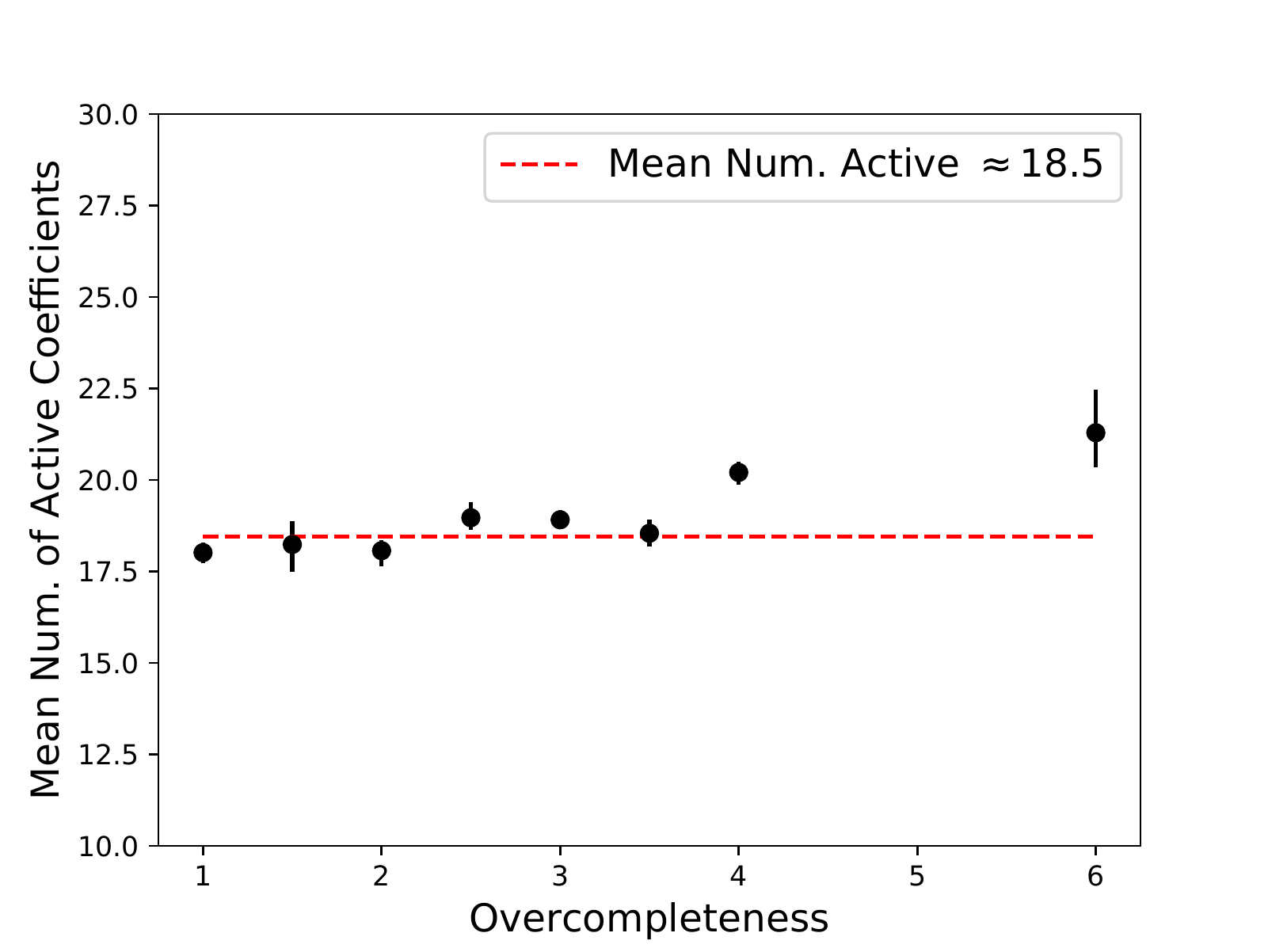}
    }
    \caption{a) Using LSC to learn dictionaries for natural scenes at different levels of completeness $\Omega$, the relationship $\pi \propto 1/\Omega$ is obtained. b) This implies that the mean number of dictionary elements used to code each image is constant irrespective of the total number of dictionary elements learned. Error bars on both plots denote the 10\% - 90\% range}
\end{figure}

Then, unfixing $\pi$, we allow the activity to be learned. Repeating the experiment at different levels of overcompleteness $\Omega$, a correspondence between the activity and overcompleteness is plotted in Figure \ref{fig:oc_vs_pi}. This relationship happens to be very well modeled by $\pi \propto \Omega^{-1}$. As a consequence, the expected number of active dictionary elements, $\pi \times K = \pi \times \Omega \times D$ stays nearly constant irrespective of the overcompleteness $\Omega$.

\section{Discussion}

Our main contribution in this paper is to show that by using Langevin dynamics to sample from posterior distributions, we obtain a set of continuous-time equations over analog state variables that enable probabilistically correct inference and learning in a latent variable model.  
While the use of Langevin dynamics for sampling in probabilistic models {\em per se} is not new \cite{cheng2018underdamped}, our emphasis here is to show how these dynamics play out in the case of the sparse coding model, and to point the way toward their efficient implementation in analog, electronic circuits that harness natural sources of stochasticity, for which we provide an example in Appendix~\ref{sec:hardware}.  The basic operations involve computing inner products, thresholding, lateral inhibition, and thresholding, in addition to injection of a Gaussian noise source.  The first four of of these are shared with LCA, for which there already exist examples of both efficient analog implementations~\cite{shapero2012low, sheridan2017sparse}, and digital implementation using spiking neurons~\cite{davies2021advancing}.  In the latter case LCA was shown to achieve the highest efficiency gains.  The only additional component required for implementing LSC or $L_0$-LSC beyond these existing implementations is the injection of a Gaussian noise source.  This would seem quite natural since noise is intrinsic to any physical system, however shaping the noise to be Gaussian and i.i.d., and whether this is strictly required, remain important issues to resolve.


Finding efficient implementations is key to making probabilistic models tractable and scalable to practical problems of interest such as image analysis.  
Indeed, latent variable models such as Boltzmann machines are often considered intractable due to the inner loop required to sample over hidden unit states conditioned on input data.  For this reason, practitioners often turn to approximations such as restricted Boltzmann machines (RBM's) \cite{hinton2006reducing} or variational inference (VAE's) \cite{kingma2013auto} so as to make the problem tractable by eliminating ``explaining away" -- i.e., dependencies among hidden units conditioned on the data. But for most problems of interest in perception, explaining away is key~\cite{olshausen201427}.  So doing away with explaining away in the interest of making the problem tractable simply dodges the very problem that needs to be solved.  Here we show that there is alternative approach that tackles sampling from posteriors head on and makes it tractable via dynamics that could be naturally realized in a physical system.

An important next step will be to improve the efficiency of sampling by developing richer dynamical models.  It is well known that the first-order Langevin dynamics we have utilized here can be slow to mix and reach equilibrium~\cite{hennequin2014fast}.  Adding higher-order terms to the dynamics such as momentum or even third-order terms has been shown to dramatically improve mixing time~\cite{mou2021high}, and it has even been proposed that the balanced excitatory and inhibitory recurrent networks in cortex could serve such a function~\cite{hennequin2014fast,echeveste2020cortical}.  The 
model we have proposed here could be modified along similar lines, and indeed this is a topic of ongoing work.  Yet another route is to harness recent improvements in Hamiltonian Monte Carlo~\cite{sohl2014hamiltonian}. 

With an efficient sampler in place, it becomes possible to adapt parameters of a sparse coding model beyond the dictionary, such as the level of sparsity or overcompleteness, which has not been possible in previous MAP-estimate based approaches.
Furthermore, through application of a threshold function to the stochastic dynamics, \emph{we demonstrate that inference with an $L_0$-sparse prior -- which has been avoided in most approaches by using $L_1$ as a proxy -- can be readily computed and implemented} (Sec. \ref{sec:l0_sparse}).  
As shown in Section \ref{sec:bars}, $L_0$-LSC is better at sampling from the posterior distribution as well as capable of learning the activation probability $\pi$ of the latent variables $\mb s$.  In applying the model to natural images (Sec. \ref{sec:vh_results}), we found that \emph{the mean number of dictionary elements used to encode an image is mostly invariant to the total dictionary size}. This runs counter to previous results \cite{olshausen2013highly} showing that, on average, the number of elements required for reconstructing a given image decreases with larger dictionaries in which the elements take on more specific and diverse shapes.
This discrepancy could possibly be reconciled by the fact that the previous work utilized MAP-estimates rather than sampling, and so the learning was biased accordingly.
Nonetheless, it is still intriguing that the mean number of dictionary elements in our case was near constant, suggesting that overcompleteness is an under-utilized degree of freedom.  However, another likely culprit is the assumption of a factorial prior, and it may be that an overcomplete dictionary loses its explanatory power under such a prior.  Thus, it will be important to consider group sparse coding or other approaches for modeling statistical dependencies among latent variables~\cite{garrigues2010group,garrigues2007learning} in order to fully realize the gains from overcompleteness. 

Finally, another contribution of this work is to show how
both learning and inference can be mapped to \emph{simultaneous dynamics} at two different time scales. An underlying assumption in all implementations of probabilistic models on digital systems is the notion of a \emph{global clock}. But the global clock is an impossibility for neural systems of any significant complexity. Our work presents an alternative approach to computing sparse coding which allows for simultaneous updates of both latent variables and model parameters such as the dictionary elements. This type of concurrent dynamics removes the need of any such global clock.

More generally, the mixed time-scale analog sampling framework on which LSC is based  opens the way to learning richer generative models that capture dependencies among latent variables via horizontal connections~\cite{garrigues2007learning} or via top-down priors~\cite{boutin2020effect}.  And this goes beyond just sparse coding. In the future we hope to develop analogous procedures for learning other latent variable models such as Boltzmann machines and hierarchical Bayesian models \cite{lee2003hierarchical}.




\newpage
\appendix
\begin{subappendices}
\section{Time-scaling property of Langevin Dynamics}

Consider the results of scaling the time variable $t$ by a constant $\tau$
\begin{align}
    \tilde t = \tau \cdot t.
\end{align}
Recall that the Gaussian white noise $\xi(t)$ was normalized such that
\begin{align}
    \langle \xi(t) \xi(t') \rangle = I \delta(t - t').
\end{align}
Using the new, scaled time, we have
\begin{align}
    \langle \xi(\tilde t) \xi(\tilde t') \rangle &= I\delta(\tau(t - t'))\\
    &= \frac1\tau I\delta(t - t').
\end{align}
If we define $\tilde \xi(\tilde t) = \sqrt(\tau) \xi(\tau t)$, we will recover
\begin{align}
    \langle \tilde \xi(\tilde t) \tilde \xi(\tilde t') \rangle = I \delta(t - t').
\end{align}

\section{Quantifying convergence to prior}
\label{app:dkl}
\begin{figure}[ht!]
    \centering
    \includegraphics[width=.6\columnwidth]{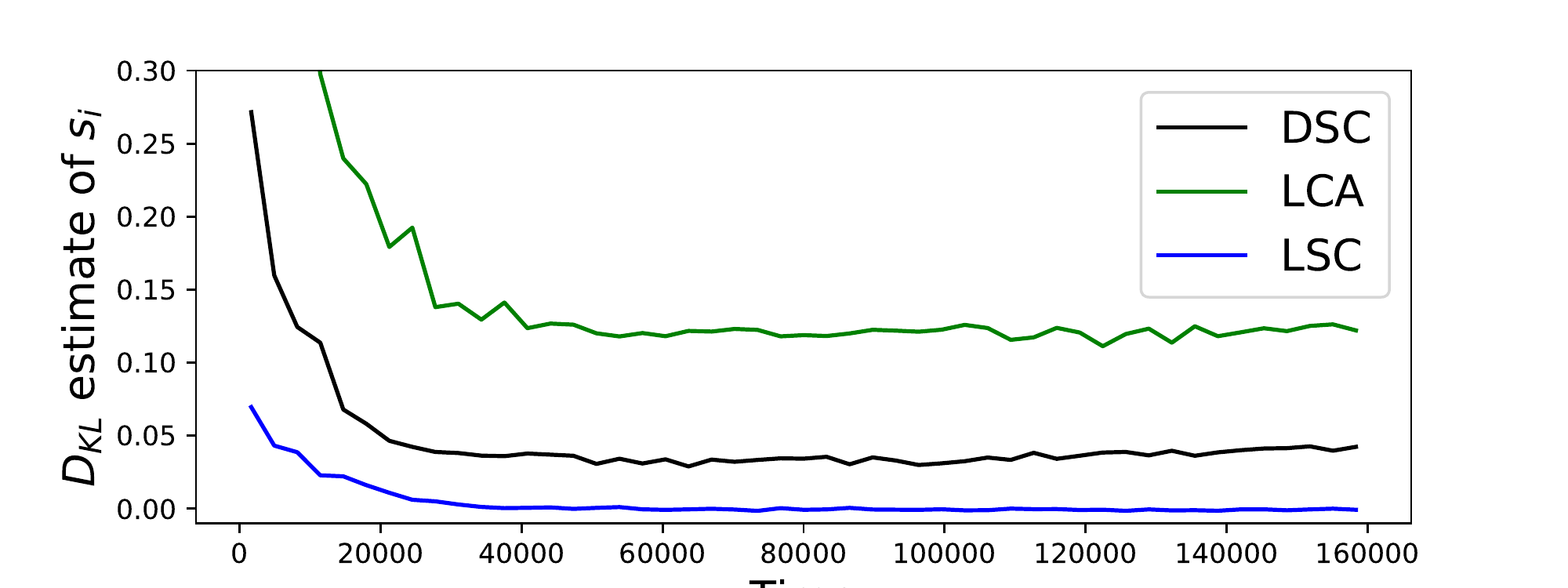}
    \caption{The KL-Divergence for coefficients $s_i$ is compared for each of the three sparse coding methods. Only with LSC, does the $D_{KL}$ approach zero.}
    \label{fig:bars_d_kl_s}
\end{figure}

To better quantify the convergence to the desired prior, we estimate the KL-divergence from $p(s_i|\lambda)$, the target prior to $p(s_i | A)$ the learned prior based on dictionary $A$. Because the learned prior cannot be easily calculated, we rely on samples taken at regular time intervals. The samples are then binned in the same way that generated the histograms in (Fig. \ref{fig:bars_distr}) .
\begin{align}
    D_{KL}(p(s|\lambda) || p(s|A)) &= \left\langle \log\left(\frac{p(s|\lambda)}{p(s|A)}\right) \right\rangle_{s|\lambda}\\
    &\approx \sum_n p_n(\lambda) \log\left(\frac{p_n(\lambda)}{q_n(A)} \right)
\end{align}
where
\begin{align}
    p_n(\lambda) = P(n \delta s < s < (n+1) \delta s)
\end{align}
with $\delta s$ being the bin width. Figure \ref{fig:bars_d_kl_s} shows the evolution of the estimated $D_{KL}$ over time. As expected, only with LSC does the KL-divergence approach 0.

\section{Hardware Implementation}
\label{sec:hardware}

While the results presented above were obtained from simulation on a digital computer using the Euler-Maruyama algorithm, LSC was designed with stochastic analog implementation in mind. In this section, we present one candidate hardware implementation making use of Gilbert cells, a type of fast analog voltage multiplier \cite{gilbert1968precise}.

The first goal is to design an analog circuit capable of simulating the coupled differential equations of continuous-time sparse coding (Eq. \ref{eq:ctsc_diffeq}), explicitly
\begin{align}
    \label{eq:hw_deq_1}
    \tau_S \dot {\mb s} &= - A^T (A \mb s - \mb x) - \lambda_1 \text{sgn}(\mb s)\\
    \label{eq:hw_deq_2}
    \tau_A \dot A &= - (A \mb s - \mb x) \mb s^T
\end{align}

The core design challenge is to dynamically update $\mb s$ through matrix multiplication with $A^T A$. Simultaneously, we require the dictionary elements $A$ to also change in accordance with the value of $\mb s$. One promising approach is using grids of memristors, or programmable resistors \cite{di2009circuit}, 
which have been proposed as an analog implementation of generative adversarial networks \cite{krestinskaya2020memristive}. However, the limited endurance of memristors prevents extensive rewrites and ultimately a fully analog implementation \cite{krestinskaya2020memristive}. As an alternative, we propose using arrays of Gilbert cells for matrix multiplication. Because both inputs and the output are voltages, continuous dynamic updates are easily possible.

\subsection{Gilbert Cell Matrix Multiplier}
To focus on the matrix multiplication, we simplify Eqs. \ref{eq:hw_deq_1}-\ref{eq:hw_deq_2}, at least initially, by ignoring the sparsity term, taking $\lambda_1 = 0$. We can also better organize the equations by introducing the reconstruction error variable, $ \Delta = A \mb s - \mb x$. Finally, integrating the differential equations, we obtain

\begin{align}
    \label{eq:d_As}
    \Delta &= A\mb s - \mb x\\
    \label{eq:s_Ad}
    \mb s &= -\tau_s^{-1} \int dt~ A^T \Delta\\
    \label{eq:A_ds}
    A &= - \tau_A^{-1} \int dt~ \Delta\, \mb s^T
\end{align}

We represent each of the variables $A, \mb s, \mb x, \Delta$ by proportional electric potentials.
\begin{align}
    V^{(A)} &\propto A\\
    V^{(S)} &\propto \mb s\\
    V^{(X)} &\propto \mb x
\end{align}

Multiplication of matrices consists of element-wise multiplication which is facilitated by Gilbert cells and summation, which is facilitated through operational amplifiers.

\begin{figure}[ht!]
    \centering
    \includegraphics[height=2.5in]{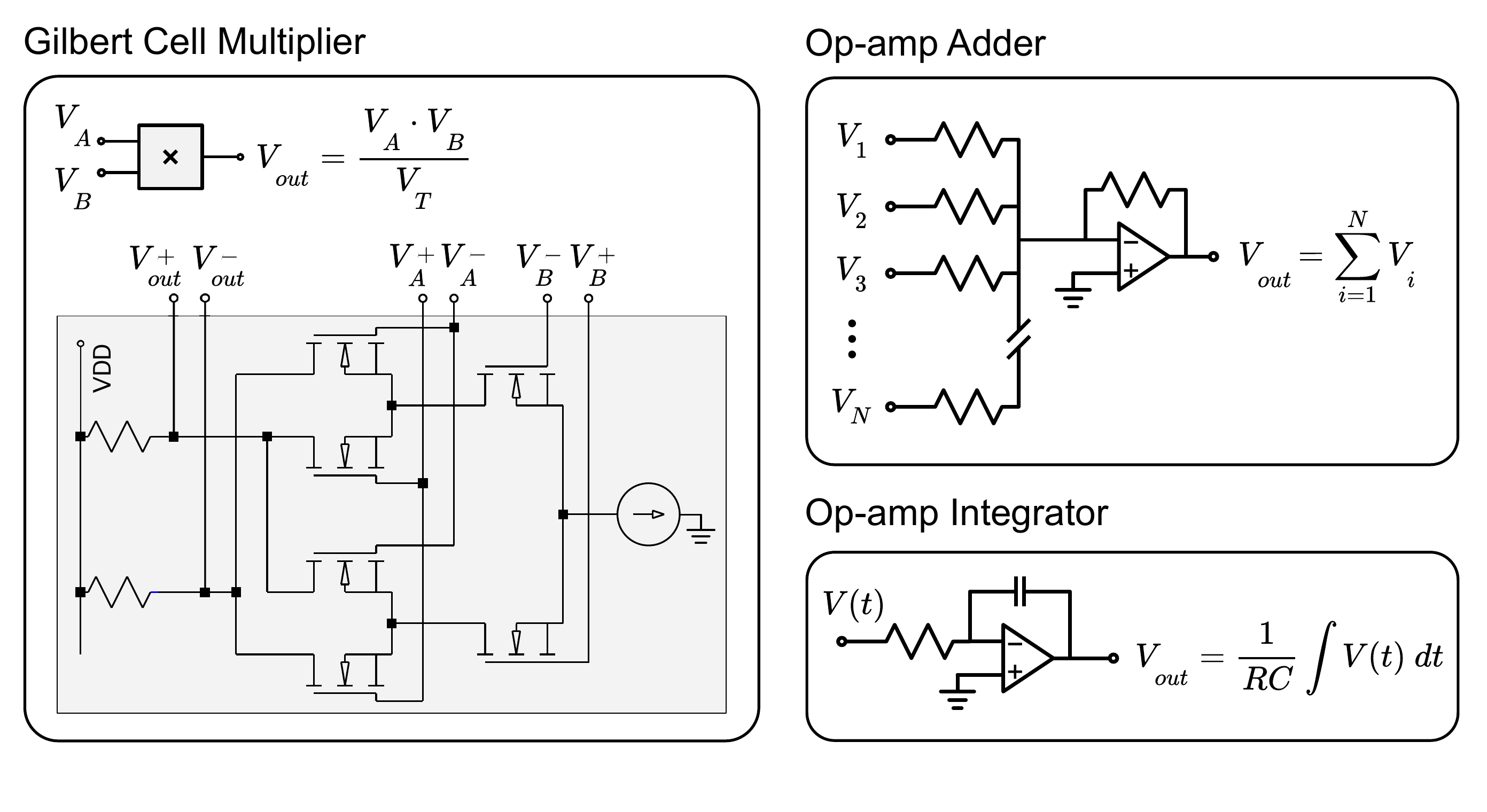}
    \caption{Circuit elements used for multiplication, summation and integration of voltages. The op-amp adder is used to add elements. The op-amp integrator is typically used to integrate over time.}
    \label{fig:elements}
\end{figure}

A schematic of a Gilbert cell is shown in Figure \ref{fig:elements}. With inputs given as the voltage differences $V_A \equiv V_A^+ - V_A^-$ and $V_B \equiv V_B^+ - V_B^-$, the multiplier produces an output proportional to the product of the inputs.

\begin{align}
    \label{eq:gilbert}
    V_{out} = V_{out}^+ - V_{out}^- = \frac{1}{V_T} (V_A^+ - V_A^-) \cdot (V_B^+ - V_B^-) = \frac{V_A \cdot V_B}{V_T}.
\end{align}
The constant $V_T$ depends on the design of the cell, choice of transistors, and other factors.

\begin{figure}[ht!]
    \centering
    \subfloat[Circuit for $\Delta = A\mb {s-x}$]{
    \label{fig:matmul_as}
    \includegraphics[height=2.5in]{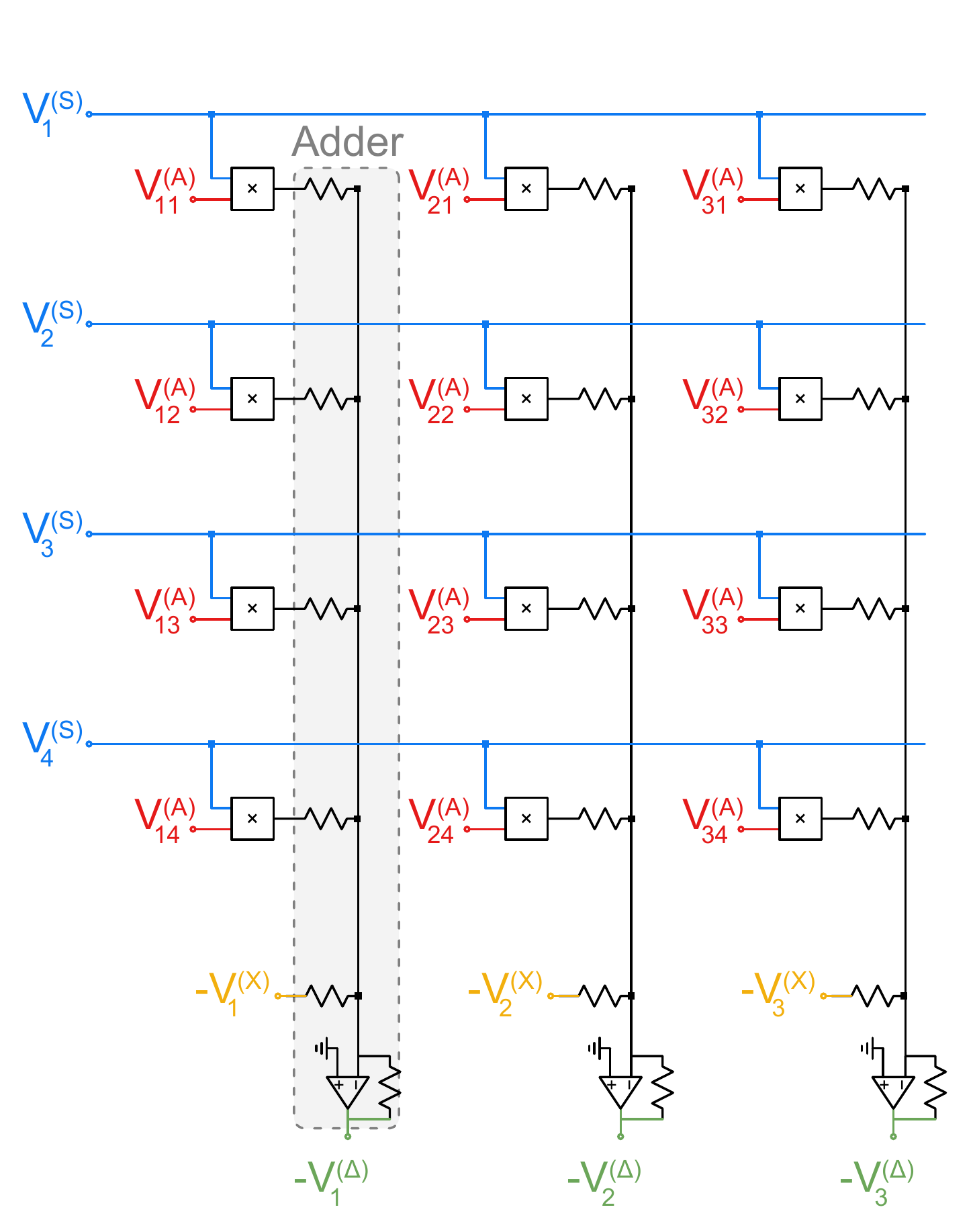}
    }
    \subfloat[$\mb s \propto - \int dt A^T \Delta$]{
    \label{fig:AAS}
    \includegraphics[height=2.5in]{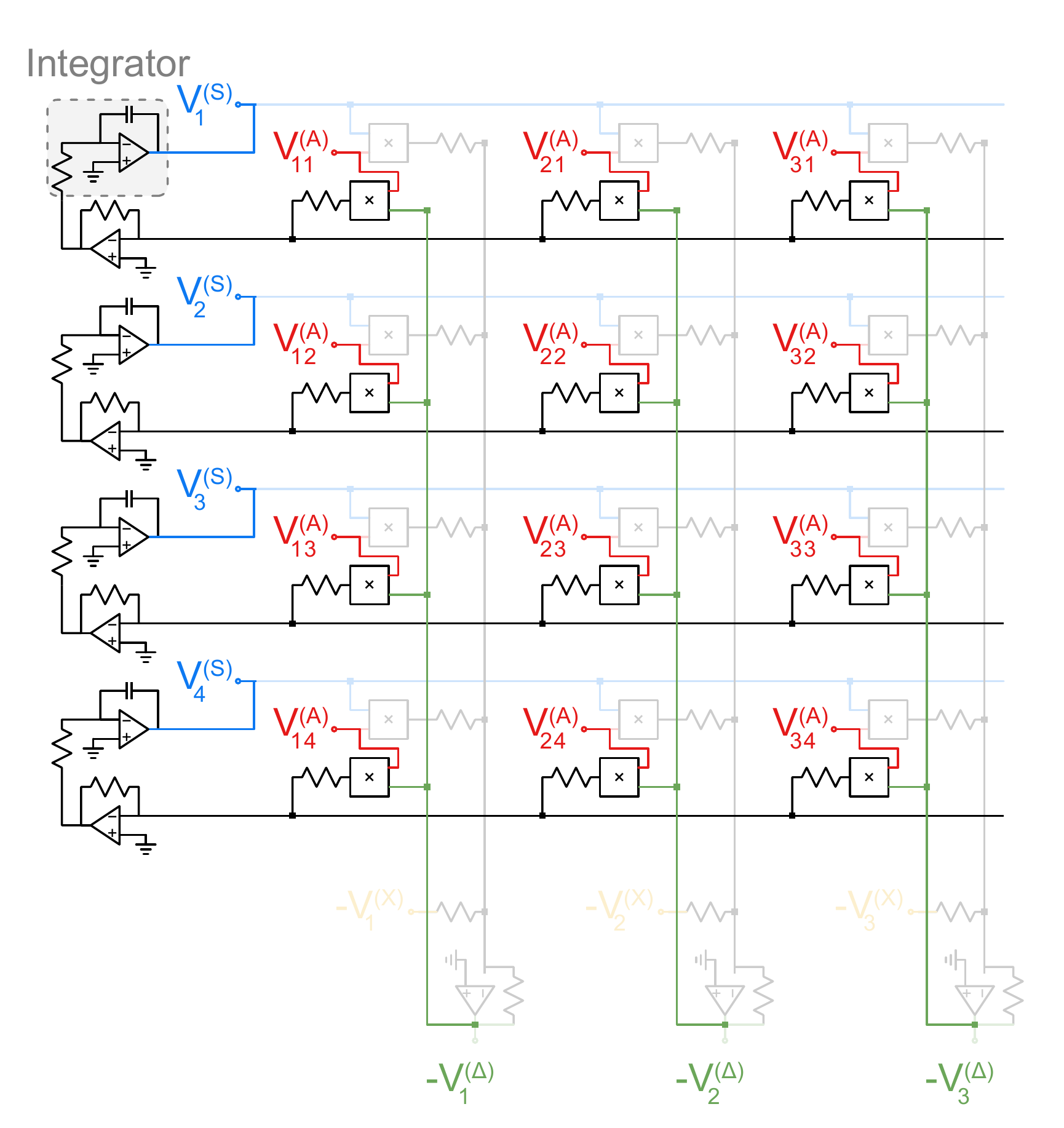}
    }
    \subfloat[Full circuit with triple multipliers]{
    \label{fig:full_loop}
    \includegraphics[height=2.5in]{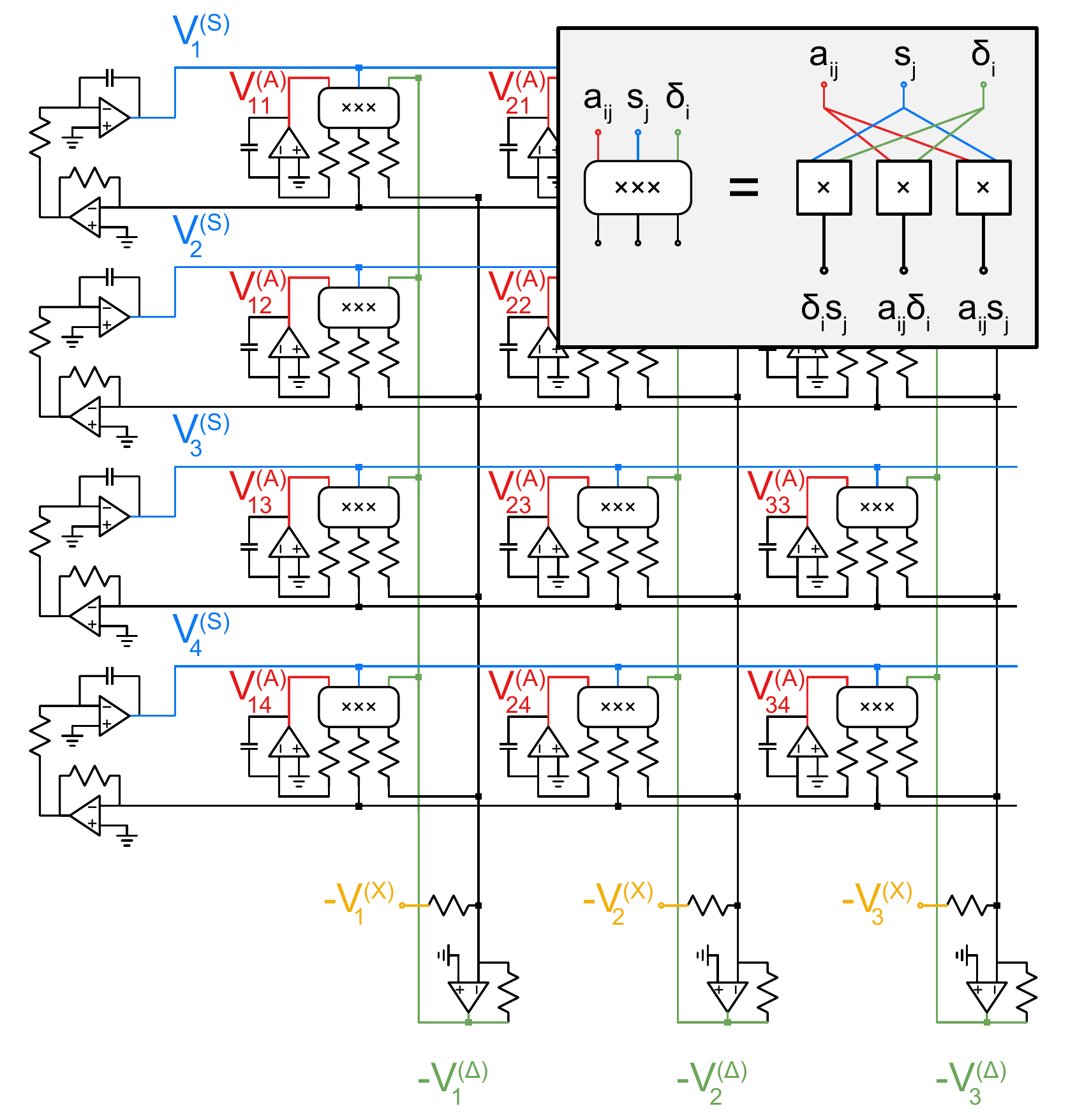}
    }
    \caption{a) An analog matrix multiplier built from a grid of Gilbert cells and op-amp adders. Note the extra row of input $V^{(X)}$ makes this technically an affine transformation. b) The output of the previous circuit, $V^{(\Delta)}$, is fed back to another set of multipliers and integrated. The new multiplier array are weaved into the existing circuit (which are faded to emphasize the newly added multipliers). c) With an array of triple multipliers, this circuit allows for the dictionary elements $V^{(A)}_{ij}$ to be updated concurrent with the coefficients $V^{(s)}_{i}$. In practice, the Gilbert cells input and output potential differences. However, for brevity each pair of potentials are represented by a single node -- see App. \ref{app:circuit} for detailed circuit diagrams.}
\end{figure}

Using operational amplifiers (op-amps) for summing electric potentials \cite{mancini2003op} (Fig. \ref{fig:elements}), matrix multiplication of analog signals can be implemented with a grid of Gilbert cells. This is demonstrated in (Fig. \ref{fig:matmul_as}) specifically for $\Delta = A \mb s - \mb x$. An array of nodes with potentials represent elements $A$ are shown in red. Wires running horizontally carrying potentials representing elements of $\mb s$ are shown in blue. Each Gilbert cell multiplies the two sets of input potentials to produce the output $V_T^{-1} V^{(A)}_{ij} V^{(S)}_{j}$. This output is then subsequently summed together by a series of op-amp adders shown running vertically in the figure. An extra row of inputs, $-V^{(X)}_{n}$ accounts for the needed bias, making this an affine transformation (rather than a matrix multiplication). Finally the output at the bottom of the figure is 
\begin{align}
    V^{(\Delta)}_i = V_T^{-1}\sum_{j} V^{(A)}_{ij} V^{(S)}_{j} - V^{(X)}_i.
\end{align}
Recall that $V_T$ is a constant which depends on specific Gilbert cell. This first circuit implements Eq. \ref{eq:d_As} as desired.

In a similar manner, an implementation of $\tau_s \dot {\mb s} = -A^T \Delta$ can be obtained. Because the voltages associated with matrix elements $V^{(A)}_{ij}$ are already present, a second set of Gilbert cells can be woven into the previous circuit (Fig. \ref{fig:AAS}). Here, the potentials $V^{(\Delta)}_i$ are propagated through vertical wires and with op-amp adders running horizontally, the resulting product is

\begin{align}
    -V_T^{-1}\sum_{i} V^{(A)}_{ij} V^{(\Delta)}_{i} = -V_T^{-1}\sum_{i} V^{(A^T)}_{ji} V^{(\Delta)}_{i} = - V^{-1}_T \left(V^{(A_T)}V^{(\Delta)}\right)_j.
\end{align}
To integrate the above output, we make use of op-amp integrators (see. Fig. \ref{fig:elements}). Passing through a set of integrators and looping back to $V^{(S)}_j$, we obtain
\begin{align}
    V^{(S)} = -\tau_S \int dx~ V^{-1}_T V^{(A_T)}V^{(\Delta)}
\end{align}

The time constant is dependent on op-amp integrator (i.e. $\tau_S = RC$) and can be adjusted accordingly. The newly introduced circuitry further enforces Eq. \ref{eq:s_Ad}.

Lastly, for Eq. \ref{eq:A_ds}, another set of Gilbert cells to multiply $\Delta_{i}$ and $s_{j}$ is added. Its output is then integrated and fedback back into $V^{(A)}_{ij}$. Figure \ref{fig:full_loop} presents the complete circuit for implementing the coupled equations \ref{eq:d_As} - \ref{eq:A_ds}. For conciseness, we introduce the triple multiplier comprising of three Gilbert cells (see Fig. \ref{fig:full_loop} inset). With inputs of $a_{ij}, s_{j}, \Delta_{i}$, it outputs all pairwise products $\delta_i s_j, a_{ij} \Delta_i$ and $ a_{ij} s_j$.

Note that in the complete circuit, the voltages $V^{(S)}_j, V^{(A)}_{ij}, V^{(\Delta)}_i$ no longer are inputs to the system but rather represent ``internal variables.''  Only $V^{(X)}_i$ is set externally and the potential at the remaining nodes evolve according to the coupled equations. The fact that the evolution of $A$ and $\mb s$ requires neither external measurement nor global clocking is exactly the desired result sought out of the fully analog system. We further demonstrate in App. \ref{app:batch} that the triple multiplier array can be easily modified to accept asynchronous batched inputs.

\subsection{$L_1$-\ctsc Circuit}
\begin{figure}[ht!]
    \centering
    \subfloat[$L_1$-\ctsc circuit operating for 30ms]{
    \label{fig:l1_short}
    \includegraphics[width=0.5\columnwidth]{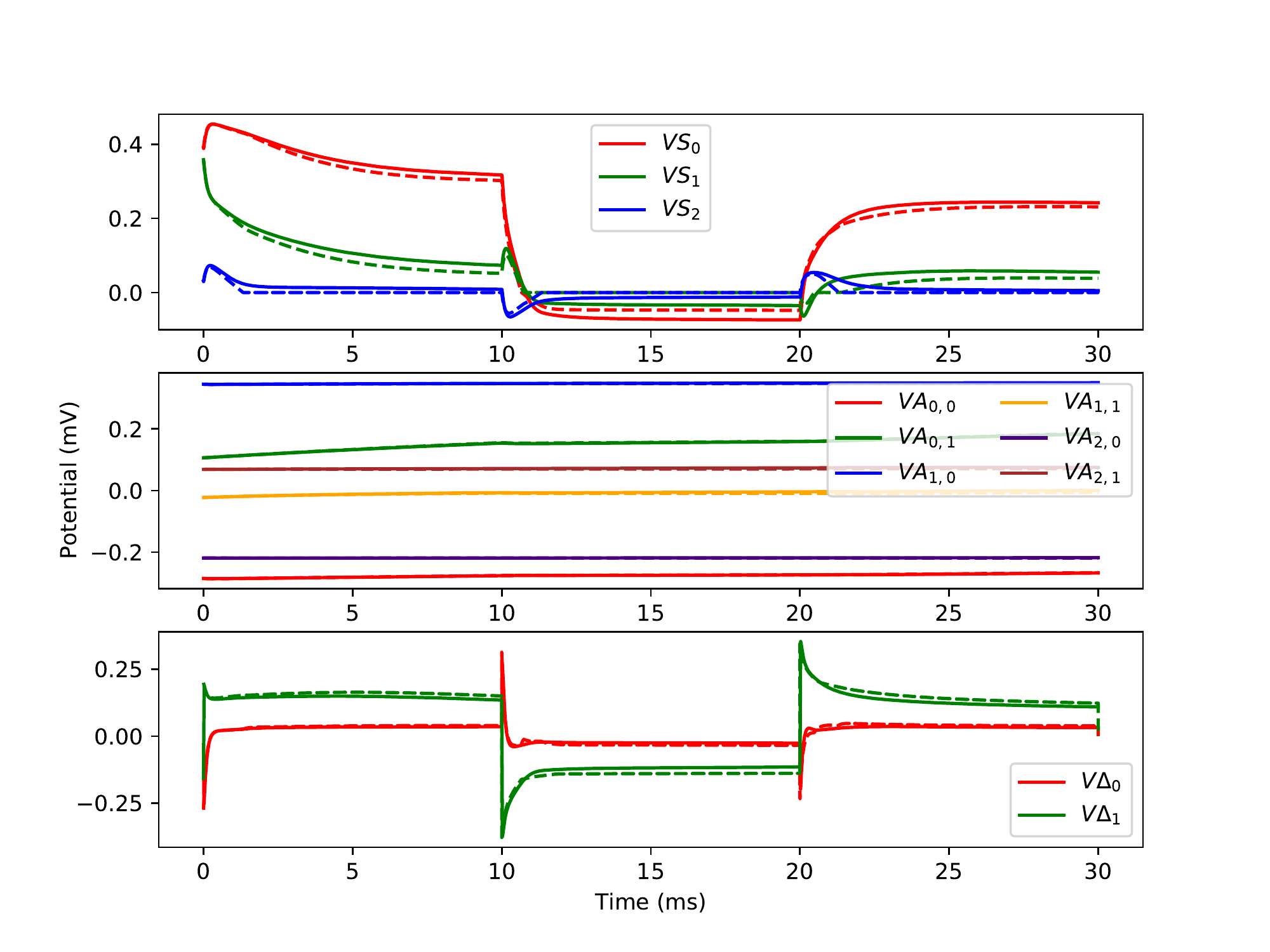}
    }
    \subfloat[$L_1$-\ctsc circuit for 1s over 100 input batches]{
    \label{fig:l1_long}
    \includegraphics[width=0.5\columnwidth]{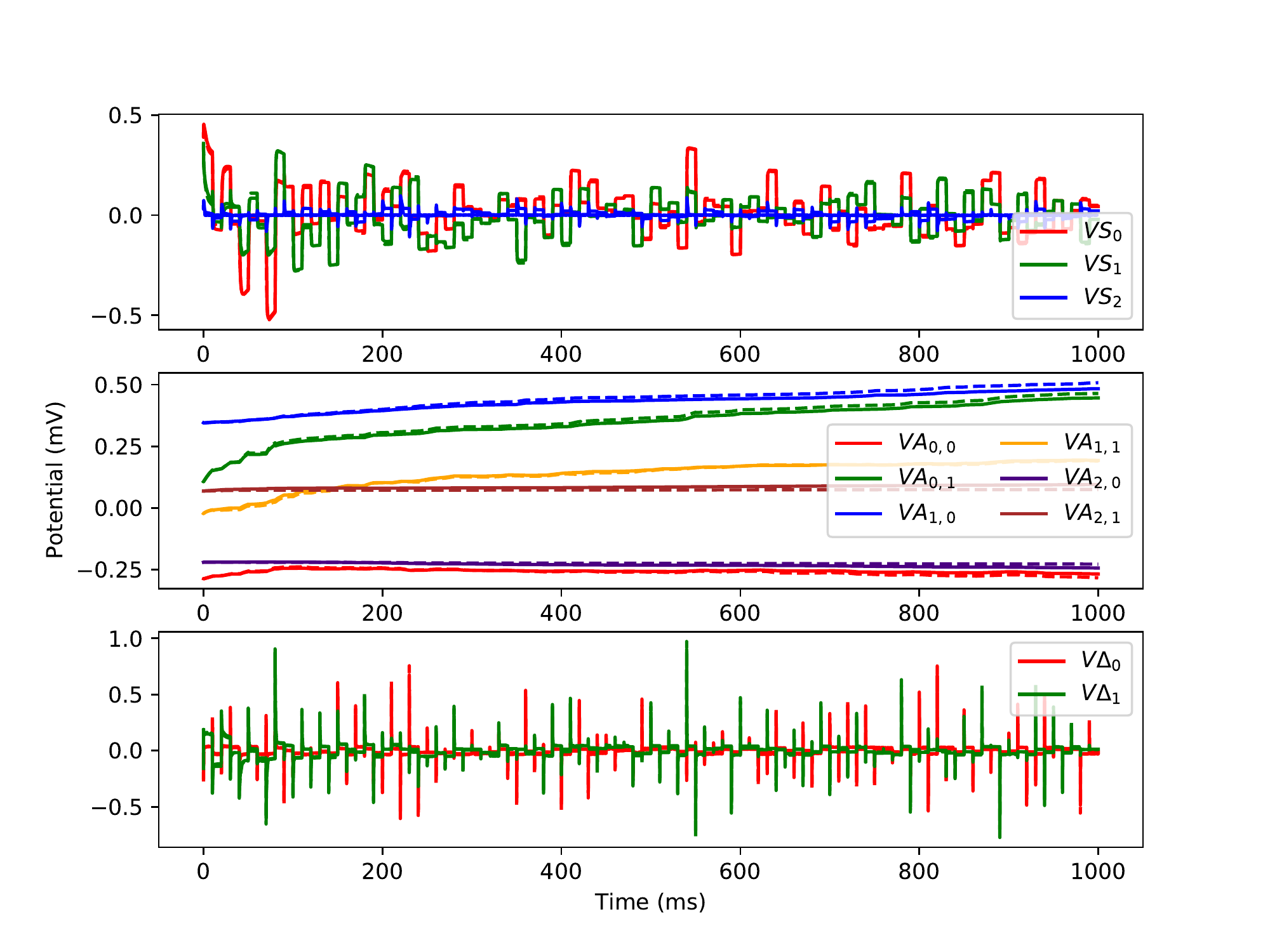}
    }\\
    \subfloat[$L_0$-LSC circuit over 100ms]{
    \label{fig:l0_sim}
    \includegraphics[width=0.5\columnwidth]{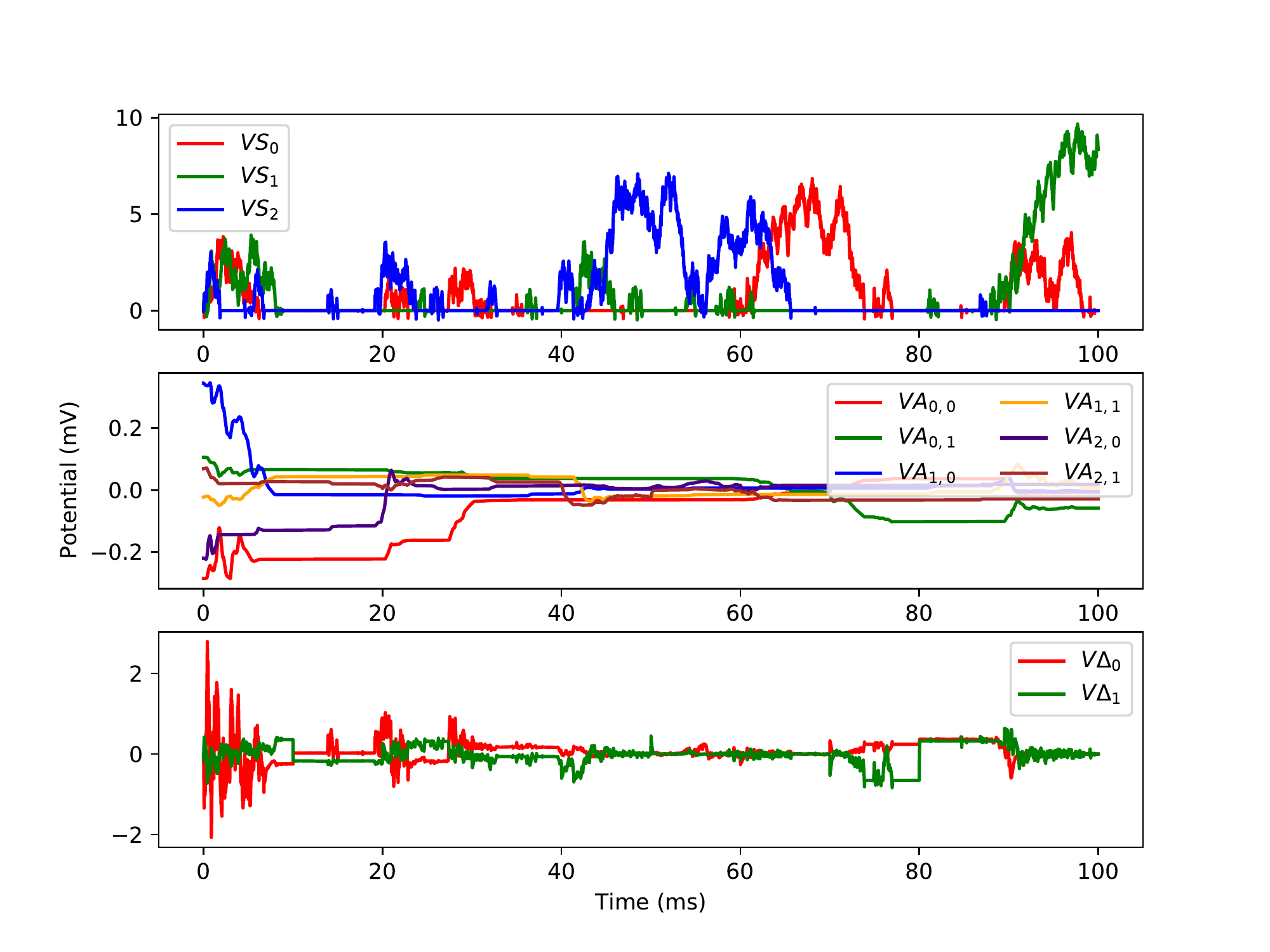}
    }
    \caption{Dynamics of potentials representing $\mb s$, $A$ and $\Delta$ in two analog circuits implementing sparse dictionary learning. The dotted lines depict simulated values and the solid lines are theoretical solutions. (Top) dynamics of latent variable coefficients. (Middle) dynamics of dictionary elements. (Bottom) dynamics of reconstruction error. a) The dynamics shown for a $L_1$-sparse \ctsc circuit. b) The same potentials in the circuit plotted for a longer interval of time where the evolution of $A$ is more apparent. c) The dynamics shown for a $L_0$-LSC circuit}
\end{figure}

With matrix multiplication accounted for, we return to the sparse penalty. Specifically, to implement the sign function in (Eq. \ref{eq:hw_deq_1}), a high gain open loop op-amp is used as an comparator (App. \ref{sec:circuit_l1}). The entire analog circuitry was drafted and simulated in LTSpice. The results are compared against solutions to (Eq. \ref{eq:ctsc_diffeq} - \ref{eq:ctsc_diffeq_2}) and shown in Fig. \ref{fig:l1_short}, \ref{fig:l1_long}.

Over the course of one second, 100 inputs $V^{(X)}$ were presented in 10ms intervals. We highlight the different response from different nodes in the circuit. The faster evolving coefficients $V^{(S)}$ converge for each input within the short 10ms window. The slower evolving dictionary elements $V^{(A)}$ exhibits slower and smoother dynamics. Finally, the reconstruction error $V^{(\Delta)}$ spikes with each presentation of new input and tends towards zero. We see that the simulated circuit dynamics closely follows the theoretical solutions both on short time scales and long time scales.

\subsection{$L_0$-LSC Circuit}
Lastly, we present the design of an analog circuit to implement $L_0$-LSC. Its dynamics are modeled by Eqs. \ref{eq:l0lsc_1}, \ref{eq:l0lsc_2}. Two major changes are required from the above $L_1$-\ctsc design. First is the inclusion of a soft-threshold function and second, the injection of white noise. Details of both can be found in App. \ref{sec:circuit_l0}.

Results from the Spice simulations are shown in Fig. \ref{fig:l0_sim}. Similar to $L1$-\ctsc, the circuit is characterized by two populations of fast evolving nodes $V^{(s)}_i$ and slow evolving nodes $V^{(A)}_{ij}$. 

While in the Spice simulation, white noise was directly added to the circuit, the ultimate aim is to leverage unavoidable, inherent noise of the circuit. An important direction for further research is the detailed characterization of noise from various electronic components. It has been demonstrated, in the context of photonic networks \cite{roques2019photonic}, systems leveraging non-Gaussian sources of noise can converge to the same distribution as those with white noise.

\section{Analog Circuit}
\label{app:circuit}
\begin{figure}
    \centering
    \includegraphics[width=.7\columnwidth]{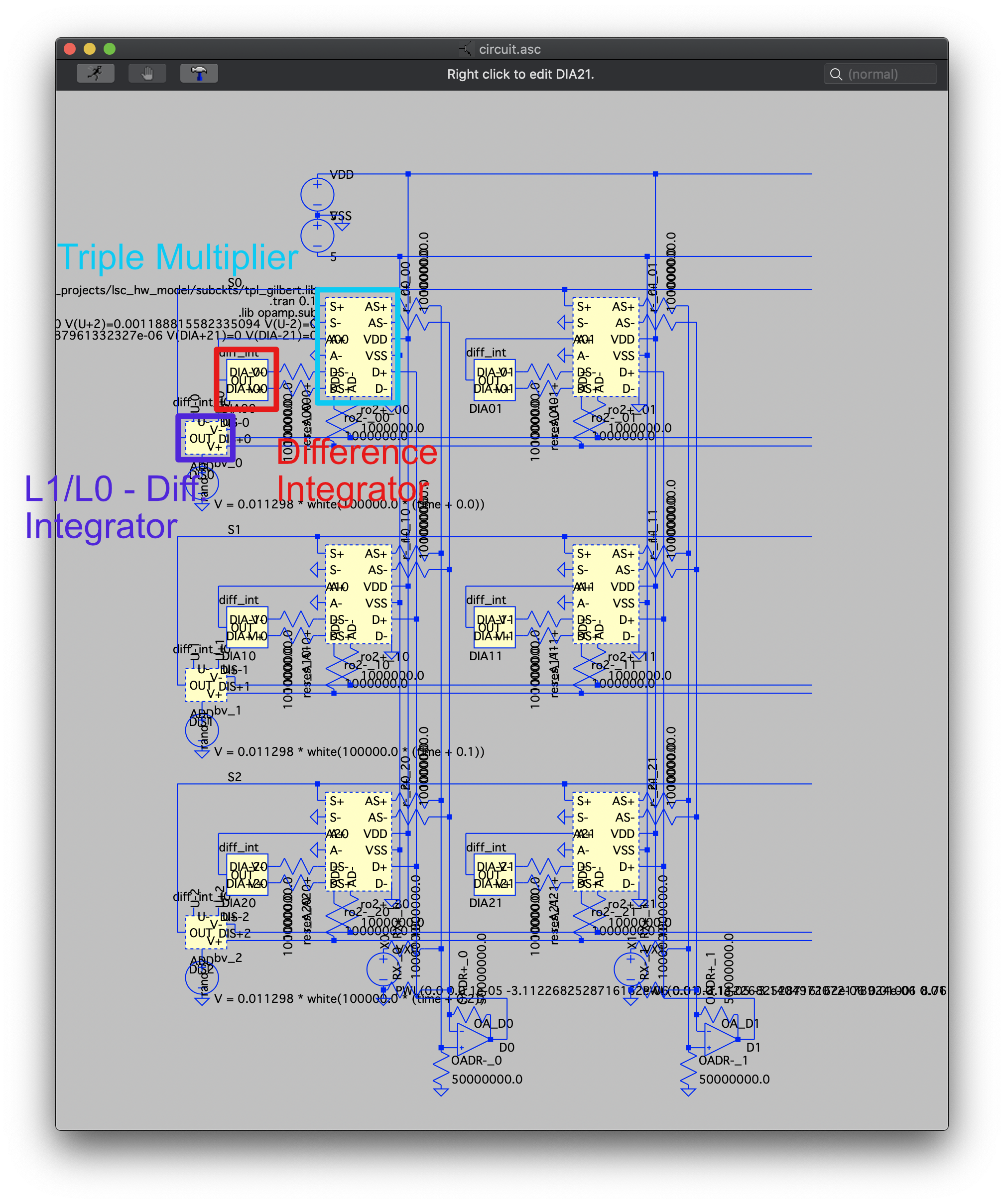}
    \caption{Circuit block diagram used in the LTSpice simulation.}
    \label{fig:lt_spice_sim}
\end{figure}

The circuitry used in the LTSpice simulation is shown in Fig. \ref{fig:lt_spice_sim}. One of the triple multipliers is highlighted by the blue box. A integrator is represented by the circuit block in the red box.

Note that unlike the simplified circuit diagram shown in Fig. \ref{fig:full_loop}, the multipliers act on differential voltages and also outputs two potentials $V_-$, $V_+$. The triple multiplier, therefore takes in six inputs and produces six outputs. Because of this, the integrator operates on the difference in voltages and contains an subtracting op-amp before the actual integrator. In practices, this is achieved through a pair of cascading opamp circuits. The first, a differential amplifier, performs a subtraction and the second, an integrator amplifier continues with integration. This differential integrator, in purple, is modified in implementing L0- and L1-sparse penalties.
\subsection{$L_1$-Sparse}
\label{sec:circuit_l1}
\begin{figure}[ht!]
    \centering
    \subfloat[Difference integrator with L1 penalty]{
    \label{fig:l1_di}
    \includegraphics[height=2in]{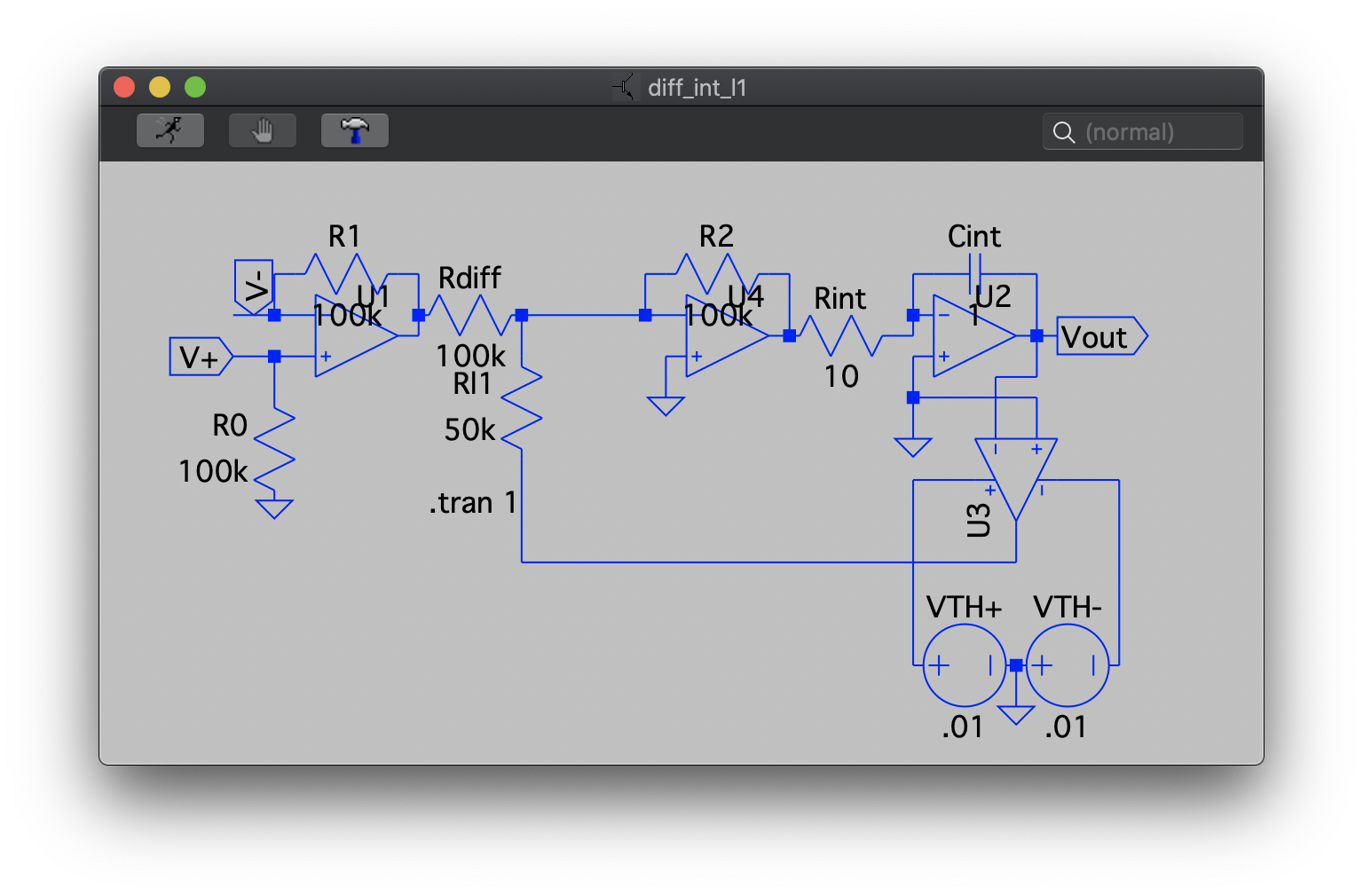}
    }
    \subfloat[Difference integrator with L0 penalty]{
    \label{fig:l0_di}
    \includegraphics[height=2in]{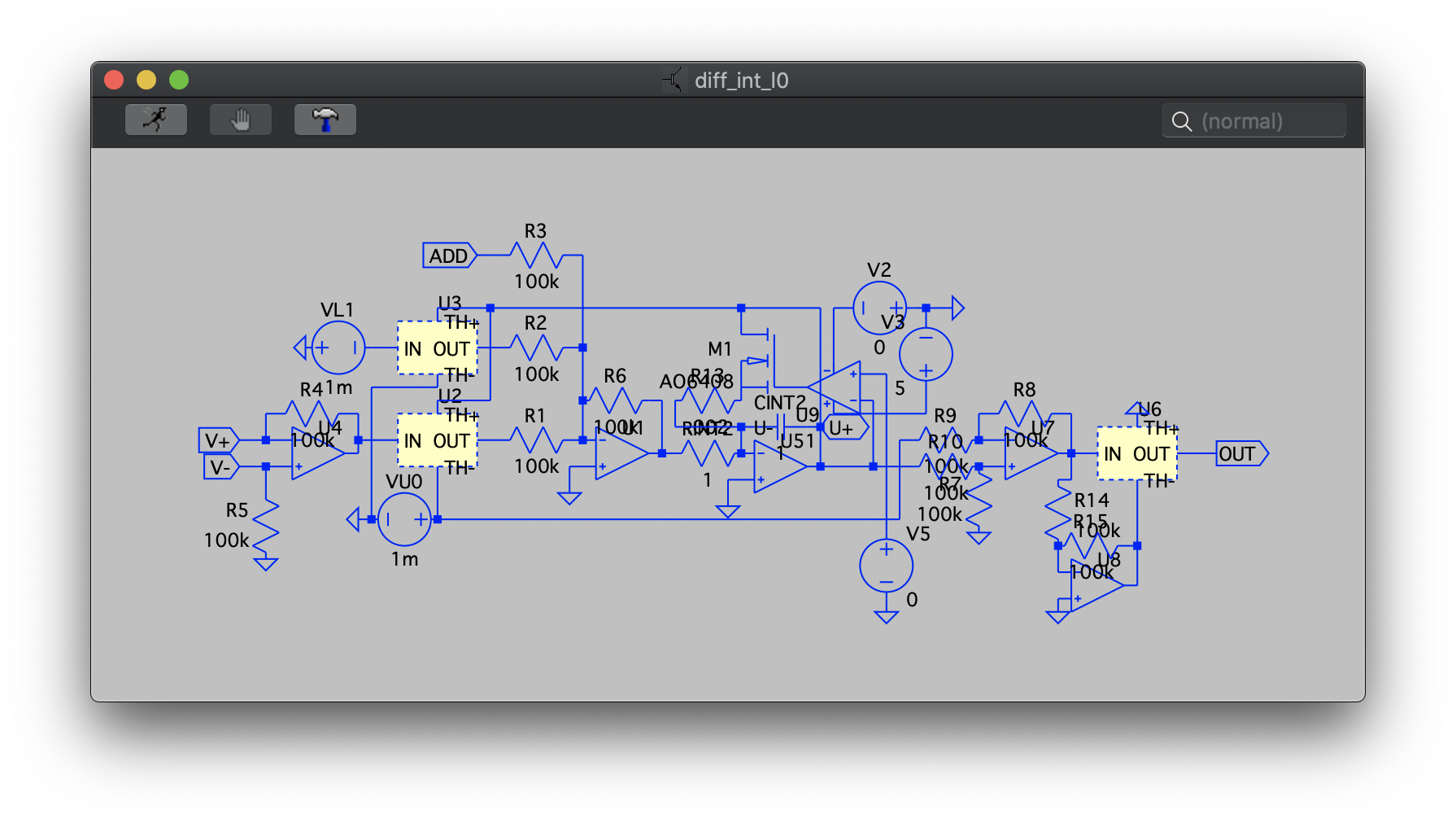}
    }
    \caption{The circuit diagram for both L1 and L0-sparse differential ingegrators. a) L1 sparse pentalty is achieved by taking the sign function of the output and subtracting the initial input. Specifically, the sign function is implemented using an op-amp. b) The L0-sparse penalty is achieved mainly by passing the output through a threshold function ($s = f(u)$).}
\end{figure}
With the L1-penalty, Eq. \ref{eq:d_As} - \ref{eq:A_ds} becomes
\begin{align}
    \Delta &= A\mb s - \mb x \\
    \mb s &= -\tau_s^{-1} \int dt~ A^T \Delta\\
    A &= - \tau_A^{-1} \int dt~ \left(\Delta \mb s^T + \lambda_1 \text{sgn}(\mb{s})\right)
\end{align}
with the addition of a $\text{sgn}(\mb s)$ term. The differential integrator circuit is modified accordingly. An open loop op-amp behaves as the sign function. The output is then summed with $\Delta \mb s^T$ and subsequently integrated. The strength of the L1 penalty $\lambda_1$ can be easily adjusted with the summation being a weighted sum. 
\subsection{$L_0$-Sparse Circuit}
\label{sec:circuit_l0}
The L0-sparse set of equations is
\begin{align}
    \Delta &= A\mb s - \mb x \\
    \mb u &= -\tau_s^{-1} \int dt~ A^T \Delta\\
    A &= - \tau_A^{-1} \int dt~ \left(\Delta \mb s^T + \lambda_1 \text{sgn}(\mb{u})\right).
\end{align}
Recall that the new internal variable $\mb u$ and $\mb s$ are related via the threshold function $s_i = f(u_i)$ (Eq. \ref{eq:threshold_fn}). In keeping with the conventions of the simulations, we restrict $u_i$ to be positive, making the sign function unnecessary. The circuit shown above details the implementation of the threshold function. The potential $u_0$ is subtracted from $u_i$ and then a ReLU circuit\cite{agarap2018deep} is applied making use of the equivalent expression
\begin{align}
    f(u_i) = \max(0, u_i - u_0)
\end{align}
in cases where $u_i \ge 0$. Finally, to maintain $u_i \ge 0$ despite the white noise, a transistor switch is added to short the capacitor of the integrator amplifier should $u_i < 0$.
\section{Batched Circuit}
\label{app:batch}
\begin{figure}
    \centering
    \includegraphics[width=0.7\columnwidth]{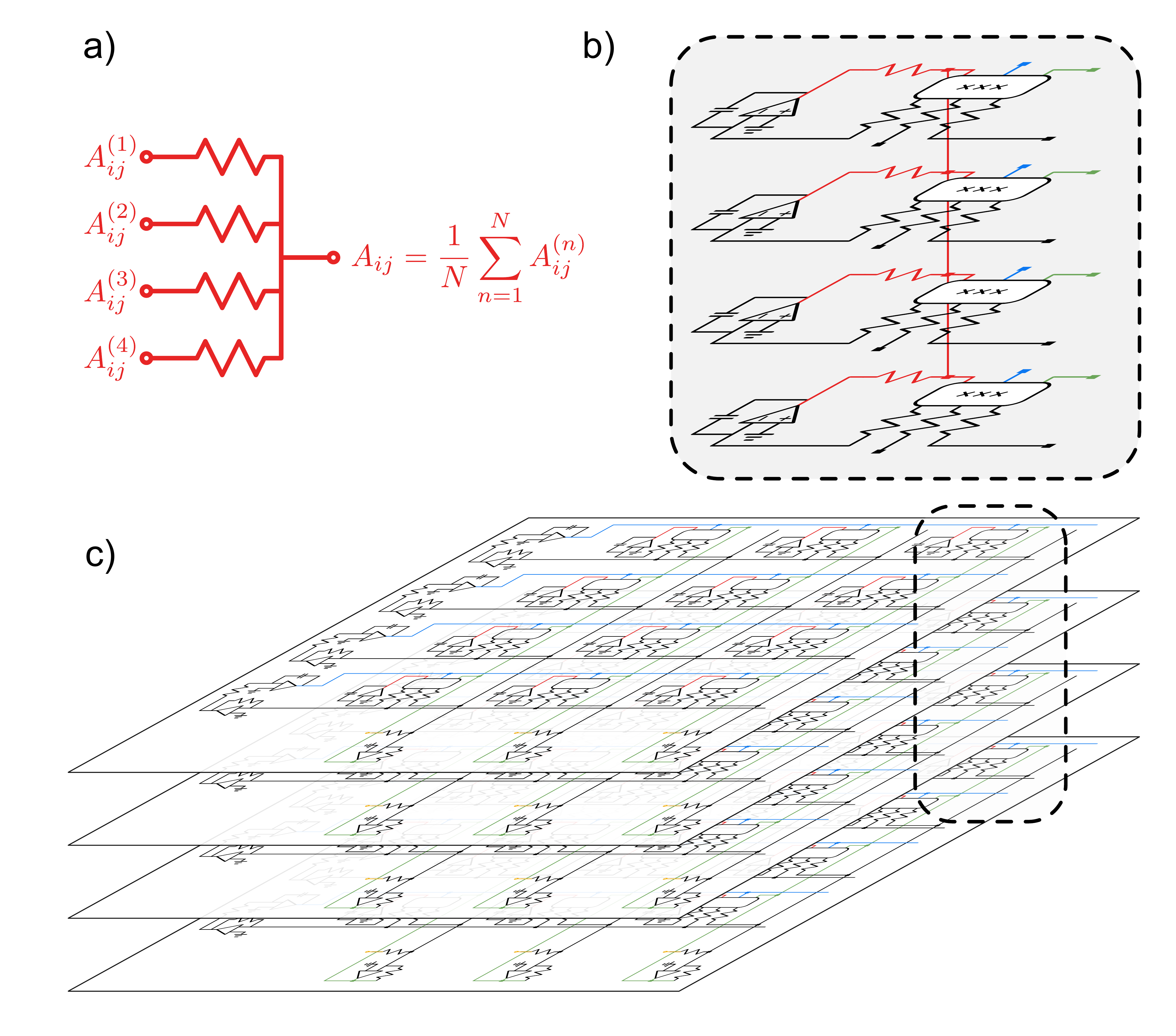}
    \caption{Circuit for batched update. a) A very simple passive averager is shown, where the central node has a potential equaling the mean of all the ``inputs''. b) Incorporating this passive averager into a column of mulitpliers. c) A 3D circuit comprised of layers of the LSC circuit. Each layer corresponds to an individual input of a batch with its unique data $\mb x_n$ and coefficients $\mb u_n$ but with shared dictionary elements $A$.}
    \label{fig:batch_circuit}
\end{figure}

In this section, we describe a simple modification to generalize the current circuit to incorporate batched inputs. The dynamics of \ctsc with input of batch size $N$ can be written as
\begin{align}
    \tau_u \mb{\dot u_n} &= \nabla_u E(A, \mb u_n, \mb x_n)\\
    \tau_A \dot A &= \frac1N \sum_{n=1}^N \nabla_A E(A, \mb u_n, \mb x_n).
\end{align}
We can implement the above dynamics by stacking $N$ identical LSC circuits (see Fig. \ref{fig:batch_circuit}). Each layer would have its own data $\mb x_n$ and latent variables $\mb u_n$ and be updated independently. However, the dictionary elements $A$ is shared among all layers and its update is averaged from each layer. A simple passive averaging circuit, shown in the figure below, can be used to link each of the layers.

\end{subappendices}

\newpage
\bibliographystyle{apacite}
\bibliography{refs}

\begin{thebibliography}{}

\bibitem [\protect \citeauthoryear {%
Ackley%
, Hinton%
\BCBL {}\ \BBA {} Sejnowski%
}{%
Ackley%
\ \protect \BOthers {.}}{%
{\protect \APACyear {1985}}%
}]{%
ackley1985learning}
\APACinsertmetastar {%
ackley1985learning}%
\begin{APACrefauthors}%
Ackley, D\BPBI H.%
, Hinton, G\BPBI E.%
\BCBL {}\ \BBA {} Sejnowski, T\BPBI J.%
\end{APACrefauthors}%
\unskip\
\newblock
\APACrefYearMonthDay{1985}{}{}.
\newblock
{\BBOQ}\APACrefatitle {A learning algorithm for Boltzmann machines} {A learning
  algorithm for boltzmann machines}.{\BBCQ}
\newblock
\APACjournalVolNumPages{Cognitive science}{9}{1}{147--169}.
\PrintBackRefs{\CurrentBib}

\bibitem [\protect \citeauthoryear {%
Agarap%
}{%
Agarap%
}{%
{\protect \APACyear {2018}}%
}]{%
agarap2018deep}
\APACinsertmetastar {%
agarap2018deep}%
\begin{APACrefauthors}%
Agarap, A\BPBI F.%
\end{APACrefauthors}%
\unskip\
\newblock
\APACrefYearMonthDay{2018}{}{}.
\newblock
{\BBOQ}\APACrefatitle {Deep learning using rectified linear units (relu)} {Deep
  learning using rectified linear units (relu)}.{\BBCQ}
\newblock
\APACjournalVolNumPages{arXiv preprint arXiv:1803.08375}{}{}{}.
\PrintBackRefs{\CurrentBib}

\bibitem [\protect \citeauthoryear {%
Beckett%
\ \protect \BOthers {.}}{%
Beckett%
\ \protect \BOthers {.}}{%
{\protect \APACyear {2014}}%
}]{%
beckett2014zero}
\APACinsertmetastar {%
beckett2014zero}%
\begin{APACrefauthors}%
Beckett, S.%
, Jee, J.%
, Ncube, T.%
, Pompilus, S.%
, Washington, Q.%
, Singh, A.%
\BCBL {}\ \BBA {} Pal, N.%
\end{APACrefauthors}%
\unskip\
\newblock
\APACrefYearMonthDay{2014}{}{}.
\newblock
{\BBOQ}\APACrefatitle {Zero-inflated Poisson (ZIP) distribution: Parameter
  estimation and applications to model data from natural calamities}
  {Zero-inflated poisson (zip) distribution: Parameter estimation and
  applications to model data from natural calamities}.{\BBCQ}
\newblock
\APACjournalVolNumPages{Involve, a Journal of Mathematics}{7}{6}{751--767}.
\PrintBackRefs{\CurrentBib}

\bibitem [\protect \citeauthoryear {%
Berkes%
, Orb{\'a}n%
, Lengyel%
\BCBL {}\ \BBA {} Fiser%
}{%
Berkes%
\ \protect \BOthers {.}}{%
{\protect \APACyear {2011}}%
}]{%
berkes2011spontaneous}
\APACinsertmetastar {%
berkes2011spontaneous}%
\begin{APACrefauthors}%
Berkes, P.%
, Orb{\'a}n, G.%
, Lengyel, M.%
\BCBL {}\ \BBA {} Fiser, J.%
\end{APACrefauthors}%
\unskip\
\newblock
\APACrefYearMonthDay{2011}{}{}.
\newblock
{\BBOQ}\APACrefatitle {Spontaneous cortical activity reveals hallmarks of an
  optimal internal model of the environment} {Spontaneous cortical activity
  reveals hallmarks of an optimal internal model of the environment}.{\BBCQ}
\newblock
\APACjournalVolNumPages{Science}{331}{6013}{83--87}.
\PrintBackRefs{\CurrentBib}

\bibitem [\protect \citeauthoryear {%
Berkes%
, Turner%
\BCBL {}\ \BBA {} Sahani%
}{%
Berkes%
\ \protect \BOthers {.}}{%
{\protect \APACyear {2008}}%
}]{%
berkes2008sparsity}
\APACinsertmetastar {%
berkes2008sparsity}%
\begin{APACrefauthors}%
Berkes, P.%
, Turner, R.%
\BCBL {}\ \BBA {} Sahani, M.%
\end{APACrefauthors}%
\unskip\
\newblock
\APACrefYearMonthDay{2008}{}{}.
\newblock
{\BBOQ}\APACrefatitle {On sparsity and overcompleteness in image models} {On
  sparsity and overcompleteness in image models}.{\BBCQ}
\newblock
\BIn{} \APACrefbtitle {Advances in neural information processing systems}
  {Advances in neural information processing systems}\ (\BPGS\ 89--96).
\PrintBackRefs{\CurrentBib}

\bibitem [\protect \citeauthoryear {%
Borders%
\ \protect \BOthers {.}}{%
Borders%
\ \protect \BOthers {.}}{%
{\protect \APACyear {2019}}%
}]{%
borders2019integer}
\APACinsertmetastar {%
borders2019integer}%
\begin{APACrefauthors}%
Borders, W\BPBI A.%
, Pervaiz, A\BPBI Z.%
, Fukami, S.%
, Camsari, K\BPBI Y.%
, Ohno, H.%
\BCBL {}\ \BBA {} Datta, S.%
\end{APACrefauthors}%
\unskip\
\newblock
\APACrefYearMonthDay{2019}{}{}.
\newblock
{\BBOQ}\APACrefatitle {Integer factorization using stochastic magnetic tunnel
  junctions} {Integer factorization using stochastic magnetic tunnel
  junctions}.{\BBCQ}
\newblock
\APACjournalVolNumPages{Nature}{573}{7774}{390--393}.
\PrintBackRefs{\CurrentBib}

\bibitem [\protect \citeauthoryear {%
Boutin%
, Franciosini%
, Ruffier%
\BCBL {}\ \BBA {} Perrinet%
}{%
Boutin%
\ \protect \BOthers {.}}{%
{\protect \APACyear {2020}}%
}]{%
boutin2020effect}
\APACinsertmetastar {%
boutin2020effect}%
\begin{APACrefauthors}%
Boutin, V.%
, Franciosini, A.%
, Ruffier, F.%
\BCBL {}\ \BBA {} Perrinet, L.%
\end{APACrefauthors}%
\unskip\
\newblock
\APACrefYearMonthDay{2020}{}{}.
\newblock
{\BBOQ}\APACrefatitle {Effect of top-down connections in Hierarchical Sparse
  Coding} {Effect of top-down connections in hierarchical sparse
  coding}.{\BBCQ}
\newblock
\APACjournalVolNumPages{Neural Computation}{32}{11}{2279--2309}.
\PrintBackRefs{\CurrentBib}

\bibitem [\protect \citeauthoryear {%
Bussi%
\ \BBA {} Parrinello%
}{%
Bussi%
\ \BBA {} Parrinello%
}{%
{\protect \APACyear {2007}}%
}]{%
bussi2007accurate}
\APACinsertmetastar {%
bussi2007accurate}%
\begin{APACrefauthors}%
Bussi, G.%
\BCBT {}\ \BBA {} Parrinello, M.%
\end{APACrefauthors}%
\unskip\
\newblock
\APACrefYearMonthDay{2007}{}{}.
\newblock
{\BBOQ}\APACrefatitle {Accurate sampling using Langevin dynamics} {Accurate
  sampling using langevin dynamics}.{\BBCQ}
\newblock
\APACjournalVolNumPages{Physical Review E}{75}{5}{056707}.
\PrintBackRefs{\CurrentBib}

\bibitem [\protect \citeauthoryear {%
Cheng%
, Chatterji%
, Bartlett%
\BCBL {}\ \BBA {} Jordan%
}{%
Cheng%
\ \protect \BOthers {.}}{%
{\protect \APACyear {2018}}%
}]{%
cheng2018underdamped}
\APACinsertmetastar {%
cheng2018underdamped}%
\begin{APACrefauthors}%
Cheng, X.%
, Chatterji, N\BPBI S.%
, Bartlett, P\BPBI L.%
\BCBL {}\ \BBA {} Jordan, M\BPBI I.%
\end{APACrefauthors}%
\unskip\
\newblock
\APACrefYearMonthDay{2018}{}{}.
\newblock
{\BBOQ}\APACrefatitle {Underdamped Langevin MCMC: A non-asymptotic analysis}
  {Underdamped langevin mcmc: A non-asymptotic analysis}.{\BBCQ}
\newblock
\BIn{} \APACrefbtitle {Conference on Learning Theory} {Conference on learning
  theory}\ (\BPGS\ 300--323).
\PrintBackRefs{\CurrentBib}

\bibitem [\protect \citeauthoryear {%
Davies%
\ \protect \BOthers {.}}{%
Davies%
\ \protect \BOthers {.}}{%
{\protect \APACyear {2021}}%
}]{%
davies2021advancing}
\APACinsertmetastar {%
davies2021advancing}%
\begin{APACrefauthors}%
Davies, M.%
, Wild, A.%
, Orchard, G.%
, Sandamirskaya, Y.%
, Guerra, G\BPBI A\BPBI F.%
, Joshi, P.%
\BDBL {}Risbud, S\BPBI R.%
\end{APACrefauthors}%
\unskip\
\newblock
\APACrefYearMonthDay{2021}{}{}.
\newblock
{\BBOQ}\APACrefatitle {Advancing neuromorphic computing with Loihi: A survey of
  results and outlook} {Advancing neuromorphic computing with loihi: A survey
  of results and outlook}.{\BBCQ}
\newblock
\APACjournalVolNumPages{Proceedings of the IEEE}{109}{5}{911--934}.
\PrintBackRefs{\CurrentBib}

\bibitem [\protect \citeauthoryear {%
Di~Ventra%
, Pershin%
\BCBL {}\ \BBA {} Chua%
}{%
Di~Ventra%
\ \protect \BOthers {.}}{%
{\protect \APACyear {2009}}%
}]{%
di2009circuit}
\APACinsertmetastar {%
di2009circuit}%
\begin{APACrefauthors}%
Di~Ventra, M.%
, Pershin, Y\BPBI V.%
\BCBL {}\ \BBA {} Chua, L\BPBI O.%
\end{APACrefauthors}%
\unskip\
\newblock
\APACrefYearMonthDay{2009}{}{}.
\newblock
{\BBOQ}\APACrefatitle {Circuit elements with memory: memristors, memcapacitors,
  and meminductors} {Circuit elements with memory: memristors, memcapacitors,
  and meminductors}.{\BBCQ}
\newblock
\APACjournalVolNumPages{Proceedings of the IEEE}{97}{10}{1717--1724}.
\PrintBackRefs{\CurrentBib}

\bibitem [\protect \citeauthoryear {%
Donoho%
}{%
Donoho%
}{%
{\protect \APACyear {2006}}%
}]{%
donoho2006compressed}
\APACinsertmetastar {%
donoho2006compressed}%
\begin{APACrefauthors}%
Donoho, D\BPBI L.%
\end{APACrefauthors}%
\unskip\
\newblock
\APACrefYearMonthDay{2006}{}{}.
\newblock
{\BBOQ}\APACrefatitle {Compressed sensing} {Compressed sensing}.{\BBCQ}
\newblock
\APACjournalVolNumPages{IEEE Transactions on information
  theory}{52}{4}{1289--1306}.
\PrintBackRefs{\CurrentBib}

\bibitem [\protect \citeauthoryear {%
Echeveste%
, Aitchison%
, Hennequin%
\BCBL {}\ \BBA {} Lengyel%
}{%
Echeveste%
\ \protect \BOthers {.}}{%
{\protect \APACyear {2020}}%
}]{%
echeveste2020cortical}
\APACinsertmetastar {%
echeveste2020cortical}%
\begin{APACrefauthors}%
Echeveste, R.%
, Aitchison, L.%
, Hennequin, G.%
\BCBL {}\ \BBA {} Lengyel, M.%
\end{APACrefauthors}%
\unskip\
\newblock
\APACrefYearMonthDay{2020}{}{}.
\newblock
{\BBOQ}\APACrefatitle {Cortical-like dynamics in recurrent circuits optimized
  for sampling-based probabilistic inference} {Cortical-like dynamics in
  recurrent circuits optimized for sampling-based probabilistic
  inference}.{\BBCQ}
\newblock
\APACjournalVolNumPages{Nature neuroscience}{23}{9}{1138--1149}.
\PrintBackRefs{\CurrentBib}

\bibitem [\protect \citeauthoryear {%
Garrigues%
\ \BBA {} Olshausen%
}{%
Garrigues%
\ \BBA {} Olshausen%
}{%
{\protect \APACyear {2007}}%
}]{%
garrigues2007learning}
\APACinsertmetastar {%
garrigues2007learning}%
\begin{APACrefauthors}%
Garrigues, P.%
\BCBT {}\ \BBA {} Olshausen, B\BPBI A.%
\end{APACrefauthors}%
\unskip\
\newblock
\APACrefYearMonthDay{2007}{}{}.
\newblock
{\BBOQ}\APACrefatitle {Learning horizontal connections in a sparse coding model
  of natural images.} {Learning horizontal connections in a sparse coding model
  of natural images.}{\BBCQ}
\newblock
\BIn{} \APACrefbtitle {Nips} {Nips}\ (\BPGS\ 505--512).
\PrintBackRefs{\CurrentBib}

\bibitem [\protect \citeauthoryear {%
Garrigues%
\ \BBA {} Olshausen%
}{%
Garrigues%
\ \BBA {} Olshausen%
}{%
{\protect \APACyear {2010}}%
}]{%
garrigues2010group}
\APACinsertmetastar {%
garrigues2010group}%
\begin{APACrefauthors}%
Garrigues, P.%
\BCBT {}\ \BBA {} Olshausen, B\BPBI A.%
\end{APACrefauthors}%
\unskip\
\newblock
\APACrefYearMonthDay{2010}{}{}.
\newblock
{\BBOQ}\APACrefatitle {Group sparse coding with a laplacian scale mixture
  prior} {Group sparse coding with a laplacian scale mixture prior}.{\BBCQ}
\newblock
\BIn{} \APACrefbtitle {Advances in neural information processing systems}
  {Advances in neural information processing systems}\ (\BPGS\ 676--684).
\PrintBackRefs{\CurrentBib}

\bibitem [\protect \citeauthoryear {%
Gilbert%
}{%
Gilbert%
}{%
{\protect \APACyear {1968}}%
}]{%
gilbert1968precise}
\APACinsertmetastar {%
gilbert1968precise}%
\begin{APACrefauthors}%
Gilbert, B.%
\end{APACrefauthors}%
\unskip\
\newblock
\APACrefYearMonthDay{1968}{}{}.
\newblock
{\BBOQ}\APACrefatitle {A precise four-quadrant multiplier with subnanosecond
  response} {A precise four-quadrant multiplier with subnanosecond
  response}.{\BBCQ}
\newblock
\APACjournalVolNumPages{IEEE journal of solid-state circuits}{3}{4}{365--373}.
\PrintBackRefs{\CurrentBib}

\bibitem [\protect \citeauthoryear {%
Hastie%
, Tibshirani%
\BCBL {}\ \BBA {} Friedman%
}{%
Hastie%
\ \protect \BOthers {.}}{%
{\protect \APACyear {2009}}%
}]{%
hastie2009elements}
\APACinsertmetastar {%
hastie2009elements}%
\begin{APACrefauthors}%
Hastie, T.%
, Tibshirani, R.%
\BCBL {}\ \BBA {} Friedman, J.%
\end{APACrefauthors}%
\unskip\
\newblock
\APACrefYear{2009}.
\newblock
\APACrefbtitle {The elements of statistical learning: data mining, inference,
  and prediction} {The elements of statistical learning: data mining,
  inference, and prediction}.
\newblock
\APACaddressPublisher{}{Springer Science \& Business Media}.
\PrintBackRefs{\CurrentBib}

\bibitem [\protect \citeauthoryear {%
Hateren%
\ \BBA {} Schaaf%
}{%
Hateren%
\ \BBA {} Schaaf%
}{%
{\protect \APACyear {1998}}%
}]{%
hateren_schaaf_1998}
\APACinsertmetastar {%
hateren_schaaf_1998}%
\begin{APACrefauthors}%
Hateren, J\BPBI H\BPBI v.%
\BCBT {}\ \BBA {} Schaaf, A\BPBI v\BPBI d.%
\end{APACrefauthors}%
\unskip\
\newblock
\APACrefYearMonthDay{1998}{Mar}{}.
\newblock
{\BBOQ}\APACrefatitle {Independent Component Filters of Natural Images Compared
  with Simple Cells in Primary Visual Cortex} {Independent component filters of
  natural images compared with simple cells in primary visual cortex}.{\BBCQ}
\newblock
\APACjournalVolNumPages{Proceedings: Biological Sciences}{265}{1394}{359-366}.
\PrintBackRefs{\CurrentBib}

\bibitem [\protect \citeauthoryear {%
Hennequin%
, Aitchison%
\BCBL {}\ \BBA {} Lengyel%
}{%
Hennequin%
\ \protect \BOthers {.}}{%
{\protect \APACyear {2014}}%
}]{%
hennequin2014fast}
\APACinsertmetastar {%
hennequin2014fast}%
\begin{APACrefauthors}%
Hennequin, G.%
, Aitchison, L.%
\BCBL {}\ \BBA {} Lengyel, M.%
\end{APACrefauthors}%
\unskip\
\newblock
\APACrefYearMonthDay{2014}{}{}.
\newblock
{\BBOQ}\APACrefatitle {Fast sampling-based inference in balanced neuronal
  networks} {Fast sampling-based inference in balanced neuronal
  networks}.{\BBCQ}
\newblock
\APACjournalVolNumPages{Advances in neural information processing
  systems}{27}{}{}.
\PrintBackRefs{\CurrentBib}

\bibitem [\protect \citeauthoryear {%
Hinton%
\ \BBA {} Salakhutdinov%
}{%
Hinton%
\ \BBA {} Salakhutdinov%
}{%
{\protect \APACyear {2006}}%
}]{%
hinton2006reducing}
\APACinsertmetastar {%
hinton2006reducing}%
\begin{APACrefauthors}%
Hinton, G\BPBI E.%
\BCBT {}\ \BBA {} Salakhutdinov, R\BPBI R.%
\end{APACrefauthors}%
\unskip\
\newblock
\APACrefYearMonthDay{2006}{}{}.
\newblock
{\BBOQ}\APACrefatitle {Reducing the dimensionality of data with neural
  networks} {Reducing the dimensionality of data with neural networks}.{\BBCQ}
\newblock
\APACjournalVolNumPages{science}{313}{5786}{504--507}.
\PrintBackRefs{\CurrentBib}

\bibitem [\protect \citeauthoryear {%
Hinton%
\ \BBA {} Sejnowski%
}{%
Hinton%
\ \BBA {} Sejnowski%
}{%
{\protect \APACyear {1983}}%
}]{%
hinton1983optimal}
\APACinsertmetastar {%
hinton1983optimal}%
\begin{APACrefauthors}%
Hinton, G\BPBI E.%
\BCBT {}\ \BBA {} Sejnowski, T\BPBI J.%
\end{APACrefauthors}%
\unskip\
\newblock
\APACrefYearMonthDay{1983}{}{}.
\newblock
{\BBOQ}\APACrefatitle {Optimal perceptual inference} {Optimal perceptual
  inference}.{\BBCQ}
\newblock
\BIn{} \APACrefbtitle {Proceedings of the IEEE conference on Computer Vision
  and Pattern Recognition} {Proceedings of the ieee conference on computer
  vision and pattern recognition}\ (\BVOL~448).
\PrintBackRefs{\CurrentBib}

\bibitem [\protect \citeauthoryear {%
Hoyer%
\ \BBA {} Hyv{\"a}rinen%
}{%
Hoyer%
\ \BBA {} Hyv{\"a}rinen%
}{%
{\protect \APACyear {2003}}%
}]{%
hoyer2003interpreting}
\APACinsertmetastar {%
hoyer2003interpreting}%
\begin{APACrefauthors}%
Hoyer, P\BPBI O.%
\BCBT {}\ \BBA {} Hyv{\"a}rinen, A.%
\end{APACrefauthors}%
\unskip\
\newblock
\APACrefYearMonthDay{2003}{}{}.
\newblock
{\BBOQ}\APACrefatitle {Interpreting neural response variability as Monte Carlo
  sampling of the posterior} {Interpreting neural response variability as monte
  carlo sampling of the posterior}.{\BBCQ}
\newblock
\BIn{} \APACrefbtitle {Advances in neural information processing systems}
  {Advances in neural information processing systems}\ (\BPGS\ 293--300).
\PrintBackRefs{\CurrentBib}

\bibitem [\protect \citeauthoryear {%
Kingma%
\ \BBA {} Welling%
}{%
Kingma%
\ \BBA {} Welling%
}{%
{\protect \APACyear {2013}}%
}]{%
kingma2013auto}
\APACinsertmetastar {%
kingma2013auto}%
\begin{APACrefauthors}%
Kingma, D\BPBI P.%
\BCBT {}\ \BBA {} Welling, M.%
\end{APACrefauthors}%
\unskip\
\newblock
\APACrefYearMonthDay{2013}{}{}.
\newblock
{\BBOQ}\APACrefatitle {Auto-encoding variational bayes} {Auto-encoding
  variational bayes}.{\BBCQ}
\newblock
\APACjournalVolNumPages{arXiv preprint arXiv:1312.6114}{}{}{}.
\PrintBackRefs{\CurrentBib}

\bibitem [\protect \citeauthoryear {%
Krestinskaya%
, Choubey%
\BCBL {}\ \BBA {} James%
}{%
Krestinskaya%
\ \protect \BOthers {.}}{%
{\protect \APACyear {2020}}%
}]{%
krestinskaya2020memristive}
\APACinsertmetastar {%
krestinskaya2020memristive}%
\begin{APACrefauthors}%
Krestinskaya, O.%
, Choubey, B.%
\BCBL {}\ \BBA {} James, A.%
\end{APACrefauthors}%
\unskip\
\newblock
\APACrefYearMonthDay{2020}{}{}.
\newblock
{\BBOQ}\APACrefatitle {Memristive GAN in Analog} {Memristive gan in
  analog}.{\BBCQ}
\newblock
\APACjournalVolNumPages{Scientific reports}{10}{1}{1--14}.
\PrintBackRefs{\CurrentBib}

\bibitem [\protect \citeauthoryear {%
Lee%
\ \BBA {} Mumford%
}{%
Lee%
\ \BBA {} Mumford%
}{%
{\protect \APACyear {2003}}%
}]{%
lee2003hierarchical}
\APACinsertmetastar {%
lee2003hierarchical}%
\begin{APACrefauthors}%
Lee, T\BPBI S.%
\BCBT {}\ \BBA {} Mumford, D.%
\end{APACrefauthors}%
\unskip\
\newblock
\APACrefYearMonthDay{2003}{}{}.
\newblock
{\BBOQ}\APACrefatitle {Hierarchical Bayesian inference in the visual cortex}
  {Hierarchical bayesian inference in the visual cortex}.{\BBCQ}
\newblock
\APACjournalVolNumPages{JOSA A}{20}{7}{1434--1448}.
\PrintBackRefs{\CurrentBib}

\bibitem [\protect \citeauthoryear {%
Lewicki%
\ \BBA {} Olshausen%
}{%
Lewicki%
\ \BBA {} Olshausen%
}{%
{\protect \APACyear {1999}}%
}]{%
lewicki1999probabilistic}
\APACinsertmetastar {%
lewicki1999probabilistic}%
\begin{APACrefauthors}%
Lewicki, M\BPBI S.%
\BCBT {}\ \BBA {} Olshausen, B\BPBI A.%
\end{APACrefauthors}%
\unskip\
\newblock
\APACrefYearMonthDay{1999}{}{}.
\newblock
{\BBOQ}\APACrefatitle {Probabilistic framework for the adaptation and
  comparison of image codes} {Probabilistic framework for the adaptation and
  comparison of image codes}.{\BBCQ}
\newblock
\APACjournalVolNumPages{JOSA A}{16}{7}{1587--1601}.
\PrintBackRefs{\CurrentBib}

\bibitem [\protect \citeauthoryear {%
Mancini%
}{%
Mancini%
}{%
{\protect \APACyear {2003}}%
}]{%
mancini2003op}
\APACinsertmetastar {%
mancini2003op}%
\begin{APACrefauthors}%
Mancini, R.%
\end{APACrefauthors}%
\unskip\
\newblock
\APACrefYear{2003}.
\newblock
\APACrefbtitle {Op amps for everyone: design reference} {Op amps for everyone:
  design reference}.
\newblock
\APACaddressPublisher{}{Newnes}.
\PrintBackRefs{\CurrentBib}

\bibitem [\protect \citeauthoryear {%
V.~Mansinghka%
\ \BBA {} Jonas%
}{%
V.~Mansinghka%
\ \BBA {} Jonas%
}{%
{\protect \APACyear {2014}}%
}]{%
mansinghka2014building}
\APACinsertmetastar {%
mansinghka2014building}%
\begin{APACrefauthors}%
Mansinghka, V.%
\BCBT {}\ \BBA {} Jonas, E.%
\end{APACrefauthors}%
\unskip\
\newblock
\APACrefYearMonthDay{2014}{}{}.
\newblock
{\BBOQ}\APACrefatitle {Building fast Bayesian computing machines out of
  intentionally stochastic, digital parts} {Building fast bayesian computing
  machines out of intentionally stochastic, digital parts}.{\BBCQ}
\newblock
\APACjournalVolNumPages{arXiv preprint arXiv:1402.4914}{}{}{}.
\PrintBackRefs{\CurrentBib}

\bibitem [\protect \citeauthoryear {%
V\BPBI K.~Mansinghka%
, Jonas%
\BCBL {}\ \BBA {} Tenenbaum%
}{%
V\BPBI K.~Mansinghka%
\ \protect \BOthers {.}}{%
{\protect \APACyear {2008}}%
}]{%
mansinghka2008stochastic}
\APACinsertmetastar {%
mansinghka2008stochastic}%
\begin{APACrefauthors}%
Mansinghka, V\BPBI K.%
, Jonas, E\BPBI M.%
\BCBL {}\ \BBA {} Tenenbaum, J\BPBI B.%
\end{APACrefauthors}%
\unskip\
\newblock
\APACrefYearMonthDay{2008}{}{}.
\newblock
{\BBOQ}\APACrefatitle {Stochastic digital circuits for probabilistic inference}
  {Stochastic digital circuits for probabilistic inference}.{\BBCQ}
\newblock
\APACjournalVolNumPages{Massachussets Institute of Technology, Technical Report
  MITCSAIL-TR}{2069}{}{}.
\PrintBackRefs{\CurrentBib}

\bibitem [\protect \citeauthoryear {%
Mead%
\ \BBA {} Mahowald%
}{%
Mead%
\ \BBA {} Mahowald%
}{%
{\protect \APACyear {1988}}%
}]{%
mead1988silicon}
\APACinsertmetastar {%
mead1988silicon}%
\begin{APACrefauthors}%
Mead, C\BPBI A.%
\BCBT {}\ \BBA {} Mahowald, M\BPBI A.%
\end{APACrefauthors}%
\unskip\
\newblock
\APACrefYearMonthDay{1988}{}{}.
\newblock
{\BBOQ}\APACrefatitle {A silicon model of early visual processing} {A silicon
  model of early visual processing}.{\BBCQ}
\newblock
\APACjournalVolNumPages{Neural networks}{1}{1}{91--97}.
\PrintBackRefs{\CurrentBib}

\bibitem [\protect \citeauthoryear {%
Mitchell%
\ \BBA {} Beauchamp%
}{%
Mitchell%
\ \BBA {} Beauchamp%
}{%
{\protect \APACyear {1988}}%
}]{%
mitchell1988bayesian}
\APACinsertmetastar {%
mitchell1988bayesian}%
\begin{APACrefauthors}%
Mitchell, T\BPBI J.%
\BCBT {}\ \BBA {} Beauchamp, J\BPBI J.%
\end{APACrefauthors}%
\unskip\
\newblock
\APACrefYearMonthDay{1988}{}{}.
\newblock
{\BBOQ}\APACrefatitle {Bayesian variable selection in linear regression}
  {Bayesian variable selection in linear regression}.{\BBCQ}
\newblock
\APACjournalVolNumPages{Journal of the american statistical
  association}{83}{404}{1023--1032}.
\PrintBackRefs{\CurrentBib}

\bibitem [\protect \citeauthoryear {%
Mou%
, Ma%
, Wainwright%
, Bartlett%
\BCBL {}\ \BBA {} Jordan%
}{%
Mou%
\ \protect \BOthers {.}}{%
{\protect \APACyear {2021}}%
}]{%
mou2021high}
\APACinsertmetastar {%
mou2021high}%
\begin{APACrefauthors}%
Mou, W.%
, Ma, Y\BHBI A.%
, Wainwright, M\BPBI J.%
, Bartlett, P\BPBI L.%
\BCBL {}\ \BBA {} Jordan, M\BPBI I.%
\end{APACrefauthors}%
\unskip\
\newblock
\APACrefYearMonthDay{2021}{}{}.
\newblock
{\BBOQ}\APACrefatitle {High-Order Langevin Diffusion Yields an Accelerated MCMC
  Algorithm.} {High-order langevin diffusion yields an accelerated mcmc
  algorithm.}{\BBCQ}
\newblock
\APACjournalVolNumPages{J. Mach. Learn. Res.}{22}{}{42--1}.
\PrintBackRefs{\CurrentBib}

\bibitem [\protect \citeauthoryear {%
Neal%
}{%
Neal%
}{%
{\protect \APACyear {2001}}%
}]{%
neal2001annealed}
\APACinsertmetastar {%
neal2001annealed}%
\begin{APACrefauthors}%
Neal, R\BPBI M.%
\end{APACrefauthors}%
\unskip\
\newblock
\APACrefYearMonthDay{2001}{}{}.
\newblock
{\BBOQ}\APACrefatitle {Annealed importance sampling} {Annealed importance
  sampling}.{\BBCQ}
\newblock
\APACjournalVolNumPages{Statistics and computing}{11}{2}{125--139}.
\PrintBackRefs{\CurrentBib}

\bibitem [\protect \citeauthoryear {%
Olshausen%
}{%
Olshausen%
}{%
{\protect \APACyear {2013}}%
}]{%
olshausen2013highly}
\APACinsertmetastar {%
olshausen2013highly}%
\begin{APACrefauthors}%
Olshausen, B\BPBI A.%
\end{APACrefauthors}%
\unskip\
\newblock
\APACrefYearMonthDay{2013}{}{}.
\newblock
{\BBOQ}\APACrefatitle {Highly overcomplete sparse coding} {Highly overcomplete
  sparse coding}.{\BBCQ}
\newblock
\BIn{} \APACrefbtitle {Human Vision and Electronic Imaging XVIII} {Human vision
  and electronic imaging xviii}\ (\BVOL\ 8651, \BPG~86510S).
\PrintBackRefs{\CurrentBib}

\bibitem [\protect \citeauthoryear {%
Olshausen%
}{%
Olshausen%
}{%
{\protect \APACyear {2014}}%
}]{%
olshausen201427}
\APACinsertmetastar {%
olshausen201427}%
\begin{APACrefauthors}%
Olshausen, B\BPBI A.%
\end{APACrefauthors}%
\unskip\
\newblock
\APACrefYearMonthDay{2014}{}{}.
\newblock
{\BBOQ}\APACrefatitle {Perception as an inference problem} {Perception as an
  inference problem}.{\BBCQ}
\newblock
\BIn{} G.~Mangun\ \BBA {} M.~Gazzaniga\ (\BEDS), \APACrefbtitle {The Cognitive
  Neurosciences} {The cognitive neurosciences}\ (\BCHAP~27).
\newblock
\APACaddressPublisher{}{MIT Press, Cambridge, MA}.
\PrintBackRefs{\CurrentBib}

\bibitem [\protect \citeauthoryear {%
Olshausen%
\ \BBA {} Field%
}{%
Olshausen%
\ \BBA {} Field%
}{%
{\protect \APACyear {1997}}%
}]{%
olshausen1997sparse}
\APACinsertmetastar {%
olshausen1997sparse}%
\begin{APACrefauthors}%
Olshausen, B\BPBI A.%
\BCBT {}\ \BBA {} Field, D\BPBI J.%
\end{APACrefauthors}%
\unskip\
\newblock
\APACrefYearMonthDay{1997}{}{}.
\newblock
{\BBOQ}\APACrefatitle {Sparse coding with an overcomplete basis set: A strategy
  employed by V1?} {Sparse coding with an overcomplete basis set: A strategy
  employed by v1?}{\BBCQ}
\newblock
\APACjournalVolNumPages{Vision research}{37}{23}{3311--3325}.
\PrintBackRefs{\CurrentBib}

\bibitem [\protect \citeauthoryear {%
Olshausen%
\ \BBA {} Millman%
}{%
Olshausen%
\ \BBA {} Millman%
}{%
{\protect \APACyear {2000}}%
}]{%
olshausen2000learning}
\APACinsertmetastar {%
olshausen2000learning}%
\begin{APACrefauthors}%
Olshausen, B\BPBI A.%
\BCBT {}\ \BBA {} Millman, K\BPBI J.%
\end{APACrefauthors}%
\unskip\
\newblock
\APACrefYearMonthDay{2000}{}{}.
\newblock
{\BBOQ}\APACrefatitle {Learning sparse codes with a mixture-of-Gaussians prior}
  {Learning sparse codes with a mixture-of-gaussians prior}.{\BBCQ}
\newblock
\BIn{} \APACrefbtitle {Advances in neural information processing systems}
  {Advances in neural information processing systems}\ (\BPGS\ 841--847).
\PrintBackRefs{\CurrentBib}

\bibitem [\protect \citeauthoryear {%
Orb{\'a}n%
, Berkes%
, Fiser%
\BCBL {}\ \BBA {} Lengyel%
}{%
Orb{\'a}n%
\ \protect \BOthers {.}}{%
{\protect \APACyear {2016}}%
}]{%
orban2016neural}
\APACinsertmetastar {%
orban2016neural}%
\begin{APACrefauthors}%
Orb{\'a}n, G.%
, Berkes, P.%
, Fiser, J.%
\BCBL {}\ \BBA {} Lengyel, M.%
\end{APACrefauthors}%
\unskip\
\newblock
\APACrefYearMonthDay{2016}{}{}.
\newblock
{\BBOQ}\APACrefatitle {Neural variability and sampling-based probabilistic
  representations in the visual cortex} {Neural variability and sampling-based
  probabilistic representations in the visual cortex}.{\BBCQ}
\newblock
\APACjournalVolNumPages{Neuron}{92}{2}{530--543}.
\PrintBackRefs{\CurrentBib}

\bibitem [\protect \citeauthoryear {%
Roques-Carmes%
\ \protect \BOthers {.}}{%
Roques-Carmes%
\ \protect \BOthers {.}}{%
{\protect \APACyear {2019}}%
}]{%
roques2019photonic}
\APACinsertmetastar {%
roques2019photonic}%
\begin{APACrefauthors}%
Roques-Carmes, C.%
, Shen, Y.%
, Zanoci, C.%
, Prabhu, M.%
, Atieh, F.%
, Jing, L.%
\BDBL {}others%
\end{APACrefauthors}%
\unskip\
\newblock
\APACrefYearMonthDay{2019}{}{}.
\newblock
{\BBOQ}\APACrefatitle {Photonic recurrent Ising sampler} {Photonic recurrent
  ising sampler}.{\BBCQ}
\newblock
\BIn{} \APACrefbtitle {CLEO: QELS\_Fundamental Science} {Cleo:
  Qels\_fundamental science}\ (\BPGS\ FTu4C--2).
\PrintBackRefs{\CurrentBib}

\bibitem [\protect \citeauthoryear {%
Rozell%
, Johnson%
, Baraniuk%
\BCBL {}\ \BBA {} Olshausen%
}{%
Rozell%
\ \protect \BOthers {.}}{%
{\protect \APACyear {2008}}%
}]{%
rozell2008sparse}
\APACinsertmetastar {%
rozell2008sparse}%
\begin{APACrefauthors}%
Rozell, C\BPBI J.%
, Johnson, D\BPBI H.%
, Baraniuk, R\BPBI G.%
\BCBL {}\ \BBA {} Olshausen, B\BPBI A.%
\end{APACrefauthors}%
\unskip\
\newblock
\APACrefYearMonthDay{2008}{}{}.
\newblock
{\BBOQ}\APACrefatitle {Sparse coding via thresholding and local competition in
  neural circuits} {Sparse coding via thresholding and local competition in
  neural circuits}.{\BBCQ}
\newblock
\APACjournalVolNumPages{Neural computation}{20}{10}{2526--2563}.
\PrintBackRefs{\CurrentBib}

\bibitem [\protect \citeauthoryear {%
Shapero%
, Charles%
, Rozell%
\BCBL {}\ \BBA {} Hasler%
}{%
Shapero%
\ \protect \BOthers {.}}{%
{\protect \APACyear {2012}}%
}]{%
shapero2012low}
\APACinsertmetastar {%
shapero2012low}%
\begin{APACrefauthors}%
Shapero, S.%
, Charles, A\BPBI S.%
, Rozell, C\BPBI J.%
\BCBL {}\ \BBA {} Hasler, P.%
\end{APACrefauthors}%
\unskip\
\newblock
\APACrefYearMonthDay{2012}{}{}.
\newblock
{\BBOQ}\APACrefatitle {Low power sparse approximation on reconfigurable analog
  hardware} {Low power sparse approximation on reconfigurable analog
  hardware}.{\BBCQ}
\newblock
\APACjournalVolNumPages{IEEE Journal on Emerging and Selected Topics in
  Circuits and Systems}{2}{3}{530--541}.
\PrintBackRefs{\CurrentBib}

\bibitem [\protect \citeauthoryear {%
Shelton%
, Sheikh%
, Berkes%
, Bornschein%
\BCBL {}\ \BBA {} L{\"u}cke%
}{%
Shelton%
\ \protect \BOthers {.}}{%
{\protect \APACyear {2011}}%
}]{%
shelton2011select}
\APACinsertmetastar {%
shelton2011select}%
\begin{APACrefauthors}%
Shelton, J\BPBI A.%
, Sheikh, A\BPBI S.%
, Berkes, P.%
, Bornschein, J.%
\BCBL {}\ \BBA {} L{\"u}cke, J.%
\end{APACrefauthors}%
\unskip\
\newblock
\APACrefYearMonthDay{2011}{}{}.
\newblock
{\BBOQ}\APACrefatitle {Select and sample-a model of efficient neural inference
  and learning} {Select and sample-a model of efficient neural inference and
  learning}.{\BBCQ}
\newblock
\BIn{} \APACrefbtitle {Advances in neural information processing systems}
  {Advances in neural information processing systems}\ (\BPGS\ 2618--2626).
\PrintBackRefs{\CurrentBib}

\bibitem [\protect \citeauthoryear {%
Sheridan%
\ \protect \BOthers {.}}{%
Sheridan%
\ \protect \BOthers {.}}{%
{\protect \APACyear {2017}}%
}]{%
sheridan2017sparse}
\APACinsertmetastar {%
sheridan2017sparse}%
\begin{APACrefauthors}%
Sheridan, P\BPBI M.%
, Cai, F.%
, Du, C.%
, Ma, W.%
, Zhang, Z.%
\BCBL {}\ \BBA {} Lu, W\BPBI D.%
\end{APACrefauthors}%
\unskip\
\newblock
\APACrefYearMonthDay{2017}{}{}.
\newblock
{\BBOQ}\APACrefatitle {Sparse coding with memristor networks} {Sparse coding
  with memristor networks}.{\BBCQ}
\newblock
\APACjournalVolNumPages{Nature nanotechnology}{12}{8}{784}.
\PrintBackRefs{\CurrentBib}

\bibitem [\protect \citeauthoryear {%
Sohl-Dickstein%
, Mudigonda%
\BCBL {}\ \BBA {} DeWeese%
}{%
Sohl-Dickstein%
\ \protect \BOthers {.}}{%
{\protect \APACyear {2014}}%
}]{%
sohl2014hamiltonian}
\APACinsertmetastar {%
sohl2014hamiltonian}%
\begin{APACrefauthors}%
Sohl-Dickstein, J.%
, Mudigonda, M.%
\BCBL {}\ \BBA {} DeWeese, M.%
\end{APACrefauthors}%
\unskip\
\newblock
\APACrefYearMonthDay{2014}{}{}.
\newblock
{\BBOQ}\APACrefatitle {Hamiltonian Monte Carlo without detailed balance}
  {Hamiltonian monte carlo without detailed balance}.{\BBCQ}
\newblock
\BIn{} \APACrefbtitle {International Conference on Machine Learning}
  {International conference on machine learning}\ (\BPGS\ 719--726).
\PrintBackRefs{\CurrentBib}

\bibitem [\protect \citeauthoryear {%
Tibshirani%
}{%
Tibshirani%
}{%
{\protect \APACyear {1996}}%
}]{%
tibshirani1996regression}
\APACinsertmetastar {%
tibshirani1996regression}%
\begin{APACrefauthors}%
Tibshirani, R.%
\end{APACrefauthors}%
\unskip\
\newblock
\APACrefYearMonthDay{1996}{}{}.
\newblock
{\BBOQ}\APACrefatitle {Regression shrinkage and selection via the lasso}
  {Regression shrinkage and selection via the lasso}.{\BBCQ}
\newblock
\APACjournalVolNumPages{Journal of the Royal Statistical Society: Series B
  (Methodological)}{58}{1}{267--288}.
\PrintBackRefs{\CurrentBib}

\bibitem [\protect \citeauthoryear {%
Tropp%
}{%
Tropp%
}{%
{\protect \APACyear {2006}}%
}]{%
tropp2006algorithms}
\APACinsertmetastar {%
tropp2006algorithms}%
\begin{APACrefauthors}%
Tropp, J\BPBI A.%
\end{APACrefauthors}%
\unskip\
\newblock
\APACrefYearMonthDay{2006}{}{}.
\newblock
{\BBOQ}\APACrefatitle {Algorithms for simultaneous sparse approximation. Part
  II: Convex relaxation} {Algorithms for simultaneous sparse approximation.
  part ii: Convex relaxation}.{\BBCQ}
\newblock
\APACjournalVolNumPages{Signal Processing}{86}{3}{589--602}.
\PrintBackRefs{\CurrentBib}

\bibitem [\protect \citeauthoryear {%
Wang%
\ \protect \BOthers {.}}{%
Wang%
\ \protect \BOthers {.}}{%
{\protect \APACyear {2015}}%
}]{%
wang2015sparse}
\APACinsertmetastar {%
wang2015sparse}%
\begin{APACrefauthors}%
Wang, Z.%
, Yang, J.%
, Zhang, H.%
, Wang, Z.%
, Huang, T\BPBI S.%
, Liu, D.%
\BCBL {}\ \BBA {} Yang, Y.%
\end{APACrefauthors}%
\unskip\
\newblock
\APACrefYear{2015}.
\newblock
\APACrefbtitle {Sparse coding and its applications in computer vision} {Sparse
  coding and its applications in computer vision}.
\newblock
\APACaddressPublisher{}{World Scientific}.
\PrintBackRefs{\CurrentBib}

\bibitem [\protect \citeauthoryear {%
Welling%
\ \BBA {} Teh%
}{%
Welling%
\ \BBA {} Teh%
}{%
{\protect \APACyear {2011}}%
}]{%
welling2011bayesian}
\APACinsertmetastar {%
welling2011bayesian}%
\begin{APACrefauthors}%
Welling, M.%
\BCBT {}\ \BBA {} Teh, Y\BPBI W.%
\end{APACrefauthors}%
\unskip\
\newblock
\APACrefYearMonthDay{2011}{}{}.
\newblock
{\BBOQ}\APACrefatitle {Bayesian learning via stochastic gradient Langevin
  dynamics} {Bayesian learning via stochastic gradient langevin
  dynamics}.{\BBCQ}
\newblock
\BIn{} \APACrefbtitle {Proceedings of the 28th international conference on
  machine learning (ICML-11)} {Proceedings of the 28th international conference
  on machine learning (icml-11)}\ (\BPGS\ 681--688).
\PrintBackRefs{\CurrentBib}

\bibitem [\protect \citeauthoryear {%
Wright%
\ \protect \BOthers {.}}{%
Wright%
\ \protect \BOthers {.}}{%
{\protect \APACyear {2010}}%
}]{%
wright2010sparse}
\APACinsertmetastar {%
wright2010sparse}%
\begin{APACrefauthors}%
Wright, J.%
, Ma, Y.%
, Mairal, J.%
, Sapiro, G.%
, Huang, T\BPBI S.%
\BCBL {}\ \BBA {} Yan, S.%
\end{APACrefauthors}%
\unskip\
\newblock
\APACrefYearMonthDay{2010}{}{}.
\newblock
{\BBOQ}\APACrefatitle {Sparse representation for computer vision and pattern
  recognition} {Sparse representation for computer vision and pattern
  recognition}.{\BBCQ}
\newblock
\APACjournalVolNumPages{Proceedings of the IEEE}{98}{6}{1031--1044}.
\PrintBackRefs{\CurrentBib}

\end{thebibliography}

\end{document}